\pdfoutput=1
\documentclass[10pt,twocolumn,letterpaper]{article}
\usepackage[pagenumbers]{cvpr}
\usepackage{cuted}      %
\usepackage{currfile}
\usepackage{caption}
\usepackage{booktabs}
\usepackage{multirow}
\usepackage{makecell}
\usepackage{xspace}
\usepackage{bm}
\usepackage[dvipsnames]{xcolor}
\usepackage{microtype}

\newcommand{\method}{Ego2Wrist\xspace}

\newcommand{\nLpOursCfgThreeAll}{0.365\xspace}
\newcommand{\nLpOursCfgThreeUnseen}{0.421\xspace}
\newcommand{\nLpOursCfgThreeEgoPro}{0.419\xspace}
\newcommand{\nLpOursCfgThreeEgoTouch}{0.248\xspace}
\newcommand{\nLpOursCfgThreeEgoViz}{0.496\xspace}
\newcommand{\nLpOursCfgThreeHumanTouch}{0.349\xspace}
\newcommand{\nPsnrOursCfgThreeAll}{17.11\xspace}
\newcommand{\nLpOursCfgOneAll}{0.374\xspace}
\newcommand{\nLpOursCfgOneEgoPro}{0.425\xspace}
\newcommand{\nLpOursCfgOneEgoTouch}{0.264\xspace}
\newcommand{\nLpOursCfgOneEgoViz}{0.496\xspace}
\newcommand{\nLpOursCfgOneHumanTouch}{0.360\xspace}
\newcommand{\nPsnrOursCfgOneAll}{17.18\xspace}
\newcommand{\nLpWwAll}{0.510\xspace}
\newcommand{\nLpWwEgoPro}{0.548\xspace}
\newcommand{\nLpWwEgoTouch}{0.415\xspace}
\newcommand{\nLpWwEgoViz}{0.577\xspace}
\newcommand{\nLpWwHumanTouch}{0.542\xspace}
\newcommand{\nPsnrWwAll}{12.88\xspace}
\newcommand{\nLpOtherAll}{0.582\xspace}
\newcommand{\nLpOtherEgoPro}{0.650\xspace}
\newcommand{\nLpOtherEgoTouch}{0.461\xspace}
\newcommand{\nLpOtherEgoViz}{0.604\xspace}
\newcommand{\nLpOtherHumanTouch}{0.657\xspace}
\newcommand{\nPsnrOtherAll}{11.40\xspace}
\newcommand{\nLpCropAll}{0.777\xspace}
\newcommand{\nLpCropEgoPro}{0.744\xspace}
\newcommand{\nLpCropEgoTouch}{0.812\xspace}
\newcommand{\nLpCropEgoViz}{0.792\xspace}
\newcommand{\nLpCropHumanTouch}{0.752\xspace}
\newcommand{\nPsnrCropAll}{8.49\xspace}
\newcommand{\nLpBaseAll}{0.920\xspace}
\newcommand{\nLpBaseEgoPro}{0.841\xspace}
\newcommand{\nLpBaseEgoTouch}{0.997\xspace}
\newcommand{\nLpBaseEgoViz}{0.891\xspace}
\newcommand{\nLpBaseHumanTouch}{0.932\xspace}
\newcommand{\nPsnrBaseAll}{8.68\xspace}
\newcommand{\nLpWwZsN}{96\xspace}
\newcommand{\nLpWwZsAll}{0.787\xspace}
\newcommand{\nLpOursOnWwZsClips}{0.365\xspace}
\newcommand{\nDiffOursVsWwAll}{$-$0.145\xspace}
\newcommand{\nDiffOursVsWwAllCI}{[$-$0.152, $-$0.137]\xspace}
\newcommand{\nDiffOursVsWwAllBetter}{98\%\xspace}
\newcommand{\nDiffOursVsWwAllPsnr}{$+$4.24\xspace}
\newcommand{\nDiffOursVsWwEgoViz}{$-$0.081\xspace}
\newcommand{\nDiffOursVsWwHumanTouch}{$-$0.193\xspace}
\newcommand{\nDiffOursVsOtherAll}{$-$0.217\xspace}
\newcommand{\nDiffOursVsOtherAllCI}{[$-$0.228, $-$0.206]\xspace}
\newcommand{\nDiffOursVsOtherAllBetter}{97\%\xspace}
\newcommand{\nDiffOursVsCropAllBetter}{100\%\xspace}
\newcommand{\nDiffOursVsBaseAllBetter}{100\%\xspace}
\newcommand{\nRelOursVsWw}{$-$28\%\xspace}
\newcommand{\nTestClips}{375\xspace}
\newcommand{\nTestClipsEgoPro}{100\xspace}
\newcommand{\nTestClipsEgoTouch}{120\xspace}
\newcommand{\nTestClipsEgoViz}{75\xspace}
\newcommand{\nTestClipsHumanTouch}{80\xspace}
\newcommand{\nTestClipsUnseen}{71\xspace}
\newcommand{\nCfgTestAll}{$-$0.0085\xspace}
\newcommand{\nCfgTestAllCI}{[$-$0.0115, $-$0.0058]\xspace}
\newcommand{\nCfgTestAllN}{375\xspace}
\newcommand{\nCfgTestAllK}{$-$8.5\xspace}
\newcommand{\nCfgTestAllKCI}{[$-$11.5, $-$5.8]\xspace}
\newcommand{\nCfgTestAllPsnr}{$-$0.07\xspace}
\newcommand{\nCfgValThreeAll}{$-$0.0097\xspace}
\newcommand{\nCfgValThreeAllCI}{[$-$0.0134, $-$0.0062]\xspace}
\newcommand{\nCfgValThreeAllN}{120\xspace}
\newcommand{\nCfgValThreeAllK}{$-$9.7\xspace}
\newcommand{\nCfgValThreeAllKCI}{[$-$13.4, $-$6.2]\xspace}
\newcommand{\nCfgValFiveAllN}{120\xspace}
\newcommand{\nCfgValFiveAllK}{$-$1.0\xspace}
\newcommand{\nCfgValFiveAllKCI}{[$-$5.3, $+$3.1]\xspace}
\newcommand{\nCfgValN}{120\xspace}
\newcommand{\nHeadSwapAll}{$-$0.201\xspace}
\newcommand{\nHeadSwapAllCI}{[$-$0.211, $-$0.191]\xspace}
\newcommand{\nHeadSwapAllBetter}{98\%\xspace}
\newcommand{\nHeadSwapLpRealAll}{0.374\xspace}
\newcommand{\nHeadSwapLpShuffleAll}{0.575\xspace}
\newcommand{\nCamNarrowAll}{$-$0.127\xspace}
\newcommand{\nCamNarrowAllN}{200\xspace}
\newcommand{\nCamRollAll}{$-$0.169\xspace}
\newcommand{\nCamFisheyeAll}{$-$0.079\xspace}
\newcommand{\nCamSigCombos}{11\xspace}
\newcommand{\nCamCombos}{12\xspace}
\newcommand{\nCamNativeAll}{$+$0.012\xspace}
\newcommand{\nCamNativeAllCI}{[$+$0.009, $+$0.016]\xspace}
\newcommand{\nCamNativeEgoTouch}{$+$0.023\xspace}
\newcommand{\nLpCamNativeAll}{0.386\xspace}
\newcommand{\nWarpNarrowAll}{$-$0.030\xspace}
\newcommand{\nWarpNarrowAllCI}{[$-$0.036, $-$0.024]\xspace}
\newcommand{\nCamLpNarrowWarpAll}{0.501\xspace}
\newcommand{\nCamGapNarrow}{81\%\xspace}
\newcommand{\nWarpRollAll}{$-$0.061\xspace}
\newcommand{\nWarpRollAllCI}{[$-$0.068, $-$0.055]\xspace}
\newcommand{\nCamLpRollWarpAll}{0.341\xspace}
\newcommand{\nCamGapRoll}{74\%\xspace}
\newcommand{\nWarpFisheyeAll}{$-$0.029\xspace}
\newcommand{\nWarpFisheyeAllCI}{[$-$0.034, $-$0.025]\xspace}
\newcommand{\nCamLpFisheyeWarpAll}{0.423\xspace}
\newcommand{\nCamGapFisheye}{73\%\xspace}
\newcommand{\nWarpSigCombos}{12\xspace}
\newcommand{\nWarpNativeAll}{$-$0.012\xspace}
\newcommand{\nWarpReproMaxDiff}{0.001\xspace}
\newcommand{\nWarpCoverRoll}{67--85\%\xspace}
\newcommand{\nWarpCoverFisheye}{90--100\%\xspace}
\newcommand{\nRsiDvsCAll}{$-$0.0020\xspace}
\newcommand{\nRsiDvsCAllCI}{[$-$0.0036, $-$0.0007]\xspace}
\newcommand{\nRsiDvsCAllN}{375\xspace}
\newcommand{\nRsiDvsCAllK}{$-$2.0\xspace}
\newcommand{\nRsiDvsCAllKCI}{[$-$3.6, $-$0.7]\xspace}
\newcommand{\nRsiDvsMAll}{$-$0.00111\xspace}
\newcommand{\nRsiDvsMAllCI}{[$-$0.00236, $-$0.00003]\xspace}
\newcommand{\nRsiDvsMAllK}{$-$1.1\xspace}
\newcommand{\nRsiDvsMAllKCI}{[$-$2.4, $-$0.03]\xspace}
\newcommand{\nRsiCvsMAll}{$+$0.0009\xspace}
\newcommand{\nRsiCvsMAllCI}{[$-$0.0005, $+$0.0026]\xspace}
\newcommand{\nRsiCvsMAllK}{$+$0.9\xspace}
\newcommand{\nRsiCvsMAllKCI}{[$-$0.5, $+$2.6]\xspace}
\newcommand{\nRsiExtDvsCAll}{$+$0.0002\xspace}
\newcommand{\nRsiExtDvsCAllCI}{[$-$0.00004, $+$0.0005]\xspace}
\newcommand{\nRsiExtDvsCAllN}{600\xspace}
\newcommand{\nRsiExtDvsMAll}{$+$0.0003\xspace}
\newcommand{\nRsiExtDvsMAllCI}{[$-$0.0003, $+$0.0009]\xspace}
\newcommand{\nRsiTwoVsOneAll}{$-$0.0014\xspace}
\newcommand{\nRsiTwoVsOneAllCI}{[$-$0.0028, $-$0.00001]\xspace}
\newcommand{\nRsiTwoExtVsOneAll}{$-$0.0014\xspace}
\newcommand{\nRsiTwoExtVsOneAllCI}{[$-$0.0019, $-$0.0008]\xspace}
\newcommand{\nRsiTwoPool}{39{,}780\xspace}
\newcommand{\nRsiTwoKept}{11{,}934\xspace}
\newcommand{\nRsiThreeVsTwoAll}{$-$0.0035\xspace}
\newcommand{\nRsiThreeVsTwoAllCI}{[$-$0.0057, $-$0.0016]\xspace}
\newcommand{\nRsiThreeVsZeroAll}{$-$0.0068\xspace}
\newcommand{\nRsiThreeVsZeroAllCI}{[$-$0.0100, $-$0.0040]\xspace}
\newcommand{\nRsiCandHoursTotal}{53.6\xspace}
\newcommand{\nRsiVsCfgShare}{80\%\xspace}
\newcommand{\nRsiThreeExtVsTwoAll}{$-$0.0025\xspace}
\newcommand{\nRsiThreeExtVsTwoAllCI}{[$-$0.0030, $-$0.0021]\xspace}
\newcommand{\nRsiThreeExtVsZeroAll}{$-$0.0035\xspace}
\newcommand{\nRsiThreeExtVsZeroAllCI}{[$-$0.0044, $-$0.0026]\xspace}
\newcommand{\nRsiThreePool}{39{,}648\xspace}
\newcommand{\nRsiThreeKept}{11{,}894\xspace}
\newcommand{\nRsiPool}{38{,}657\xspace}
\newcommand{\nRsiKept}{11{,}597\xspace}
\newcommand{\nRsiCut}{0.535\xspace}
\newcommand{\nXfamN}{82\xspace}
\newcommand{\nXfamDiff}{$-$0.012\xspace}
\newcommand{\nXfamDiffCI}{[$-$0.018, $-$0.006]\xspace}
\newcommand{\nXfamOursHigher}{27\xspace}
\newcommand{\nXfamLpOursBetter}{80\xspace}
\newcommand{\nXfamThreeDiff}{$+$0.013\xspace}
\newcommand{\nXfamThreeDiffCI}{[$+$0.008, $+$0.019]\xspace}
\newcommand{\nXfamThreeOursHigher}{56\xspace}
\newcommand{\nXfamThreeLpOursBetter}{79\xspace}
\newcommand{\nXfamGuidanceGain}{$+$0.025\xspace}
\newcommand{\nXfamGuidanceGainCI}{[$+$0.021, $+$0.028]\xspace}
\newcommand{\nCropHccN}{373\xspace}
\newcommand{\nCropHcc}{0.781\xspace}
\newcommand{\nCropHccCI}{[0.775, 0.787]\xspace}
\newcommand{\nCropHccReal}{0.619\xspace}
\newcommand{\nCropHccRealCI}{[0.616, 0.623]\xspace}
\newcommand{\nCropHccOverRealShare}{100\%\xspace}
\newcommand{\nCropHccSubN}{82\xspace}
\newcommand{\nCropHccSubOurs}{0.590\xspace}
\newcommand{\nCropHccSubWw}{0.577\xspace}
\newcommand{\nCropHccSub}{0.775\xspace}
\newcommand{\nCropHccSubReal}{0.620\xspace}
\newcommand{\nCropHccMinusOurs}{$+$0.185\xspace}
\newcommand{\nCropHccMinusOursCI}{[$+$0.170, $+$0.200]\xspace}
\newcommand{\nCropHccSubOursOne}{0.565\xspace}
\newcommand{\nCropSubLpCrop}{0.772\xspace}
\newcommand{\nCropSubLpOursThree}{0.387\xspace}
\newcommand{\nCropSubLpOursOne}{0.390\xspace}
\newcommand{\nCropSubLpWw}{0.516\xspace}
\newcommand{\nCropHccPartCropPrec}{0.779\xspace}
\newcommand{\nCropHccPartCropRec}{0.771\xspace}
\newcommand{\nCropHccPartRealPrec}{0.602\xspace}
\newcommand{\nCropHccPartRealRec}{0.638\xspace}
\newcommand{\nCopyOwnKeptROne}{0.770\xspace}
\newcommand{\nCopyOwnDropROne}{0.783\xspace}
\newcommand{\nCopyOwnDiffROne}{$-$0.013\xspace}
\newcommand{\nCopyOwnDiffROneCI}{[$-$0.016, $-$0.009]\xspace}
\newcommand{\nCopyNPerGroup}{300\xspace}
\newcommand{\nCopyOwnKeptRTwo}{0.806\xspace}
\newcommand{\nCopyOwnDropRTwo}{0.825\xspace}
\newcommand{\nCopyOwnDiffRTwo}{$-$0.019\xspace}
\newcommand{\nCopyOwnDiffRTwoCI}{[$-$0.023, $-$0.015]\xspace}
\newcommand{\nCopyOwnKeptRThree}{0.820\xspace}
\newcommand{\nCopyOwnDropRThree}{0.830\xspace}
\newcommand{\nCopyOwnDiffRThree}{$-$0.010\xspace}
\newcommand{\nCopyOwnDiffRThreeCI}{[$-$0.013, $-$0.007]\xspace}
\newcommand{\nCopyOwnKeptPool}{0.798\xspace}
\newcommand{\nCopyOwnDropPool}{0.812\xspace}
\newcommand{\nCopyOwnDiffPool}{$-$0.014\xspace}
\newcommand{\nCopyOwnDiffPoolCI}{[$-$0.016, $-$0.012]\xspace}
\newcommand{\nCopyHccDiffPool}{$+$0.042\xspace}
\newcommand{\nCopyMisDiffROne}{$-$0.013\xspace}
\newcommand{\nCopyMisDiffROneCI}{[$-$0.017, $-$0.009]\xspace}
\newcommand{\nCopyMisDiffRTwo}{$-$0.019\xspace}
\newcommand{\nCopyMisDiffRTwoCI}{[$-$0.023, $-$0.016]\xspace}
\newcommand{\nCopyMisDiffRThree}{$-$0.011\xspace}
\newcommand{\nCopyMisDiffRThreeCI}{[$-$0.014, $-$0.007]\xspace}
\newcommand{\nCopyMisDiffPool}{$-$0.014\xspace}
\newcommand{\nCopyMisDiffPoolCI}{[$-$0.016, $-$0.012]\xspace}
\newcommand{\nCopySpecDiffROne}{$+$0.0004\xspace}
\newcommand{\nCopySpecDiffROneCI}{[$-$0.0039, $+$0.0047]\xspace}
\newcommand{\nCopySpecDiffRTwo}{$+$0.0007\xspace}
\newcommand{\nCopySpecDiffRTwoCI}{[$-$0.0037, $+$0.0050]\xspace}
\newcommand{\nCopySpecDiffRThree}{$+$0.0003\xspace}
\newcommand{\nCopySpecDiffRThreeCI}{[$-$0.0032, $+$0.0039]\xspace}
\newcommand{\nCopySpecDiffPool}{$+$0.0005\xspace}
\newcommand{\nCopySpecDiffPoolCI}{[$-$0.0019, $+$0.0028]\xspace}
\newcommand{\nCopyNTotal}{1{,}800\xspace}
\newcommand{\nCopyReal}{0.770\xspace}
\newcommand{\nCopyRealCI}{[0.765, 0.774]\xspace}
\newcommand{\nCopyRealPfive}{0.703\xspace}
\newcommand{\nCopyShiftShare}{1.8\%\xspace}
\newcommand{\nCopyMin}{0.712\xspace}
\newcommand{\nCopyFactoryDiff}{$-$0.007\xspace}
\newcommand{\nRoiCrossOneFive}{0.414\xspace}
\newcommand{\nRoiCrossTwoFive}{0.441\xspace}
\newcommand{\nRoiCrossFour}{0.431\xspace}
\newcommand{\nHumOursCross}{$+$0.441\xspace}
\newcommand{\nHumOursCrossCI}{[$+$0.35, $+$0.52]\xspace}
\newcommand{\nHumOursSame}{$+$0.197\xspace}
\newcommand{\nHumOursSameCI}{[$+$0.11, $+$0.28]\xspace}
\newcommand{\nHumNCross}{407\xspace}
\newcommand{\nHumNSame}{542\xspace}
\newcommand{\nHumWholeCross}{$+$0.394\xspace}
\newcommand{\nHumWholeCrossCI}{[$+$0.30, $+$0.48]\xspace}
\newcommand{\nHumWholeSame}{$+$0.158\xspace}
\newcommand{\nHumWholeSameCI}{[$+$0.07, $+$0.25]\xspace}
\newcommand{\nHumMastrCross}{$+$0.239\xspace}
\newcommand{\nHumMastrCrossCI}{[$+$0.12, $+$0.35]\xspace}
\newcommand{\nHumMastrSame}{$+$0.095\xspace}
\newcommand{\nHumMastrSameCI}{[0.00, $+$0.19]\xspace}
\newcommand{\nHumClipCross}{$+$0.232\xspace}
\newcommand{\nHumClipCrossCI}{[$+$0.13, $+$0.34]\xspace}
\newcommand{\nHumClipSame}{$+$0.096\xspace}
\newcommand{\nHumClipSameCI}{[$+$0.01, $+$0.19]\xspace}
\newcommand{\nHumMetrCross}{$-$0.116\xspace}
\newcommand{\nHumMetrCrossCI}{[$-$0.23, $+$0.01]\xspace}
\newcommand{\nHumMetrSame}{$+$0.064\xspace}
\newcommand{\nHumMetrSameCI}{[$-$0.02, $+$0.15]\xspace}
\newcommand{\nHumLpipsCross}{$+$0.410\xspace}
\newcommand{\nHumLpipsCrossCI}{[$+$0.30, $+$0.51]\xspace}
\newcommand{\nHumLpipsSame}{$+$0.104\xspace}
\newcommand{\nHumLpipsSameCI}{[$+$0.01, $+$0.20]\xspace}
\newcommand{\nHumHumanCross}{0.42\xspace}
\newcommand{\nHumHumanSame}{0.22\xspace}
\newcommand{\nHumAnswers}{1{,}200\xspace}
\newcommand{\nHumRaters}{6\xspace}
\newcommand{\nHumScored}{949\xspace}
\newcommand{\nHumAnalysed}{950\xspace}
\newcommand{\nHumAnalysedRaters}{5\xspace}
\newcommand{\nHumExcluded}{1\xspace}
\newcommand{\nAucSwapOurs}{0.84\xspace}
\newcommand{\nAucSwapMastr}{0.81\xspace}
\newcommand{\nAucSwapMetr}{0.47\xspace}
\newcommand{\nAucDriftOurs}{0.75\xspace}
\newcommand{\nCeilOracleEight}{$-$0.0221\xspace}
\newcommand{\nCeilRankerCapture}{5\%\xspace}
\newcommand{\nCeilFeatCapture}{9\%\xspace}
\newcommand{\nCeilScoreCapture}{14\%\xspace}
\newcommand{\nHoursEgoPro}{771\xspace}
\newcommand{\nHoursEgoTouch}{15\xspace}
\newcommand{\nHoursEgoViz}{41\xspace}
\newcommand{\nHoursHumanTouch}{114\xspace}
\newcommand{\nHoursTotal}{940\xspace}
\newcommand{\nTrainWindows}{2M\xspace}
\newcommand{\nWwWindows}{128K\xspace}
\newcommand{\nWarpZeroAll}{$-$0.0043\xspace}
\newcommand{\nWarpZeroAllN}{160\xspace}
\newcommand{\nWarpZeroAllK}{$-$4.3\xspace}
\newcommand{\nWarpZeroAllKCI}{[$-$8.6, $-$0.04]\xspace}
\newcommand{\nWarpTwentyAllN}{160\xspace}
\newcommand{\nWarpTwentyAllK}{$-$4.0\xspace}
\newcommand{\nWarpTwentyAllKCI}{[$-$8.1, $+$0.01]\xspace}
\newcommand{\nWarpFortyAll}{$-$0.0015\xspace}
\newcommand{\nWarpFortyAllN}{160\xspace}
\newcommand{\nWarpFortyAllK}{$-$1.5\xspace}
\newcommand{\nWarpFortyAllKCI}{[$-$5.4, $+$2.5]\xspace}
\newcommand{\nZoomAll}{$+$0.0039\xspace}
\newcommand{\nZoomAllN}{200\xspace}
\newcommand{\nZoomAllK}{$+$3.9\xspace}
\newcommand{\nZoomAllKCI}{[$+$0.01, $+$7.8]\xspace}
\newcommand{\nCamLpNarrowWithAll}{0.531\xspace}
\newcommand{\nCamLpNarrowWithoutAll}{0.658\xspace}
\newcommand{\nCamLpRollWithAll}{0.402\xspace}
\newcommand{\nCamLpRollWithoutAll}{0.572\xspace}
\newcommand{\nCamLpFisheyeWithAll}{0.451\xspace}
\newcommand{\nCamLpFisheyeWithoutAll}{0.531\xspace}
\newcommand{\nCtEgoTouchClips}{198\xspace}
\newcommand{\nCtEgoTouchRows}{6{,}706\xspace}
\newcommand{\nCtEgoTouchHeadCropAuc}{0.941\xspace}
\newcommand{\nCtEgoTouchHeadCropAucCI}{[0.921, 0.958]\xspace}
\newcommand{\nCtEgoTouchHeadCropWithin}{0.801\xspace}
\newcommand{\nCtEgoTouchHeadCropWithinCI}{[0.753, 0.852]\xspace}
\newcommand{\nCtEgoTouchHeadFullAuc}{0.922\xspace}
\newcommand{\nCtEgoTouchHeadFullAucCI}{[0.893, 0.945]\xspace}
\newcommand{\nCtEgoTouchHeadFullWithin}{0.728\xspace}
\newcommand{\nCtEgoTouchHeadFullWithinCI}{[0.659, 0.795]\xspace}
\newcommand{\nCtEgoTouchWristAuc}{0.937\xspace}
\newcommand{\nCtEgoTouchWristAucCI}{[0.910, 0.960]\xspace}
\newcommand{\nCtEgoTouchWristWithin}{0.756\xspace}
\newcommand{\nCtEgoTouchWristWithinCI}{[0.690, 0.819]\xspace}
\newcommand{\nCtEgoTouchGenAuc}{0.799\xspace}
\newcommand{\nCtEgoTouchGenAucCI}{[0.744, 0.848]\xspace}
\newcommand{\nCtEgoTouchGenWithin}{0.608\xspace}
\newcommand{\nCtEgoTouchGenWithinCI}{[0.536, 0.678]\xspace}
\newcommand{\nCtEgoTouchPriorAuc}{0.725\xspace}
\newcommand{\nCtEgoTouchPriorAucCI}{[0.666, 0.779]\xspace}
\newcommand{\nCtEgoTouchPriorWithin}{0.500\xspace}
\newcommand{\nCtEgoTouchWithinN}{65\xspace}
\newcommand{\nCtEgoTouchVaeAuc}{0.931\xspace}
\newcommand{\nCtEgoTouchVaeAucCI}{[0.901, 0.956]\xspace}
\newcommand{\nCtEgoTouchVaeWithin}{0.725\xspace}
\newcommand{\nCtEgoTouchVaeWithinCI}{[0.654, 0.794]\xspace}
\newcommand{\nCtEgoTouchWristMinusCrop}{$-$0.004\xspace}
\newcommand{\nCtEgoTouchWristMinusCropCI}{[$-$0.025, $+$0.014]\xspace}
\newcommand{\nCtEgoTouchWristMinusCropWithin}{$-$0.045\xspace}
\newcommand{\nCtEgoTouchWristMinusCropWithinCI}{[$-$0.119, $+$0.027]\xspace}
\newcommand{\nCtEgoTouchGenMinusCrop}{$-$0.142\xspace}
\newcommand{\nCtEgoTouchGenMinusCropCI}{[$-$0.194, $-$0.099]\xspace}
\newcommand{\nCtEgoTouchGenMinusCropWithin}{$-$0.193\xspace}
\newcommand{\nCtEgoTouchGenMinusCropWithinCI}{[$-$0.272, $-$0.111]\xspace}
\newcommand{\nCtEgoTouchWristMinusGen}{$+$0.138\xspace}
\newcommand{\nCtEgoTouchWristMinusGenCI}{[$+$0.095, $+$0.187]\xspace}
\newcommand{\nCtEgoTouchWristMinusGenWithin}{$+$0.148\xspace}
\newcommand{\nCtEgoTouchWristMinusGenWithinCI}{[$+$0.079, $+$0.214]\xspace}
\newcommand{\nCtEgoTouchGenMinusPrior}{$+$0.075\xspace}
\newcommand{\nCtEgoTouchGenMinusPriorCI}{[$+$0.006, $+$0.143]\xspace}
\newcommand{\nCtEgoTouchGenMinusPriorWithin}{$+$0.108\xspace}
\newcommand{\nCtEgoTouchGenMinusPriorWithinCI}{[$+$0.036, $+$0.178]\xspace}
\newcommand{\nCtEgoTouchWristMinusVae}{$+$0.006\xspace}
\newcommand{\nCtEgoTouchWristMinusVaeCI}{[$+$0.002, $+$0.010]\xspace}
\newcommand{\nCtEgoTouchWristMinusVaeWithin}{$+$0.027\xspace}
\newcommand{\nCtEgoTouchWristMinusVaeWithinCI}{[$+$0.007, $+$0.048]\xspace}
\newcommand{\nCtHumanTouchClips}{200\xspace}
\newcommand{\nCtHumanTouchRows}{7{,}755\xspace}
\newcommand{\nCtHumanTouchHeadCropAuc}{0.756\xspace}
\newcommand{\nCtHumanTouchHeadCropAucCI}{[0.718, 0.794]\xspace}
\newcommand{\nCtHumanTouchHeadCropWithin}{0.679\xspace}
\newcommand{\nCtHumanTouchHeadCropWithinCI}{[0.645, 0.714]\xspace}
\newcommand{\nCtHumanTouchHeadFullAuc}{0.715\xspace}
\newcommand{\nCtHumanTouchHeadFullAucCI}{[0.672, 0.756]\xspace}
\newcommand{\nCtHumanTouchHeadFullWithin}{0.615\xspace}
\newcommand{\nCtHumanTouchHeadFullWithinCI}{[0.577, 0.652]\xspace}
\newcommand{\nCtHumanTouchWristAuc}{0.721\xspace}
\newcommand{\nCtHumanTouchWristAucCI}{[0.670, 0.768]\xspace}
\newcommand{\nCtHumanTouchWristWithin}{0.662\xspace}
\newcommand{\nCtHumanTouchWristWithinCI}{[0.624, 0.699]\xspace}
\newcommand{\nCtHumanTouchGenAuc}{0.680\xspace}
\newcommand{\nCtHumanTouchGenAucCI}{[0.640, 0.720]\xspace}
\newcommand{\nCtHumanTouchGenWithin}{0.626\xspace}
\newcommand{\nCtHumanTouchGenWithinCI}{[0.594, 0.658]\xspace}
\newcommand{\nCtHumanTouchPriorAuc}{0.569\xspace}
\newcommand{\nCtHumanTouchPriorAucCI}{[0.516, 0.619]\xspace}
\newcommand{\nCtHumanTouchPriorWithin}{0.500\xspace}
\newcommand{\nCtHumanTouchWithinN}{156\xspace}
\newcommand{\nCtHumanTouchVaeAuc}{0.721\xspace}
\newcommand{\nCtHumanTouchVaeAucCI}{[0.670, 0.766]\xspace}
\newcommand{\nCtHumanTouchVaeWithin}{0.673\xspace}
\newcommand{\nCtHumanTouchVaeWithinCI}{[0.637, 0.707]\xspace}
\newcommand{\nCtHumanTouchWristMinusCrop}{$-$0.035\xspace}
\newcommand{\nCtHumanTouchWristMinusCropCI}{[$-$0.080, $+$0.006]\xspace}
\newcommand{\nCtHumanTouchWristMinusCropWithin}{$-$0.017\xspace}
\newcommand{\nCtHumanTouchWristMinusCropWithinCI}{[$-$0.058, $+$0.021]\xspace}
\newcommand{\nCtHumanTouchGenMinusCrop}{$-$0.076\xspace}
\newcommand{\nCtHumanTouchGenMinusCropCI}{[$-$0.120, $-$0.033]\xspace}
\newcommand{\nCtHumanTouchGenMinusCropWithin}{$-$0.053\xspace}
\newcommand{\nCtHumanTouchGenMinusCropWithinCI}{[$-$0.093, $-$0.013]\xspace}
\newcommand{\nCtHumanTouchWristMinusGen}{$+$0.041\xspace}
\newcommand{\nCtHumanTouchWristMinusGenCI}{[$-$0.006, $+$0.086]\xspace}
\newcommand{\nCtHumanTouchWristMinusGenWithin}{$+$0.036\xspace}
\newcommand{\nCtHumanTouchWristMinusGenWithinCI}{[$-$0.004, $+$0.076]\xspace}
\newcommand{\nCtHumanTouchGenMinusPrior}{$+$0.111\xspace}
\newcommand{\nCtHumanTouchGenMinusPriorCI}{[$+$0.061, $+$0.162]\xspace}
\newcommand{\nCtHumanTouchGenMinusPriorWithin}{$+$0.126\xspace}
\newcommand{\nCtHumanTouchGenMinusPriorWithinCI}{[$+$0.094, $+$0.158]\xspace}
\newcommand{\nCtHumanTouchWristMinusVae}{$-$0.001\xspace}
\newcommand{\nCtHumanTouchWristMinusVaeCI}{[$-$0.007, $+$0.004]\xspace}
\newcommand{\nCtHumanTouchWristMinusVaeWithin}{$-$0.005\xspace}
\newcommand{\nCtHumanTouchWristMinusVaeWithinCI}{[$-$0.017, $+$0.007]\xspace}
\newcommand{\nCtThrEgoTouchLo}{2\xspace}
\newcommand{\nCtThrEgoTouchHi}{10\xspace}
\newcommand{\nCtThrHumanTouchLo}{0.5\xspace}
\newcommand{\nCtThrHumanTouchHi}{3\xspace}
\newcommand{\nCtPcs}{1{,}024\xspace}
\newcommand{\nCtBoot}{2{,}000\xspace}
\newcommand{\nCtCrop}{224\xspace}
\newcommand{\nCtRoi}{2.5\xspace}
\newcommand{\nStudyHOneA}{48\xspace}
\newcommand{\nStudyHOneAPerDs}{12\xspace}
\newcommand{\nStudyHOneB}{48\xspace}
\newcommand{\nStudyHTwo}{60\xspace}
\newcommand{\nStudyHOneBEligible}{46{,}889\xspace}
\newcommand{\nStudyHOneBFactories}{39\xspace}
\newcommand{\nStudyHTwoEligible}{1{,}620\xspace}
\newcommand{\nStudyHTwoFactories}{43\xspace}
\newcommand{\nStudyHTwoGapFrames}{150\xspace}
\newcommand{\nStudyHTwoScoreGap}{0.044\xspace}
\newcommand{\nStudyUnique}{156\xspace}
\newcommand{\nStudyChecks}{6\xspace}
\newcommand{\nStudyBlocks}{5\xspace}
\newcommand{\nStudyBlocksPerQ}{3\xspace}
\newcommand{\nStudyRepeats}{4\xspace}
\newcommand{\nStudyUniqueMin}{93\xspace}
\newcommand{\nStudyUniqueMax}{95\xspace}
\newcommand{\nStudyPerRaterMin}{103\xspace}
\newcommand{\nStudyPerRaterMax}{105\xspace}
\newcommand{\nStudyHOneBPerFactory}{2\xspace}
\newcommand{\nStudyHTwoPerFactory}{2\xspace}
\newcommand{\nRsiKeepShare}{30\%\xspace}
\newcommand{\nStudyRepeatGap}{40\xspace}
\newcommand{\nStudyMinChecks}{3\xspace}
\newcommand{\nStudyFailExclude}{2\xspace}
\newcommand{\nStudyRatersTarget}{5\xspace}
\newcommand{\nStudyBoot}{10{,}000\xspace}
\newcommand{\nStudyRaters}{5\xspace}
\newcommand{\nStudyAnswers}{518\xspace}
\newcommand{\nStudyHOneAPref}{76\%\xspace}
\newcommand{\nStudyHOneAPrefCI}{[69\%, 83\%]\xspace}
\newcommand{\nStudyHOneAMajFor}{41\xspace}
\newcommand{\nStudyHOneAMajN}{48\xspace}
\newcommand{\nStudyHOneAItems}{48\xspace}
\newcommand{\nStudyHOneAEgoPro}{58\%\xspace}
\newcommand{\nStudyHOneAEgoProCI}{[44\%, 72\%]\xspace}
\newcommand{\nStudyHOneAEgoTouch}{78\%\xspace}
\newcommand{\nStudyHOneAEgoTouchCI}{[69\%, 86\%]\xspace}
\newcommand{\nStudyHOneAEgoViz}{81\%\xspace}
\newcommand{\nStudyHOneAEgoVizCI}{[67\%, 94\%]\xspace}
\newcommand{\nStudyHOneAHumanTouch}{89\%\xspace}
\newcommand{\nStudyHOneAHumanTouchCI}{[81\%, 97\%]\xspace}
\newcommand{\nStudyHOneBPref}{51\%\xspace}
\newcommand{\nStudyHOneBPrefCI}{[42\%, 61\%]\xspace}
\newcommand{\nStudyHOneBMajFor}{24\xspace}
\newcommand{\nStudyHOneBMajN}{48\xspace}
\newcommand{\nStudyHOneBMajAgainst}{24\xspace}
\newcommand{\nStudyHOneBHcc}{69\%\xspace}
\newcommand{\nStudyHOneBHccCI}{[60\%, 77\%]\xspace}
\newcommand{\nStudyHOneBHccMaj}{37\xspace}
\newcommand{\nStudyHOneBHccMajN}{48\xspace}
\newcommand{\nStudyHTwoPref}{77\%\xspace}
\newcommand{\nStudyHTwoPrefCI}{[69\%, 83\%]\xspace}
\newcommand{\nStudyHTwoMajFor}{51\xspace}
\newcommand{\nStudyHTwoMajN}{60\xspace}
\newcommand{\nStudyHTwoLow}{75\%\xspace}
\newcommand{\nStudyHTwoMid}{68\%\xspace}
\newcommand{\nStudyHTwoHigh}{87\%\xspace}
\newcommand{\nStudyHTwoHighLow}{$+$0.12\xspace}
\newcommand{\nStudyHTwoHighLowCI}{[$-$0.03, $+$0.28]\xspace}
\newcommand{\nStudyHOneAAgree}{62\%\xspace}
\newcommand{\nStudyHOneBAgree}{64\%\xspace}
\newcommand{\nStudyRepeatSame}{65\%\xspace}
\newcommand{\nStudyRepeatSameK}{13\xspace}
\newcommand{\nStudyRepeatSameN}{20\xspace}
\newcommand{\nStudyHOneAPrefAllRaters}{76\%\xspace}
\newcommand{\nStudyHOneBPrefAllRaters}{52\%\xspace}
\newcommand{\nStudyHTwoPrefAllRaters}{77\%\xspace}
\newcommand{\nStudyHOneAPrefNoEarlier}{78\%\xspace}
\newcommand{\nStudyHOneBPrefNoEarlier}{50\%\xspace}
\newcommand{\nStudyHTwoPrefNoEarlier}{77\%\xspace}
\newcommand{\nStudyRatersNoEarlier}{4\xspace}
\newcommand{\nSAucSwapOurs}{0.84\xspace}
\newcommand{\nSAucSwapPrec}{0.82\xspace}
\newcommand{\nSAucSwapRec}{0.82\xspace}
\newcommand{\nSAucSwapWhole}{0.86\xspace}
\newcommand{\nSAucSwapTime}{0.70\xspace}
\newcommand{\nSAucSwapRefl}{0.57\xspace}
\newcommand{\nSAucSwapMastr}{0.81\xspace}
\newcommand{\nSAucSwapMetr}{0.47\xspace}
\newcommand{\nSAucSwapClip}{0.71\xspace}
\newcommand{\nSAucDriftWeakOurs}{0.64\xspace}
\newcommand{\nSAucDriftWeakPrec}{0.63\xspace}
\newcommand{\nSAucDriftWeakRec}{0.59\xspace}
\newcommand{\nSAucDriftWeakWhole}{0.65\xspace}
\newcommand{\nSAucDriftWeakTime}{0.49\xspace}
\newcommand{\nSAucDriftWeakRefl}{0.50\xspace}
\newcommand{\nSAucDriftWeakMastr}{0.57\xspace}
\newcommand{\nSAucDriftWeakMetr}{0.50\xspace}
\newcommand{\nSAucDriftWeakClip}{0.58\xspace}
\newcommand{\nSAucDriftOurs}{0.75\xspace}
\newcommand{\nSAucDriftPrec}{0.73\xspace}
\newcommand{\nSAucDriftRec}{0.69\xspace}
\newcommand{\nSAucDriftWhole}{0.77\xspace}
\newcommand{\nSAucDriftTime}{0.52\xspace}
\newcommand{\nSAucDriftRefl}{0.50\xspace}
\newcommand{\nSAucDriftMastr}{0.64\xspace}
\newcommand{\nSAucDriftMetr}{0.49\xspace}
\newcommand{\nSAucDriftClip}{0.63\xspace}
\newcommand{\nSAucFlipOurs}{0.61\xspace}
\newcommand{\nSAucFlipPrec}{0.59\xspace}
\newcommand{\nSAucFlipRec}{0.61\xspace}
\newcommand{\nSAucFlipWhole}{0.59\xspace}
\newcommand{\nSAucFlipTime}{0.52\xspace}
\newcommand{\nSAucFlipRefl}{0.60\xspace}
\newcommand{\nSAucFlipMastr}{0.68\xspace}
\newcommand{\nSAucFlipMetr}{0.50\xspace}
\newcommand{\nSAucFlipClip}{0.51\xspace}
\newcommand{\nSAucRotateOurs}{0.64\xspace}
\newcommand{\nSAucRotatePrec}{0.59\xspace}
\newcommand{\nSAucRotateRec}{0.65\xspace}
\newcommand{\nSAucRotateWhole}{0.59\xspace}
\newcommand{\nSAucRotateTime}{0.52\xspace}
\newcommand{\nSAucRotateRefl}{0.63\xspace}
\newcommand{\nSAucRotateMastr}{0.63\xspace}
\newcommand{\nSAucRotateMetr}{0.49\xspace}
\newcommand{\nSAucRotateClip}{0.54\xspace}
\newcommand{\nSAucHalvesOurs}{0.63\xspace}
\newcommand{\nSAucHalvesPrec}{0.60\xspace}
\newcommand{\nSAucHalvesRec}{0.63\xspace}
\newcommand{\nSAucHalvesWhole}{0.53\xspace}
\newcommand{\nSAucHalvesTime}{0.59\xspace}
\newcommand{\nSAucHalvesRefl}{0.65\xspace}
\newcommand{\nSAucHalvesMastr}{0.62\xspace}
\newcommand{\nSAucHalvesMetr}{0.57\xspace}
\newcommand{\nSAucHalvesClip}{0.61\xspace}
\newcommand{\nSAucMirrorOurs}{0.54\xspace}
\newcommand{\nSAucMirrorPrec}{0.51\xspace}
\newcommand{\nSAucMirrorRec}{0.56\xspace}
\newcommand{\nSAucMirrorWhole}{0.51\xspace}
\newcommand{\nSAucMirrorTime}{0.50\xspace}
\newcommand{\nSAucMirrorRefl}{0.65\xspace}
\newcommand{\nSAucMirrorMastr}{0.65\xspace}
\newcommand{\nSAucMirrorMetr}{0.52\xspace}
\newcommand{\nSAucMirrorClip}{0.50\xspace}
\newcommand{\nSAucHandwarpOurs}{0.60\xspace}
\newcommand{\nSAucHandwarpPrec}{0.57\xspace}
\newcommand{\nSAucHandwarpRec}{0.58\xspace}
\newcommand{\nSAucHandwarpWhole}{0.59\xspace}
\newcommand{\nSAucHandwarpTime}{0.51\xspace}
\newcommand{\nSAucHandwarpRefl}{0.52\xspace}
\newcommand{\nSAucHandwarpMastr}{0.53\xspace}
\newcommand{\nSAucHandwarpMetr}{0.55\xspace}
\newcommand{\nSAucHandwarpClip}{0.53\xspace}
\newcommand{\nSAucElasticOurs}{0.56\xspace}
\newcommand{\nSAucElasticPrec}{0.54\xspace}
\newcommand{\nSAucElasticRec}{0.56\xspace}
\newcommand{\nSAucElasticWhole}{0.54\xspace}
\newcommand{\nSAucElasticTime}{0.51\xspace}
\newcommand{\nSAucElasticRefl}{0.50\xspace}
\newcommand{\nSAucElasticMastr}{0.52\xspace}
\newcommand{\nSAucElasticMetr}{0.55\xspace}
\newcommand{\nSAucElasticClip}{0.51\xspace}
\newcommand{\nSAucShuffleOurs}{0.53\xspace}
\newcommand{\nSAucShufflePrec}{0.52\xspace}
\newcommand{\nSAucShuffleRec}{0.52\xspace}
\newcommand{\nSAucShuffleWhole}{0.51\xspace}
\newcommand{\nSAucShuffleTime}{0.71\xspace}
\newcommand{\nSAucShuffleRefl}{0.51\xspace}
\newcommand{\nSAucShuffleMastr}{0.52\xspace}
\newcommand{\nSAucShuffleMetr}{0.51\xspace}
\newcommand{\nSAucShuffleClip}{0.49\xspace}
\newcommand{\nSAucFreezeOurs}{0.54\xspace}
\newcommand{\nSAucFreezePrec}{0.54\xspace}
\newcommand{\nSAucFreezeRec}{0.53\xspace}
\newcommand{\nSAucFreezeWhole}{0.52\xspace}
\newcommand{\nSAucFreezeTime}{0.74\xspace}
\newcommand{\nSAucFreezeRefl}{0.51\xspace}
\newcommand{\nSAucFreezeMastr}{0.56\xspace}
\newcommand{\nSAucFreezeMetr}{0.57\xspace}
\newcommand{\nSAucFreezeClip}{0.51\xspace}
\newcommand{\nSAucFlickerOurs}{0.51\xspace}
\newcommand{\nSAucFlickerPrec}{0.51\xspace}
\newcommand{\nSAucFlickerRec}{0.50\xspace}
\newcommand{\nSAucFlickerWhole}{0.51\xspace}
\newcommand{\nSAucFlickerTime}{0.50\xspace}
\newcommand{\nSAucFlickerRefl}{0.49\xspace}
\newcommand{\nSAucFlickerMastr}{0.50\xspace}
\newcommand{\nSAucFlickerMetr}{0.50\xspace}
\newcommand{\nSAucFlickerClip}{0.50\xspace}
\newcommand{\nSAucTimewarpOurs}{0.50\xspace}
\newcommand{\nSAucTimewarpPrec}{0.50\xspace}
\newcommand{\nSAucTimewarpRec}{0.50\xspace}
\newcommand{\nSAucTimewarpWhole}{0.50\xspace}
\newcommand{\nSAucTimewarpTime}{0.51\xspace}
\newcommand{\nSAucTimewarpRefl}{0.49\xspace}
\newcommand{\nSAucTimewarpMastr}{0.51\xspace}
\newcommand{\nSAucTimewarpMetr}{0.49\xspace}
\newcommand{\nSAucTimewarpClip}{0.49\xspace}
\newcommand{\nSAucJitterOurs}{0.51\xspace}
\newcommand{\nSAucJitterPrec}{0.50\xspace}
\newcommand{\nSAucJitterRec}{0.52\xspace}
\newcommand{\nSAucJitterWhole}{0.51\xspace}
\newcommand{\nSAucJitterTime}{0.51\xspace}
\newcommand{\nSAucJitterRefl}{0.49\xspace}
\newcommand{\nSAucJitterMastr}{0.51\xspace}
\newcommand{\nSAucJitterMetr}{0.48\xspace}
\newcommand{\nSAucJitterClip}{0.47\xspace}
\newcommand{\nSAucBreatheOurs}{0.51\xspace}
\newcommand{\nSAucBreathePrec}{0.51\xspace}
\newcommand{\nSAucBreatheRec}{0.51\xspace}
\newcommand{\nSAucBreatheWhole}{0.51\xspace}
\newcommand{\nSAucBreatheTime}{0.51\xspace}
\newcommand{\nSAucBreatheRefl}{0.49\xspace}
\newcommand{\nSAucBreatheMastr}{0.50\xspace}
\newcommand{\nSAucBreatheMetr}{0.50\xspace}
\newcommand{\nSAucBreatheClip}{0.50\xspace}
\newcommand{\nSAucN}{180\xspace}
\newcommand{\nHumPrecCross}{$+$0.392\xspace}
\newcommand{\nHumPrecCrossCI}{[$+$0.30, $+$0.48]\xspace}
\newcommand{\nHumPrecSame}{$+$0.176\xspace}
\newcommand{\nHumPrecSameCI}{[$+$0.09, $+$0.27]\xspace}
\newcommand{\nHumRecCross}{$+$0.443\xspace}
\newcommand{\nHumRecCrossCI}{[$+$0.35, $+$0.52]\xspace}
\newcommand{\nHumRecSame}{$+$0.190\xspace}
\newcommand{\nHumRecSameCI}{[$+$0.10, $+$0.28]\xspace}
\newcommand{\nHumMetrOvCross}{$+$0.145\xspace}
\newcommand{\nHumMetrOvCrossCI}{[$+$0.03, $+$0.26]\xspace}
\newcommand{\nHumMetrOvSame}{$+$0.055\xspace}
\newcommand{\nHumMetrOvSameCI}{[$-$0.04, $+$0.15]\xspace}
\newcommand{\nStgCTwoSteps}{12{,}000\xspace}
\newcommand{\nStgCTwoBatch}{4\xspace}
\newcommand{\nStgCTwoLr}{$10^{-4}$\xspace}
\newcommand{\nStgCTwoWindows}{48{,}000\xspace}
\newcommand{\nStgCTwoLongSteps}{48{,}000\xspace}
\newcommand{\nStgCTwoLongBatch}{6\xspace}
\newcommand{\nStgCTwoLongLr}{$10^{-4}$\xspace}
\newcommand{\nStgCTwoLongDecayAt}{52{,}000\xspace}
\newcommand{\nStgCTwoLongWindows}{288{,}000\xspace}
\newcommand{\nStgBigSteps}{4{,}000\xspace}
\newcommand{\nStgBigBatch}{24\xspace}
\newcommand{\nStgBigLr}{$5{\times}10^{-5}$\xspace}
\newcommand{\nStgBigWarm}{200\xspace}
\newcommand{\nStgBigDecaySteps}{2{,}000\xspace}
\newcommand{\nStgBigWindows}{96{,}000\xspace}
\newcommand{\nStgLongSteps}{53{,}500\xspace}
\newcommand{\nStgLongBatch}{24\xspace}
\newcommand{\nStgLongLr}{$5{\times}10^{-5}$\xspace}
\newcommand{\nStgLongWarm}{200\xspace}
\newcommand{\nStgLongDecaySteps}{4{,}500\xspace}
\newcommand{\nStgLongWindows}{1{,}284{,}000\xspace}
\newcommand{\nStgConvSteps}{12{,}550\xspace}
\newcommand{\nStgConvBatch}{24\xspace}
\newcommand{\nStgConvLr}{$5{\times}10^{-5}$\xspace}
\newcommand{\nStgConvWarm}{500\xspace}
\newcommand{\nStgConvDecayAt}{8{,}050\xspace}
\newcommand{\nStgConvDecaySteps}{4{,}500\xspace}
\newcommand{\nStgConvWindows}{301{,}200\xspace}
\newcommand{\nStgWindowsTotal}{2{,}017{,}200\xspace}
\newcommand{\nStgLongSwitch}{44{,}500\xspace}
\newcommand{\nStgLongDataSwap}{10{,}500\xspace}
\newcommand{\nStgEma}{0.9995\xspace}
\newcommand{\nStgSeedsDrop}{0.1\xspace}
\newcommand{\nStgGradClip}{1\xspace}
\newcommand{\nMixEgoPro}{0.4\xspace}
\newcommand{\nMixEgoTouch}{0.18\xspace}
\newcommand{\nMixEgoViz}{0.24\xspace}
\newcommand{\nMixHumanTouch}{0.18\xspace}
\newcommand{\nLoraParams}{188.3M\xspace}
\newcommand{\nCamParams}{236.1M\xspace}
\newcommand{\nLoraModules}{368\xspace}
\newcommand{\nShift}{12\xspace}
\newcommand{\nAdamBetaOne}{0.9\xspace}
\newcommand{\nAdamBetaTwo}{0.95\xspace}
\newcommand{\nAdamWd}{0.01\xspace}
\newcommand{\nSampleSteps}{30\xspace}
\newcommand{\nSeedBase}{1{,}000\xspace}
\newcommand{\nGpuHoursMain}{2{,}000\xspace}
\newcommand{\nGpuHoursWw}{190\xspace}
\newcommand{\nSecPerClip}{39\xspace}
\newcommand{\nStepsTwentyAll}{$+$0.0013\xspace}
\newcommand{\nStepsFifteenAll}{$+$0.0040\xspace}
\newcommand{\nStepsTenAll}{$+$0.0072\xspace}
\newcommand{\nStepsN}{200\xspace}
\newcommand{\nVcKeepWrist}{35\%\xspace}
\newcommand{\nVcKeepHead}{50\%\xspace}
\newcommand{\nVcRollPWrist}{50\%\xspace}
\newcommand{\nVcRollPHead}{30\%\xspace}
\newcommand{\nVcRollWrist}{30\xspace}
\newcommand{\nVcRollHead}{10\xspace}
\newcommand{\nVcFovLo}{60\%\xspace}
\newcommand{\nVcFovHi}{100\%\xspace}
\newcommand{\nVcPinMax}{110\xspace}
\newcommand{\nVcAspect}{3\xspace}
\newcommand{\nVcPpX}{10\xspace}
\newcommand{\nVcPpY}{8\xspace}
\newcommand{\nVcLetterP}{30\%\xspace}
\newcommand{\nVcBarLo}{24\xspace}
\newcommand{\nVcBarHi}{56\xspace}
\newcommand{\nTokPx}{32\xspace}
\newcommand{\nSubPts}{2\xspace}
\newcommand{\nRayFeat}{21\xspace}
\newcommand{\nMounts}{5\xspace}
\newcommand{\nInjHidden}{512\xspace}
\newcommand{\nInjEmb}{512\xspace}
\newcommand{\nCamNarrowFov}{70\xspace}
\newcommand{\nCamRollDeg}{25\xspace}
\newcommand{\nCamStart}{53{,}500\xspace}
\newcommand{\nCamSteps}{12{,}550\xspace}
\newcommand{\nCamBatch}{24\xspace}
\newcommand{\nCamDecayAt}{8{,}050\xspace}
\newcommand{\nCamDrop}{10\%\xspace}
\newcommand{\nWwSteps}{8{,}001\xspace}
\newcommand{\nWwBatch}{16\xspace}
\newcommand{\nWwLr}{$10^{-4}$\xspace}
\newcommand{\nWwParams}{201.5M\xspace}
\newcommand{\nWwZsEvery}{4\xspace}
\newcommand{\nLoraRank}{64\xspace}
\newcommand{\nFrames}{49\xspace}
\newcommand{\nRes}{$384{\times}512$\xspace}
\newcommand{\nResLow}{$288{\times}384$\xspace}
\newcommand{\nCanvas}{$512{\times}384$\xspace}
\newcommand{\nStepsCmpA}{20\xspace}
\newcommand{\nStepsCmpB}{15\xspace}
\newcommand{\nStepsCmpC}{10\xspace}
\newcommand{\nPoolSteps}{15\xspace}
\newcommand{\nWwSampleSteps}{50\xspace}
\newcommand{\nWwCfg}{5\xspace}
\newcommand{\nWristMounts}{4\xspace}
\newcommand{\nGuideW}{3\xspace}
\newcommand{\nEpEgoProTrain}{22{,}342\xspace}
\newcommand{\nHrEgoProTrain}{770.7\xspace}
\newcommand{\nEpEgoProVal}{90\xspace}
\newcommand{\nHrEgoProVal}{3.7\xspace}
\newcommand{\nEpEgoProSeen}{118\xspace}
\newcommand{\nHrEgoProSeen}{4.4\xspace}
\newcommand{\nEpEgoProUnseen}{292\xspace}
\newcommand{\nHrEgoProUnseen}{15.2\xspace}
\newcommand{\nEpEgoProHeld}{1{,}706\xspace}
\newcommand{\nHrEgoProHeld}{60.6\xspace}
\newcommand{\nEpEgoTouchTrain}{1{,}201\xspace}
\newcommand{\nHrEgoTouchTrain}{15.0\xspace}
\newcommand{\nEpEgoTouchVal}{135\xspace}
\newcommand{\nHrEgoTouchVal}{1.5\xspace}
\newcommand{\nEpEgoTouchSeen}{135\xspace}
\newcommand{\nHrEgoTouchSeen}{1.8\xspace}
\newcommand{\nEpEgoTouchUnseen}{65\xspace}
\newcommand{\nHrEgoTouchUnseen}{0.6\xspace}
\newcommand{\nEpEgoVizTrain}{2{,}734\xspace}
\newcommand{\nHrEgoVizTrain}{40.6\xspace}
\newcommand{\nEpEgoVizVal}{43\xspace}
\newcommand{\nHrEgoVizVal}{0.6\xspace}
\newcommand{\nEpEgoVizSeen}{75\xspace}
\newcommand{\nHrEgoVizSeen}{1.1\xspace}
\newcommand{\nEpHumanTouchTrain}{15{,}981\xspace}
\newcommand{\nHrHumanTouchTrain}{113.6\xspace}
\newcommand{\nEpHumanTouchVal}{356\xspace}
\newcommand{\nHrHumanTouchVal}{2.5\xspace}
\newcommand{\nEpHumanTouchSeen}{527\xspace}
\newcommand{\nHrHumanTouchSeen}{3.8\xspace}
\newcommand{\nEpHumanTouchUnseen}{711\xspace}
\newcommand{\nHrHumanTouchUnseen}{12.3\xspace}
\newcommand{\nEpAllTrain}{42{,}258\xspace}
\newcommand{\nHrAllTrain}{939.8\xspace}
\newcommand{\nEpAllVal}{624\xspace}
\newcommand{\nHrAllVal}{8.3\xspace}
\newcommand{\nEpAllSeen}{855\xspace}
\newcommand{\nHrAllSeen}{11.1\xspace}
\newcommand{\nEpAllUnseen}{1{,}068\xspace}
\newcommand{\nHrAllUnseen}{28.1\xspace}
\newcommand{\nEpAllHeld}{1{,}706\xspace}
\newcommand{\nHrAllHeld}{60.6\xspace}
\newcommand{\nHoursVTwo}{395\xspace}
\newcommand{\nSetStd}{375\xspace}
\newcommand{\nSetStdEgoPro}{100\xspace}
\newcommand{\nSetStdEgoProUnseen}{71\xspace}
\newcommand{\nSetStdEgoTouch}{120\xspace}
\newcommand{\nSetStdEgoTouchUnseen}{40\xspace}
\newcommand{\nSetStdEgoViz}{75\xspace}
\newcommand{\nSetStdHumanTouch}{80\xspace}
\newcommand{\nSetStdHumanTouchUnseen}{45\xspace}
\newcommand{\nSetVal}{120\xspace}
\newcommand{\nSetValEgoPro}{30\xspace}
\newcommand{\nSetExt}{600\xspace}
\newcommand{\nSetCompact}{200\xspace}
\newcommand{\nSetCompactEgoPro}{60\xspace}
\newcommand{\nSetCompactEgoTouch}{40\xspace}
\newcommand{\nSetCompactEgoViz}{60\xspace}
\newcommand{\nSetCompactHumanTouch}{40\xspace}
\newcommand{\nSetCompactOverlap}{118\xspace}
\newcommand{\nSetExtUnseen}{87\xspace}
\newcommand{\nSetExtHeld}{513\xspace}
\newcommand{\nFps}{30\xspace}
\newcommand{\nUnselEgoPro}{22\xspace}
\newcommand{\nUnselOther}{20\xspace}
\newcommand{\nWildClips}{2{,}916\xspace}
\newcommand{\nWildCand}{114{,}592\xspace}
\newcommand{\nWildHands}{55{,}437\xspace}
\newcommand{\nWildScreened}{48{,}192\xspace}
\newcommand{\nWildScorable}{48{,}087\xspace}
\newcommand{\nWildPoolHours}{17.5\xspace}
\newcommand{\nWildKeptHours}{5.3\xspace}
\newcommand{\nWildFactoriesKept}{81\xspace}
\newcommand{\nWildFactoriesPool}{85\xspace}
\newcommand{\nCalEgoProHeadH}{101\xspace}
\newcommand{\nCalEgoProHeadV}{85\xspace}
\newcommand{\nCalEgoProLeftH}{78\xspace}
\newcommand{\nCalEgoProLeftV}{65\xspace}
\newcommand{\nCalEgoProRightH}{86\xspace}
\newcommand{\nCalEgoProRightV}{72\xspace}
\newcommand{\nCalEgoTouchHeadH}{88\xspace}
\newcommand{\nCalEgoTouchHeadV}{67\xspace}
\newcommand{\nCalEgoTouchLeftH}{154\xspace}
\newcommand{\nCalEgoTouchLeftV}{145\xspace}
\newcommand{\nCalEgoTouchRightH}{167\xspace}
\newcommand{\nCalEgoTouchRightV}{147\xspace}
\newcommand{\nCalEgoVizHeadH}{94\xspace}
\newcommand{\nCalEgoVizHeadV}{68\xspace}
\newcommand{\nCalEgoVizLeftH}{107\xspace}
\newcommand{\nCalEgoVizLeftV}{82\xspace}
\newcommand{\nCalEgoVizRightH}{108\xspace}
\newcommand{\nCalEgoVizRightV}{82\xspace}
\newcommand{\nCalHumanTouchHeadH}{96\xspace}
\newcommand{\nCalHumanTouchHeadV}{62\xspace}
\newcommand{\nCalHumanTouchLeftH}{111\xspace}
\newcommand{\nCalHumanTouchLeftV}{83\xspace}
\newcommand{\nCalHumanTouchRightH}{116\xspace}
\newcommand{\nCalHumanTouchRightV}{88\xspace}
\newcommand{\nCalWildHeadH}{85\xspace}
\newcommand{\nCalWildHeadV}{62\xspace}
\newcommand{\nCalBugR}{1.35\xspace}
\newcommand{\nCalBugDeg}{5.7\xspace}
\newcommand{\nCalBugFlatR}{5\xspace}
\newcommand{\nCalBugDisk}{37\xspace}
\newcommand{\nCalImpNarrow}{$-$0.129\xspace}
\newcommand{\nCalImpNarrowCI}{[$-$0.142, $-$0.117]\xspace}
\newcommand{\nCalEtNarrowL}{$-$0.257\xspace}
\newcommand{\nCalEtNarrowR}{$-$0.241\xspace}
\newcommand{\nCalImpRoll}{$-$0.169\xspace}
\newcommand{\nCalImpRollCI}{[$-$0.181, $-$0.158]\xspace}
\newcommand{\nCalEtRollL}{$-$0.032\xspace}
\newcommand{\nCalEtRollR}{$-$0.033\xspace}
\newcommand{\nCalImpFisheye}{$-$0.083\xspace}
\newcommand{\nCalImpFisheyeCI}{[$-$0.094, $-$0.072]\xspace}
\newcommand{\nCalEtFisheyeL}{$-$0.191\xspace}
\newcommand{\nCalEtFisheyeR}{$-$0.154\xspace}
\newcommand{\nCalEtN}{40\xspace}
\newcommand{\nCalNativeL}{$+$0.023\xspace}
\newcommand{\nCalNativeLCI}{[$+$0.016, $+$0.032]\xspace}
\newcommand{\nCalNativeR}{$+$0.022\xspace}
\newcommand{\nCalNativeRCI}{[$+$0.014, $+$0.032]\xspace}
\newcommand{\nXvFrames}{13\xspace}
\newcommand{\nXvStride}{4\xspace}
\newcommand{\nXvConf}{0.3\xspace}
\newcommand{\nXvOwl}{0.15\xspace}
\newcommand{\nXvMinPatches}{4\xspace}
\newcommand{\nXvMinFrames}{3\xspace}
\newcommand{\nWildRowsNan}{0.1\%\xspace}
\newcommand{\nXvRoi}{2.5\xspace}
\newcommand{\nRoiSmall}{1.5\xspace}
\newcommand{\nRoiLarge}{4\xspace}
\newcommand{\nCeilNA}{2\xspace}
\newcommand{\nCeilNB}{4\xspace}
\newcommand{\nCeilNC}{8\xspace}
\newcommand{\nCeilClips}{603\xspace}
\newcommand{\nCeilSeeds}{8\xspace}
\newcommand{\nCeilVarClips}{97\%\xspace}
\newcommand{\nCeilVarSeeds}{3\%\xspace}
\newcommand{\nCeilOracleTwo}{$-$0.0092\xspace}
\newcommand{\nCeilOracleFour}{$-$0.0163\xspace}
\newcommand{\nRoneDvsCAll}{$+$0.0012\xspace}
\newcommand{\nRoneDvsCAllCI}{[$-$0.0006, $+$0.0032]\xspace}
\newcommand{\nRoneDvsMAll}{$+$0.0025\xspace}
\newcommand{\nRoneDvsMAllCI}{[$+$0.0008, $+$0.0045]\xspace}
\newcommand{\nRoneAvsMAll}{$+$0.0028\xspace}
\newcommand{\nRoneAvsMAllCI}{[$-$0.0001, $+$0.0061]\xspace}
\newcommand{\nRsiPairOneDvsCAllK}{$-$0.8\xspace}
\newcommand{\nRsiPairOneDvsCAllKCI}{[$-$2.0, $+$0.4]\xspace}
\newcommand{\nRsiPairOneDvsMAllK}{$-$0.4\xspace}
\newcommand{\nRsiPairOneDvsMAllKCI}{[$-$1.3, $+$0.7]\xspace}
\newcommand{\nRsiPairTwoDvsCAllK}{$-$3.3\xspace}
\newcommand{\nRsiPairTwoDvsCAllKCI}{[$-$6.4, $-$0.7]\xspace}
\newcommand{\nRsiPairTwoDvsMAllK}{$-$1.9\xspace}
\newcommand{\nRsiPairTwoDvsMAllKCI}{[$-$4.0, $-$0.2]\xspace}
\newcommand{\nRsiLr}{$2{\times}10^{-5}$\xspace}
\newcommand{\nRsiWarm}{200\xspace}
\newcommand{\nRsiSteps}{4{,}000\xspace}
\newcommand{\nRsiDecayAt}{2{,}000\xspace}
\newcommand{\nRsiBatch}{30\xspace}
\newcommand{\nRsiSynShare}{25\%\xspace}
\newcommand{\nRsiOverlap}{3{,}542\xspace}
\newcommand{\nRoneLr}{$5{\times}10^{-5}$\xspace}
\newcommand{\nRoneSteps}{6{,}000\xspace}
\newcommand{\nRoneSynShare}{25\%\xspace}
\newcommand{\nProgCTwoAll}{0.519\xspace}
\newcommand{\nProgCTwoUnseen}{0.550\xspace}
\newcommand{\nProgCTwoEgoPro}{0.550\xspace}
\newcommand{\nProgCTwoEgoTouch}{0.427\xspace}
\newcommand{\nProgCTwoEgoViz}{0.563\xspace}
\newcommand{\nProgCTwoHumanTouch}{0.577\xspace}
\newcommand{\nProgCTwoWindows}{48{,}000\xspace}
\newcommand{\nProgCTwoLongAll}{0.426\xspace}
\newcommand{\nProgCTwoLongUnseen}{0.470\xspace}
\newcommand{\nProgCTwoLongEgoPro}{0.470\xspace}
\newcommand{\nProgCTwoLongEgoTouch}{0.323\xspace}
\newcommand{\nProgCTwoLongEgoViz}{0.531\xspace}
\newcommand{\nProgCTwoLongHumanTouch}{0.428\xspace}
\newcommand{\nProgCTwoLongWindows}{336{,}000\xspace}
\newcommand{\nProgBigAll}{0.417\xspace}
\newcommand{\nProgBigUnseen}{0.462\xspace}
\newcommand{\nProgBigEgoPro}{0.461\xspace}
\newcommand{\nProgBigEgoTouch}{0.315\xspace}
\newcommand{\nProgBigEgoViz}{0.527\xspace}
\newcommand{\nProgBigHumanTouch}{0.413\xspace}
\newcommand{\nProgBigWindows}{432{,}000\xspace}
\newcommand{\nProgLongAll}{0.382\xspace}
\newcommand{\nProgLongUnseen}{0.430\xspace}
\newcommand{\nProgLongEgoPro}{0.429\xspace}
\newcommand{\nProgLongEgoTouch}{0.279\xspace}
\newcommand{\nProgLongEgoViz}{0.500\xspace}
\newcommand{\nProgLongHumanTouch}{0.365\xspace}
\newcommand{\nProgLongWindows}{1{,}716{,}000\xspace}
\newcommand{\nProgConvAll}{0.374\xspace}
\newcommand{\nProgConvUnseen}{0.426\xspace}
\newcommand{\nProgConvEgoPro}{0.425\xspace}
\newcommand{\nProgConvEgoTouch}{0.264\xspace}
\newcommand{\nProgConvEgoViz}{0.496\xspace}
\newcommand{\nProgConvHumanTouch}{0.360\xspace}
\newcommand{\nProgConvWindows}{2{,}017{,}200\xspace}
\newcommand{\nRestLpOursCfgThreeAll}{0.351\xspace}
\newcommand{\nRestLpOursCfgThreeEgoPro}{0.417\xspace}
\newcommand{\nRestLpOursCfgThreeEgoTouch}{0.252\xspace}
\newcommand{\nRestLpOursCfgThreeEgoViz}{0.475\xspace}
\newcommand{\nRestLpOursCfgThreeHumanTouch}{0.356\xspace}
\newcommand{\nRestPsnrOursCfgThreeAll}{17.50\xspace}
\newcommand{\nRestLpOursCfgOneAll}{0.359\xspace}
\newcommand{\nRestLpOursCfgOneEgoPro}{0.422\xspace}
\newcommand{\nRestLpOursCfgOneEgoTouch}{0.264\xspace}
\newcommand{\nRestLpOursCfgOneEgoViz}{0.478\xspace}
\newcommand{\nRestLpOursCfgOneHumanTouch}{0.366\xspace}
\newcommand{\nRestPsnrOursCfgOneAll}{17.56\xspace}
\newcommand{\nRestLpWwAll}{0.504\xspace}
\newcommand{\nRestLpWwEgoPro}{0.542\xspace}
\newcommand{\nRestLpWwEgoTouch}{0.415\xspace}
\newcommand{\nRestLpWwEgoViz}{0.562\xspace}
\newcommand{\nRestLpWwHumanTouch}{0.544\xspace}
\newcommand{\nRestPsnrWwAll}{13.07\xspace}
\newcommand{\nRestLpOtherAll}{0.591\xspace}
\newcommand{\nRestLpOtherEgoPro}{0.648\xspace}
\newcommand{\nRestLpOtherEgoTouch}{0.463\xspace}
\newcommand{\nRestLpOtherEgoViz}{0.605\xspace}
\newcommand{\nRestLpOtherHumanTouch}{0.658\xspace}
\newcommand{\nRestPsnrOtherAll}{11.36\xspace}
\newcommand{\nRestLpCropAll}{0.769\xspace}
\newcommand{\nRestLpCropEgoPro}{0.744\xspace}
\newcommand{\nRestLpCropEgoTouch}{0.812\xspace}
\newcommand{\nRestLpCropEgoViz}{0.779\xspace}
\newcommand{\nRestLpCropHumanTouch}{0.752\xspace}
\newcommand{\nRestPsnrCropAll}{8.37\xspace}
\newcommand{\nRestLpBaseAll}{0.918\xspace}
\newcommand{\nRestLpBaseEgoPro}{0.839\xspace}
\newcommand{\nRestLpBaseEgoTouch}{0.998\xspace}
\newcommand{\nRestLpBaseEgoViz}{0.842\xspace}
\newcommand{\nRestLpBaseHumanTouch}{0.937\xspace}
\newcommand{\nRestPsnrBaseAll}{8.66\xspace}
\newcommand{\nRestDiffOursVsWwAll}{$-$0.153\xspace}
\newcommand{\nRestDiffOursVsWwAllCI}{[$-$0.162, $-$0.145]\xspace}
\newcommand{\nRestDiffOursVsWwAllBetter}{99\%\xspace}
\newcommand{\nRestDiffOursVsOtherAllBetter}{99\%\xspace}
\newcommand{\nRestDiffOursVsCropAllBetter}{100\%\xspace}
\newcommand{\nRestDiffOursVsBaseAllBetter}{100\%\xspace}
\newcommand{\nRestN}{257\xspace}
\newcommand{\nRestNEgoPro}{86\xspace}
\newcommand{\nRestNEgoTouch}{80\xspace}
\newcommand{\nRestNEgoViz}{15\xspace}
\newcommand{\nRestNHumanTouch}{76\xspace}
\newcommand{\nRestRelOursVsWw}{$-$30\%\xspace}
\newcommand{\nMonDiffOursVsWwAll}{$-$0.126\xspace}
\newcommand{\nMonDiffOursVsWwAllCI}{[$-$0.140, $-$0.112]\xspace}
\newcommand{\nLongRawAll}{0.379\xspace}
\newcommand{\nLongEmaAll}{0.382\xspace}

\definecolor{cvprblue}{rgb}{0.21,0.49,0.74}
\usepackage[pagebackref,breaklinks,colorlinks,allcolors=cvprblue]{hyperref}
\def\confName{CVPR}
\def\confYear{2027}

\title{Ego2Wrist: Bootstrapping Wrist-View Synthesis from Egocentric Video}
\author{%
Ziyao Zeng$^{1}$\quad
Xiatao Sun$^{1}$\quad
Hao Wang$^{2}$\quad
Ziyu Guo$^{2}$\quad
Youheng Yao$^{1}$\quad
Jingcheng Ni$^{3}$\\
Shenyuan Gao$^{4}$\quad
Fengyu Yang$^{1}$\quad
Renrui Zhang$^{5}$\quad
Bangya Liu$^{6}$\quad
Daniel Rakita$^{1}$\\[0.6em]
{\normalsize
$^{1}$Yale University\quad
$^{2}$Google DeepMind\quad
$^{3}$Brown University\quad
$^{4}$NVIDIA\quad
$^{5}$Independent Researcher\quad
$^{6}$2077 AI}\\[0.4em]
{\normalsize Project page: \url{https://adonis-galaxy.github.io/Ego2Wrist/}}
}

\begin{document}
\maketitle
\begin{strip}
\centering
\includegraphics[width=\textwidth]{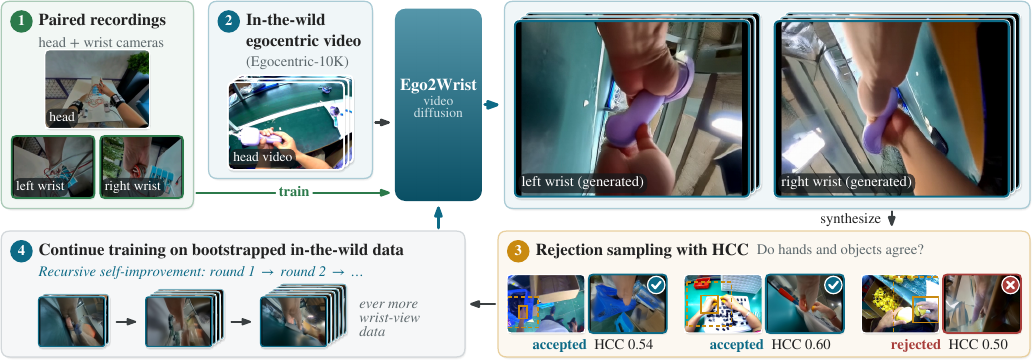}
\captionsetup{hypcap=false}
\captionof{figure}{\textbf{\method turns head-camera video into wrist-camera video and improves itself on video that has no wrist cameras.}
(1)~The model is trained on recordings with head and wrist cameras.
(2)~It then generates both wrist videos for in-the-wild egocentric video from Egocentric-10K~\cite{buildai2025egocentric10k}, which has only a head camera.
(3)~HCC compares each generated sample with the region around the hand in the head view (box) and rejects samples that disagree.
(4)~Training continues on the accepted samples, and the improved model generates the next round.}
\captionsetup{hypcap=true}
\label{fig:teaser}
\vspace{2mm}
\end{strip}

\begin{abstract}
Wrist cameras provide an action-centric view that is directly relevant to robot manipulation.
However, collecting human wrist-camera videos at scale is challenging due to cumbersome setups that can interfere with natural hand movements.
In contrast, head-mounted human egocentric videos are much easier to scale.
We introduce \method, which synthesizes synchronized wrist-camera views from human egocentric videos and task descriptions.
Trained on approximately \nHoursTotal hours of paired head- and wrist-camera recordings, \method jointly generates three views.
To evaluate generated wrist views without ground truth, we propose hand-centric cross-view consistency (HCC), which measures consistency between generated wrist views and corresponding hand regions in the input head view.
HCC ranks model checkpoints comparably to ground-truth-based LPIPS, and human evaluators prefer HCC-selected generations in \nStudyHTwoPref of responses.
Evaluated on four datasets, \method outperforms fine-tuned WorldWander, reducing LPIPS by 28\% and receiving human preference in \nStudyHOneAPref of responses.
More importantly, HCC facilitates recursive self-improvement on in-the-wild egocentric videos by selecting generated wrist views for further training.
Generation quality progressively improves over three rounds.
These results demonstrate a scalable approach to generating real-world wrist-view human manipulation videos for robot policy training.
\end{abstract}

\section{Introduction}
\label{sec:intro}

Wrist cameras provide an action-centric view that is directly relevant to robot manipulation~\cite{zhao2023learning,khazatsky2024droid,chi2024universal}.
However, collecting human wrist-camera videos at scale is challenging, as the camera setup is cumbersome and can interfere with natural hand movements.
In contrast, head-mounted human egocentric videos are much easier to scale~\cite{grauman2022ego4d,grauman2024egoexo4d,buildai2025egocentric10k}.
Synthesizing wrist views from these videos could therefore transform scalable real-world human demonstrations into action-centric data for robot policy training.

Paired recordings of head and wrist cameras were rare until 2026, when several datasets released about a thousand hours of synchronized head and wrist video of people performing manipulation tasks~\cite{lightwheel2026egosuite,zhou2026touchanything,humaid2026egoviz,lyu2026humantouch,chen2026spark0}.
This makes it possible to learn the mapping from a head video to both wrist videos directly (head-to-wrist synthesis, \cref{fig:teaser}).

Head-to-wrist synthesis presents two key challenges beyond conventional novel-view synthesis.
\textbf{First,} wrist cameras move with the forearms, but their trajectories cannot be directly inferred from head views.
Existing methods either require known target camera trajectories~\cite{fu2026plenoptic} or estimate them from static views~\cite{qian2025wristworld}, neither of which is readily available for egocentric human videos.
Even with estimated wrist poses, depth-based reprojection provides only marginal improvements (\cref{sec:ablation}).
We therefore train a video diffusion model that jointly generates both wrist views alongside the clean head view in a shared latent panel, allowing attention to establish cross-view relationships without explicit geometric modeling. 
\textbf{Second,} in-the-wild egocentric videos lack wrist-view ground truth, making reference-based evaluation impossible.
To address this, we propose hand-centric cross-view consistency (HCC), a ground-truth-free metric that measures whether generated wrist views are consistent with the corresponding hand regions in the input head view.

Beyond evaluation, HCC facilitates recursive self-improvement using in-the-wild human egocentric videos without wrist-view ground truth.
Specifically, we synthesize wrist views for \nRsiPool egocentric video windows and select the top 30\% based on HCC scores for further training alongside paired data.
We iteratively repeat this process with the updated model, progressively improving generation quality through self-generated training data.
In summary, our contributions are:
\begin{itemize}
    \item \textbf{We introduce \method, which synthesizes both wrist-camera views from human egocentric videos and task descriptions.}
    Evaluated on four datasets, \method generates wrist views consistent with the input head video and outperforms fine-tuned WorldWander~\cite{song2026worldwander}, reducing LPIPS by 28\% and achieving \nStudyHOneAPref human preference.

    \item \textbf{We propose hand-centric cross-view consistency (HCC), a ground-truth-free metric for evaluating generated wrist views.}
    HCC agrees with human judgments comparably to ground-truth-based LPIPS and provides a practical criterion for selecting synthetic training data.

    \item \textbf{We demonstrate recursive self-improvement using in-the-wild human egocentric videos.}
    By iteratively generating and selecting wrist views with HCC, \method progressively improves over three rounds of self-training.
    HCC-based selection also outperforms random selection, demonstrating the potential to scale wrist-view training data from real-world human demonstrations.
\end{itemize}

\section{Related Work}
\label{sec:related}

\paragraph{Cross-view video synthesis.}
Exocentric-to-egocentric methods render the first-person view of a person seen by third-person cameras, from several exocentric views~\cite{liu2024exo2egov}, with future prediction~\cite{xu2025egoexogen}, by point-cloud reprojection~\cite{park2026egoworld}, or by adapting pretrained video diffusion models~\cite{kang2026egox,mahdi2025exo2egosyn}; WorldWander~\cite{song2026worldwander} translates in both directions with in-context conditioning.
None of these works targets a wrist camera.
For robots, WristWorld~\cite{qian2025wristworld} generates one wrist view from two static third-person views through a predicted wrist pose, WARPED~\cite{freeman2026warped} renders a robot wrist view from egocentric human video with Gaussian splatting, and PlenopticDreamer~\cite{fu2026plenoptic} re-renders a head-camera video into both gripper views given their camera trajectories.
OpenLongTail~\cite{liu2026openlongtail} completes the surround cameras of a driving rig from the front view, and camera-controlled video models~\cite{vanhoorick2024generative,bai2025recammaster,yu2025trajectorycrafter,ren2025gen3c} likewise follow a given camera path.
We generate both wrist views of a person from a moving head camera, and the wrist poses are not given at inference.

\paragraph{Camera-controllable video generation.}
Camera poses condition video generation directly~\cite{wang2024motionctrl} or as per-pixel Pl\"ucker embeddings~\cite{he2025cameractrl,sitzmann2021light}, also in video diffusion transformers~\cite{bahmani2025vd3d,bahmani2025ac3d}.
Most of this work assumes a pinhole camera; AKiRa~\cite{wang2025akira}, UCPE~\cite{zhang2026unified} and CRePE~\cite{jin2026crepe} use ray-based representations that include lens distortion.
Our ray condition specifies the lens and mount of a target wrist camera, and our virtual-camera augmentation, close to that of AKiRa, lets one model serve wrist lenses from circular fisheye to rectilinear.

\paragraph{Evaluating generated video without ground truth.}
FVD~\cite{unterthiner2018towards} compares distributions of videos, VBench~\cite{huang2024vbench} and WorldScore~\cite{duan2025worldscore} score predefined dimensions, CrashTwin~\cite{chen2026physics} tests whether generated multi-agent dynamics obey physical laws, and VideoScore~\cite{he2024videoscore} and WorldModelBench~\cite{li2025worldmodelbench} train vision-language models on human ratings.
LPIPS~\cite{zhang2018unreasonable} agrees well with human similarity judgments but needs a reference.
MEt3R~\cite{asim2025met3r} measures multi-view consistency without a reference by warping one view into another with DUSt3R~\cite{wang2024dust3r}; dense matchers such as MASt3R~\cite{leroy2024grounding}, DINO features~\cite{caron2021emerging,oquab2024dinov2,simeoni2025dinov3} and CLIP~\cite{radford2021learning} also relate views.
HCC asks a narrower question, whether a generated wrist view shows the content of the head view around the same-side hand, and we compare it with MASt3R, CLIP and MEt3R.

\paragraph{Learning manipulation from human and synthetic video.}
A hand-centric camera improves training efficiency and out-of-distribution generalization~\cite{hsu2022vision}, and ALOHA~\cite{zhao2023learning}, DROID~\cite{khazatsky2024droid} and UMI~\cite{chi2024universal} record wrist cameras.
Human egocentric video is easier to collect at scale~\cite{grauman2022ego4d,grauman2024egoexo4d,buildai2025egocentric10k}, sometimes together with touch~\cite{li2026touchscale}, and it is used to pretrain representations~\cite{nair2023r3m} and to learn manipulation with 3D hand tracking~\cite{kareer2025egomimic,hoque2026egodex}.
Phantom~\cite{lepert2025phantom}, Masquerade~\cite{lepert2025masquerade} and DexUMI~\cite{xu2025dexumi} replace the human arm or hand by a robot within the recorded view, whereas we synthesize views that were not recorded.
Video generation also supplies robot data, for planning~\cite{du2023learning}, as an interactive simulator~\cite{yang2024learning}, as synthetic training videos~\cite{jang2025dreamgen}, and with tactile conditioning~\cite{zeng2026dtwm}.
Self-training methods keep the model samples that pass a reward or a verifier and fine-tune on them~\cite{zelikman2022star,gulcehre2023reinforced,singh2024beyond,dong2023raft}, and large image-text datasets are filtered with a learned score~\cite{gadre2023datacomp}; HCC plays this role for synthesized wrist views.
Training a generative model on its own samples can degrade it unless enough real data enters every round~\cite{shumailov2024ai,alemohammad2024self,bertrand2024stability}; our loop keeps the full real stream in every round and adds the bootstrapped windows at about one in four.

\section{Method}
\label{sec:method}
\begin{figure*}[t]
\centering
\includegraphics[width=\textwidth]{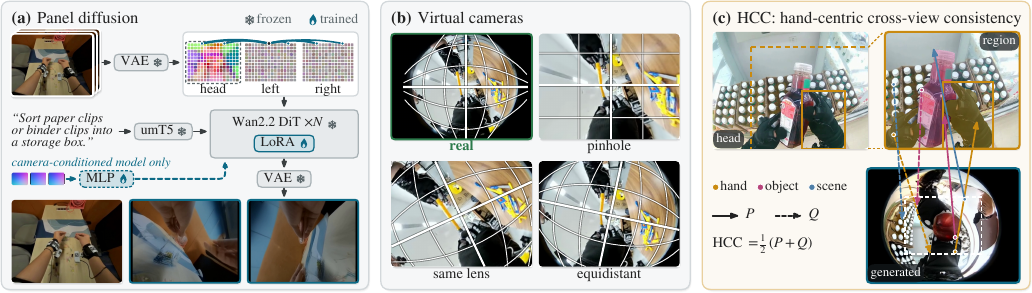}
\caption{\textbf{\method generates the two wrist views next to the head view in one panel, and HCC checks the result against the head view.}
(a)~The head video and the two wrist videos are encoded and placed side by side. A frozen video diffusion transformer with LoRA adapters denoises the wrist parts while the head part stays clean, so the wrist views can attend to the head view. Dashed: the camera branch of the camera-conditioned model.
(b)~Wrist cameras have different configurations, so training frames are re-rendered into random virtual cameras. The grid shows how each lens bends the same directions.
(c)~HCC compares image patches of the generated wrist frame with the head-view region: the area around the same-side hand in the head frame. $P$ (solid arrows): does each patch at the center of the wrist view have a similar patch in the head-view region? It is low when the wrist view shows content absent from the head view. $Q$ (dashed arrows): does each patch of the head-view region have a similar patch in the wrist view? It is low when the wrist view misses the hand or the object.}
\label{fig:method}
\end{figure*}

\subsection{Problem setup}
\label{sec:setup}

Given a head-camera video $\mathbf{H}$ of $T{=}49$ frames and a one-line description $c$ of the task, we generate the two wrist-camera videos $\mathbf{W}_l$ and $\mathbf{W}_r$ of the same interval, each view at $384{\times}512$ pixels.
Optionally, the user also specifies the target wrist cameras $\kappa_l$ and $\kappa_r$, each given by a lens model, a field of view, and a mounting pose.
Training uses paired recordings in which the head and wrist cameras run synchronously (\cref{sec:exp_setup}).
At test time only $\mathbf{H}$ and $c$ are available, and the wrist cameras are not observed.

A wrist camera is strapped to the forearm, so its pose is not visible from the head, and explicit reprojection helps little (\cref{sec:ablation}).
We therefore learn the mapping end to end, with the head video, both wrist views and the text in one attention context.

\subsection{Panel diffusion}
\label{sec:panel}

\paragraph{Latent panel.}
We build on Wan2.2-Fun-5B-InP\footnote{\url{https://huggingface.co/alibaba-pai/Wan2.2-Fun-5B-InP}}, the inpainting variant of the Wan latent video diffusion model~\cite{wan2025wan}, trained with flow matching~\cite{lipman2023flow,liu2023flow} and conditioned on a masked reference video.
Its VAE compresses each view by $16{\times}$ in space and $4{\times}$ in time.
We encode the three views separately and concatenate the latents along the width into a panel
\begin{equation}
  \mathbf{Z} = \left[\, \mathbf{z}_h \mid \mathbf{z}_l \mid \mathbf{z}_r \,\right], \qquad \mathbf{z}_v = \mathcal{E}(\mathbf{V}_v),
\end{equation}
which the transformer processes as one token sequence.
Every wrist token can therefore attend to every head token and to the other wrist, which lets the model carry content from the head view into each wrist view without an explicit geometric model and keeps the two wrists consistent.

\paragraph{Conditioning.}
The head latent fills the head columns of the noisy input, with their timestep set to zero, and of the inpainting reference, whose mask marks them as known; the wrist columns of the reference hold a grey video marked as unknown.
The task description is encoded by umT5~\cite{chung2023unimax} and enters through cross-attention.
Language can resolve ambiguities that a single view leaves open~\cite{zeng2024wordepth,zeng2024rsa,zeng2026iris}; here it tells the model which object and action to render where the head view is uninformative.
We fine-tune the transformer with LoRA~\cite{hu2022lora} of rank 64.

\paragraph{Training.}
For a panel $\mathbf{Z}$, we draw a noise level $\sigma$, noise $\boldsymbol{\epsilon}$, and form $\mathbf{x}_\sigma = (1-\sigma)\mathbf{Z} + \sigma\boldsymbol{\epsilon}$ in the wrist columns while the head columns stay clean.
The network $v_\theta$ predicts the velocity, and the loss is taken over the wrist columns only:
\begin{equation}
  \mathcal{L} = \mathbb{E}_{\mathbf{Z},\sigma,\boldsymbol{\epsilon}} \left\| \left(v_\theta(\mathbf{x}_\sigma, \sigma, \mathbf{y}, c) - (\boldsymbol{\epsilon} - \mathbf{Z})\right)_{\text{wrist}} \right\|_2^2 ,
\end{equation}
where $\mathbf{y}$ is the inpainting reference.
With probability 0.1 each, we drop the head video, replacing it by the grey latent in both the input and the reference, and the text.

\paragraph{Head guidance.}
Dropping the head during training gives an unconditional branch $v_\varnothing$.
At inference we apply classifier-free guidance~\cite{ho2022classifier} to the head condition only, $\hat{v} = v_\varnothing + w\,(v_h - v_\varnothing)$, where both branches see the same text.
Without guidance ($w{=}1$) the samples are blurrier than real wrist video, because the model averages over the many wrist views that are compatible with the head view.
We select $w{=}3$ from $\{1,3,5\}$ on the validation set (\cref{sec:ablation}) and sample with 30 Euler steps.

\subsection{Camera-conditioned generation}
\label{sec:camera}

The four datasets use different wrist cameras, from circular fisheye lenses to pinhole views, and the wrist cameras of robots differ again.
A model trained on this mixture learns the camera from the data source and cannot be told which camera to render, so we make the target camera an input.

\paragraph{Ray conditioning.}
As in recent lens-aware camera encodings~\cite{wang2025akira,zhang2026unified,jin2026crepe}, we describe a camera by its viewing rays rather than by a pinhole matrix.
Each latent token receives the ray directions of its footprint, a validity flag for pixels outside the image circle, and a code of the mounting position.
An MLP and zero-initialized linear layers add this condition to the hidden states of every transformer block, so the model starts identical to the unconditioned one.
Lens models come from AnyCalib~\cite{tiradogarin2025anycalib} (\cref{sec:s_calib}).

\paragraph{Virtual-camera augmentation.}
Training data cover only a few cameras, so we synthesize others: each view of a training sample is re-rendered into a random virtual camera that changes the lens, the field of view and the orientation (\cref{fig:method}b).
The image and its ray map are rendered with the same camera model, so the image always agrees with its condition.
We remove the sentence that describes the recording device from the text, so that the camera is not entangled with the prompt~\cite{zeng2025coffee}.
Ranges and probabilities are in \cref{tab:s_virtual}.

\subsection{Hand-centric cross-view consistency (HCC)}
\label{sec:score}

On in-the-wild video there is no real wrist video, so reference metrics such as LPIPS~\cite{zhang2018unreasonable} cannot be computed.
We instead compare the generated wrist view with the part of the head view that the wrist camera observes: a region around the same-side hand, which shows the hand, the object it manipulates and the nearby scene.
\emph{Hand-centric cross-view consistency} (HCC) measures whether the generated wrist view and this region show the same content.

For each of 13 frames ($t = 0, 4, \dots, 48$), we detect the same-side hand in the head frame with the WiLoR detector~\cite{potamias2025wilor}, falling back to OWLv2~\cite{minderer2023scaling} for gloved hands, and take a square region $R_t$ centered on the hand whose side is 2.5 times the longer side of the box.
We extract DINOv3~\cite{simeoni2025dinov3} patch features from the head frame and from the generated wrist frame, giving unit vectors $\mathbf{h}_j$ and $\mathbf{w}_i$.
Precision measures whether the center of the wrist view appears in the region, and recall measures whether the region appears anywhere in the wrist view:
\begin{equation}
  P_t = \frac{1}{|C|} \sum_{i \in C} \max_{j \in R_t} \mathbf{w}_i^{\top}\mathbf{h}_j, \qquad
  Q_t = \frac{1}{|R_t|} \sum_{j \in R_t} \max_{i} \mathbf{w}_i^{\top}\mathbf{h}_j ,
\end{equation}
where $C$ is the central half of the wrist frame in each direction.
HCC of a wrist is the mean of $\tfrac{1}{2}(P_t + Q_t)$ over the frames in which its hand is detected, and HCC of a sample is the mean over the two wrists.
Each direction takes the nearest neighbor of every patch, as BERTScore~\cite{zhang2020bertscore} does for tokens, rather than one-to-one correspondences, because the two views differ strongly in scale and overlap only partly.
Verified correspondences (MASt3R inliers~\cite{leroy2024grounding}) agree less with human judgments; the cost of our choice is that HCC checks what the wrist view shows but not where it shows it (\cref{sec:s_score_design}).
HCC does not use the wrist ground truth, the camera parameters, or any learned component trained on our data.

\subsection{Recursive self-improvement with bootstrapped data}
\label{sec:select}

HCC also lets the model learn from unlabeled egocentric video.
We cut head-only videos into windows, keep those in which the wearer handles an object with both hands visible, caption them with a vision-language model, and let the current model generate both wrist videos for each window.
We call these samples \emph{bootstrapped} data.
HCC then serves for rejection sampling: we accept the samples in the top 30\% of HCC and reject the rest.
The accepted samples are added to the paired training stream, so that every training step sees the same real data as before plus a share of bootstrapped data, and training continues from the current model.
Repeating these steps with the updated model and new unlabeled video is recursive self-improvement (\cref{fig:teaser}).
One sample is generated per window, so the selection acts across inputs; we do not use HCC to choose among samples of the same input, where it is not reliable (\cref{sec:exp_score}).

\section{Experiments}
\label{sec:experiments}

\subsection{Setup}
\label{sec:exp_setup}

\paragraph{Data.}
We train on four datasets of synchronized head and wrist video of people performing manipulation tasks, \nHoursTotal hours in total: EgoPro~\cite{lightwheel2026egosuite} (\nHoursEgoPro hours, pinhole wrist cameras), EgoTouch~\cite{zhou2026touchanything} (\nHoursEgoTouch, circular fisheye), EgoViz~\cite{humaid2026egoviz} (\nHoursEgoViz, wide angle) and HumanTouch~\cite{lyu2026humantouch,chen2026spark0} (\nHoursHumanTouch, wide angle).
Splits keep whole recording units apart (\cref{sec:s_splits}).
The standard test set has \nTestClips clips of 49 frames from the four datasets; \nTestClipsUnseen of its EgoPro clips come from environments that never appear in training.
A separate validation set of \nCfgValN clips is used to choose the guidance weight.

\paragraph{Decisions made on test clips.}
Three decisions about our own training schedule, none involving a baseline, used test clips: when to switch to the full resolution and when to start the final learning-rate decay were decided on \nSetCompact clips from the test splits, \nSetCompactOverlap of them in the standard test set, and the final stage starts from the raw rather than the averaged long-run weights, which were slightly better on the standard test set (\nLongRawAll against \nLongEmaAll, far less than the gap to any baseline).
On the \nRestN clips that were never monitored (\cref{tab:main}, \emph{Unmon.}), the gap to WorldWander, \nRestDiffOursVsWwAll \nRestDiffOursVsWwAllCI, is slightly larger than on all \nTestClips.

\paragraph{Metrics.}
Our main metric is LPIPS~\cite{zhang2018unreasonable} (AlexNet features) between the generated and the real wrist video, averaged over frames and both wrists.
When people compare two checkpoints of our model (\cref{sec:exp_score}), LPIPS agrees with them about as well as two people agree with each other.
We also report layout PSNR, the PSNR after shrinking each frame by 16 in each direction, which measures whether the coarse layout is right.
All comparisons are paired by clip: we report the mean difference with a 95\% confidence interval from 10{,}000 bootstrap resamples of clips, and the share of clips on which the first method is better.

\paragraph{Baselines.}
We compare against four baselines.
\emph{Untuned Wan2.2} is the base model with our conditioning but without fine-tuning.
\emph{Head-view crop} crops the head video around the same-side hand; it tests whether a wrist view is more than a zoomed head view.
\emph{Another clip's wrist video} is the real wrist video of a different clip from the same dataset: realistic, but with the wrong content.
\emph{WorldWander}~\cite{song2026worldwander} translates videos between egocentric and exocentric viewpoints with a video diffusion model of the same backbone family and without camera poses, which makes it the closest prior method that can be trained on our data.
We fine-tune it on our training data with its public recipe and sample it with its own settings (\cref{sec:s_ww}).

\paragraph{Implementation.}
We fine-tune LoRA weights of rank 64 on 49 frames at $384{\times}512$ per view and evaluate an EMA of the weights; the training schedule and all hyperparameters are in \cref{sec:s_impl}.

\begin{table*}[t]
\centering
\small
\setlength{\tabcolsep}{3.8pt}
\begin{tabular}{@{}lccccccccc@{}}
\toprule
& \multicolumn{6}{c}{LPIPS $\downarrow$} & & \multicolumn{2}{c}{All clips} \\
\cmidrule(lr){2-7}\cmidrule(l){9-10}
Method & EgoPro & EgoTouch & EgoViz & HumanTouch & All & Unmon. & & Layout PSNR $\uparrow$ & Ours better \\
\midrule
Untuned Wan2.2~\cite{wan2025wan} & \nLpBaseEgoPro & \nLpBaseEgoTouch & \nLpBaseEgoViz & \nLpBaseHumanTouch & \nLpBaseAll & \nRestLpBaseAll & & \nPsnrBaseAll & \nDiffOursVsBaseAllBetter \\
Head-view crop & \nLpCropEgoPro & \nLpCropEgoTouch & \nLpCropEgoViz & \nLpCropHumanTouch & \nLpCropAll & \nRestLpCropAll & & \nPsnrCropAll & \nDiffOursVsCropAllBetter \\
Another clip's wrist video & \nLpOtherEgoPro & \nLpOtherEgoTouch & \nLpOtherEgoViz & \nLpOtherHumanTouch & \nLpOtherAll & \nRestLpOtherAll & & \nPsnrOtherAll & \nDiffOursVsOtherAllBetter \\
WorldWander~\cite{song2026worldwander}, fine-tuned & \nLpWwEgoPro & \nLpWwEgoTouch & \nLpWwEgoViz & \nLpWwHumanTouch & \nLpWwAll & \nRestLpWwAll & & \nPsnrWwAll & \nDiffOursVsWwAllBetter \\
\midrule
\method, $w{=}1$ & \nLpOursCfgOneEgoPro & \nLpOursCfgOneEgoTouch & \nLpOursCfgOneEgoViz & \nLpOursCfgOneHumanTouch & \nLpOursCfgOneAll & \nRestLpOursCfgOneAll & & \textbf{\nPsnrOursCfgOneAll} & \\
\method, $w{=}3$ & \textbf{\nLpOursCfgThreeEgoPro} & \textbf{\nLpOursCfgThreeEgoTouch} & \textbf{\nLpOursCfgThreeEgoViz} & \textbf{\nLpOursCfgThreeHumanTouch} & \textbf{\nLpOursCfgThreeAll} & \textbf{\nRestLpOursCfgThreeAll} & & \nPsnrOursCfgThreeAll & \\
\bottomrule
\end{tabular}
\caption{\textbf{\method has the lowest LPIPS on every dataset} (standard test set, \nTestClips clips). LPIPS compares the generated with the real wrist videos (lower is better), and layout PSNR measures the coarse layout (higher is better). ``Ours better'' is the share of clips on which \method ($w{=}3$) beats the method of that row. In a human study with the real wrist videos shown, people prefer \method over WorldWander in \nStudyHOneAPref of answers. $w$ is the head-guidance weight. \emph{Unmon.}: the \nRestN clips never monitored during training (\cref{sec:exp_setup}).}
\label{tab:main}
\end{table*}

\subsection{Comparison with baselines}
\label{sec:exp_main}
\begin{figure*}[t]
\centering
\includegraphics[width=\textwidth]{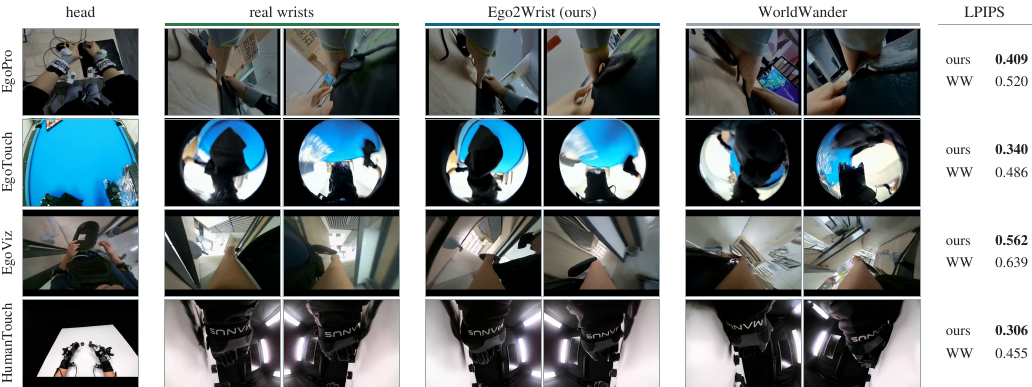}
\caption{\textbf{\method keeps the viewpoint and layout of the real wrist view, while WorldWander often shows the hand and object from a wrong viewpoint.}
Each row is one test clip: the head input, then the left and right wrist views from the real cameras, from \method ($w{=}3$) and from WorldWander fine-tuned on our data. Numbers on the right are LPIPS (lower is better).
The rows are typical cases: for each dataset we show the clip with the median LPIPS gap between the two methods, among \nXfamN test clips drawn without looking at any score.}
\label{fig:qual}
\end{figure*}

In \cref{tab:main}, \method reaches an LPIPS of \nLpOursCfgThreeAll, and its output is specific to the input clip: the real wrist video of another clip is fully realistic, yet ours is better by \nDiffOursVsOtherAll \nDiffOursVsOtherAllCI.
On the \nTestClipsUnseen EgoPro clips from unseen environments, LPIPS is \nLpOursCfgThreeUnseen, as on all EgoPro clips (\nLpOursCfgThreeEgoPro).

\paragraph{The model uses the head view.}
A video model could produce plausible wrist views from the text and the dataset prior alone, so we replace the head video by that of another clip from the same dataset while keeping the text.
With $w{=}1$, LPIPS rises from \nHeadSwapLpRealAll to \nHeadSwapLpShuffleAll (paired difference \nHeadSwapAll \nHeadSwapAllCI, correct head video better on \nHeadSwapAllBetter of the clips), and the gap is significant on each dataset.

\paragraph{Comparison with WorldWander.}
Against WorldWander, the paired LPIPS difference is \nDiffOursVsWwAll \nDiffOursVsWwAllCI, a relative change of \nRelOursVsWw, and lower on \nDiffOursVsWwAllBetter of the clips; without fine-tuning, WorldWander reaches \nLpWwZsAll on the \nLpWwZsN clips we ran it on, where \method reaches \nLpOursOnWwZsClips.
The gap holds on every dataset, from \nDiffOursVsWwEgoViz on EgoViz to \nDiffOursVsWwHumanTouch on HumanTouch, and layout PSNR is higher by \nDiffOursVsWwAllPsnr\,dB.
WorldWander often produces a plausible hand and object from the wrong viewpoint, while our wrist views keep the layout of the real ones (\cref{fig:qual}).
In a human study, \nStudyRaters raters see the head video and the real wrist videos of \nStudyHOneAItems test clips and choose which of two unnamed outputs, ours or WorldWander's, is closer to the real one.
They prefer ours in \nStudyHOneAPref \nStudyHOneAPrefCI of their answers; the preference holds on EgoTouch (\nStudyHOneAEgoTouch), EgoViz (\nStudyHOneAEgoViz) and HumanTouch (\nStudyHOneAHumanTouch) and is not significant on EgoPro (\nStudyHOneAEgoPro \nStudyHOneAEgoProCI).
This comparison, however, is not at matched compute: WorldWander is fine-tuned from its released weights on \nWwWindows windows and \method from the base model on about \nTrainWindows (\cref{sec:s_train,sec:s_ww}).
With fewer windows (\nProgCTwoWindows) and \nHoursVTwo of the \nHoursTotal hours of data, our first stage was already close to WorldWander without guidance (\nProgCTwoAll against \nLpWwAll).
This comparison thus does not separate architecture from training budget; the evidence above that the output follows the input clip and its head view does not depend on this baseline.

\subsection{Camera control}
\label{sec:exp_camera}

As a side capability, the camera-conditioned model of \cref{sec:camera} renders the wrist views in a requested camera (\cref{sec:s_camera}).
We request a narrower, a rotated or a fisheye camera and score against the real wrist video re-rendered into it.
The oracle generates in the native camera with \method and re-renders the output with the renderer that produced the reference.
The condition closes \nCamGapNarrow, \nCamGapRoll and \nCamGapFisheye of the gap between the model without it and the oracle, and the oracle stays better on each camera (oracle minus condition: \nWarpNarrowAll, \nWarpRollAll and \nWarpFisheyeAll).
On the native cameras the condition costs \nCamNativeAll, so we report \method without it in \cref{tab:main}.
For cameras inside the native view, generating in the native camera and re-rendering is therefore the better choice; the condition could matter for wider cameras, which re-rendering cannot produce and which we cannot evaluate without such recordings.

\begin{figure}[t]
\centering
\includegraphics[width=0.85\linewidth]{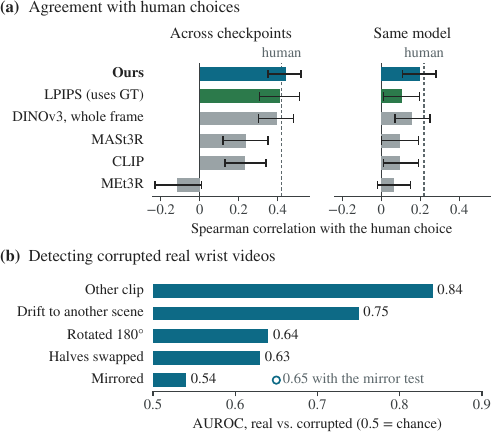}
\caption{\textbf{HCC agrees with people as well as LPIPS does, although it never sees the real wrist video.}
(a)~People chose the better of two generated videos, and the bars show how well each score predicts their choice (Spearman correlation, 95\% CIs). Left: the two videos come from different checkpoints (\nHumNCross answers). Right: they are two samples of the same checkpoint (\nHumNSame answers), where every score is weak. Ours: HCC. Dashed line: agreement between two people.
(b)~How often HCC ranks a real wrist video above a corrupted copy (0.5 is chance). It detects wrong content well and geometric changes such as mirroring less well. Open circle: with an explicit test for mirrored layouts.}
\label{fig:score}
\end{figure}

\subsection{Evaluating without ground truth}
\label{sec:exp_score}

\paragraph{Agreement with people.}
In a first study, people saw the real video of one wrist camera and two generated videos of it, from different checkpoints of our model or different samples of one checkpoint, and chose the one closer to the real video (\nHumScored answers from \nHumAnalysedRaters raters; \cref{sec:s_study}).
When two checkpoints are compared, HCC agrees with people as well as LPIPS does, although it does not see the real wrist video (\cref{fig:score}a, \nHumOursCross against \nHumLpipsCross), and it is above the other scores we tested that need no ground truth.
Matching against the region around the hand rather than the whole head frame raises the agreement from \nHumWholeCross to \nHumOursCross (overlapping intervals).
We chose the region size on these answers; all three sizes we tried stay at or above LPIPS (\nRoiCrossOneFive, \nRoiCrossTwoFive and \nRoiCrossFour for 1.5, 2.5 and 4 times the hand box).
On samples of the same model, agreement is lower for every score and between two people (HCC \nHumOursSame, LPIPS \nHumLpipsSame, people \nHumHumanSame).

\paragraph{A test on new data.}
HCC was designed after the answers above were collected, so a second study, collected after HCC was fixed, tests it on in-the-wild video.
On \nStudyHOneB Egocentric-10K windows, people compared our wrist videos with WorldWander's without wrist ground truth.
They prefer ours in \nStudyHOneBPref \nStudyHOneBPrefCI of their answers, a tie, and HCC agrees with their choice in \nStudyHOneBHcc \nStudyHOneBHccCI of the answers (\cref{sec:s_study}).

\paragraph{Corrupted videos.}
We corrupt real wrist videos and ask whether HCC prefers the original (\cref{fig:score}b).
HCC detects a wrist video from another clip (AUROC \nAucSwapOurs, against \nAucSwapMastr for MASt3R and \nAucSwapMetr for MEt3R) and a gradual drift to another scene (\nAucDriftOurs), and it is weaker on geometric edits that keep the content, such as rotation or mirroring: it measures whether the right content appears, not where it appears.

\paragraph{What HCC cannot do.}
HCC cannot choose among samples of one input: among eight samples, picking the one with the best HCC captures only \nCeilScoreCapture of the gain of picking by LPIPS, and a ranker trained on several GT-free features captures \nCeilRankerCapture.
HCC is also sensitive to blur.
On \nXfamN test clips not selected by any score, our unguided samples ($w{=}1$), which are blurrier than real video, score below WorldWander (\nXfamDiff \nXfamDiffCI), although LPIPS prefers them on \nXfamLpOursBetter clips.
With head guidance the samples are sharper and HCC prefers \method (\nXfamThreeDiff \nXfamThreeDiffCI, \nXfamThreeOursHigher of \nXfamN clips), in the same direction as LPIPS (\nXfamThreeLpOursBetter of \nXfamN).
Clip by clip, the two agree only weakly across methods: WorldWander's errors lie mostly in viewpoint and layout, which HCC does not check.

\paragraph{Where HCC can be used.}
HCC asks whether the content around the hand in the head view appears in the wrist view, and a copy of that region satisfies this best: the head-view crop of \cref{tab:main}, which LPIPS ranks last, scores \nCropHcc against \nCropHccReal for the real wrist video (\cref{sec:s_score_copy}).
As a reward, HCC would teach a model to copy the head view; as a metric to compare methods, it would rank a method that copies above the real video.
To select training data, it only ranks windows that one model has synthesized: the model is trained on the windows that HCC keeps, never on the score itself.
The selection does not favor copies either: the kept windows are slightly closer to their head-view crop than the dropped ones (LPIPS \nCopyOwnDiffPool \nCopyOwnDiffPoolCI), but by as much as they are closer to the crop of an unrelated window, and none of them is as close to its crop as the closest 5\% of real wrist videos (\cref{sec:s_score_copy}).
We therefore use HCC to rank checkpoints of one model and to select its training data, and compare methods with LPIPS and people.

\subsection{Recursive self-improvement}
\label{sec:exp_select}
\begin{figure*}[t]
\centering
\includegraphics[width=\textwidth]{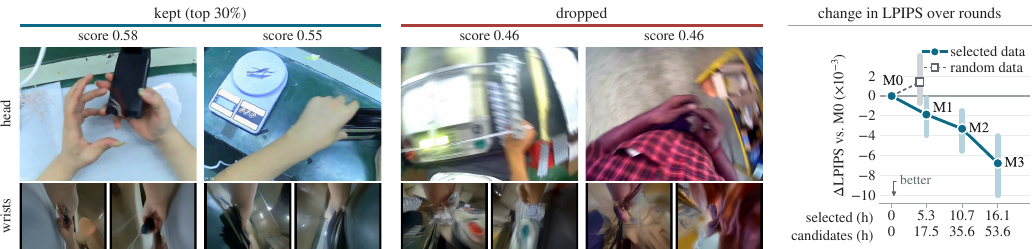}
\caption{\textbf{Training on its own HCC-selected wrist videos improves the model round after round.}
Left: in-the-wild windows with the head frame (top), the generated wrists (bottom) and their HCC. The top 30\% are kept for training (\nRsiKept of \nRsiPool) and the rest are dropped; shown are random draws from the kept and from the bottom 30\%.
Right: LPIPS on the standard test set relative to the starting model M0 after each round (lower is better, 95\% CIs). Teal: trained on HCC-selected data (M1 to M3). Grey: trained on the same amount of randomly chosen data in the first round; M1 and the random point are the seed pair that continues. The axis gives the cumulative hours of selected and of candidate video.}
\label{fig:select}
\end{figure*}

\paragraph{Does selection help?}
We generate wrist videos for \nRsiPool windows of in-the-wild egocentric video from Egocentric-10K~\cite{buildai2025egocentric10k}, which has no wrist cameras, and keep the \nRsiKept with the highest HCC (HCC $\geq$ \nRsiCut; \cref{fig:select}, left).
The kept windows come from \nWildFactoriesKept of the \nWildFactoriesPool factories in the pool, so HCC does not narrow the data to a few scenes.
Before scoring, we drop windows whose wrists show a circular-fisheye look that the target wrist camera never produces (\cref{sec:s_wild}).
In this first round we compare with a random control: the same number of windows drawn at random from the same pool.
Each arm continues training from \method (M0) for 4{,}000 steps with two seeds, with the bootstrapped samples added to the real stream at about one in four.
The decision rule was fixed before training: selection must beat random selection and the starting model.

Data selected by HCC is better than random data, by \nRsiDvsCAll \nRsiDvsCAllCI on the standard test set.
Random data leaves the model where it started (\nRsiCvsMAll \nRsiCvsMAllCI), while selected data gives a small gain (\nRsiDvsMAll \nRsiDvsMAllCI).
On the larger extended EgoPro set (\nRsiExtDvsCAllN clips), neither selected $-$ random (\nRsiExtDvsCAll \nRsiExtDvsCAllCI) nor selected $-$ start (\nRsiExtDvsMAll \nRsiExtDvsMAllCI) is significant (\cref{sec:s_select}).

\paragraph{Does the model keep improving?}
M1, the selected-data model of the seed with the lower validation LPIPS, synthesizes wrist videos for \nRsiTwoPool new windows, HCC keeps the top 30\%, and training continues for another 4{,}000 steps to give M2; a third round gives M3 in the same way (\nRsiThreePool new windows; \cref{sec:s_select}).
Each round trains on windows that earlier rounds did not use: the second round draws on workers absent from the first, and the third on new windows of the same videos.
The rule of each round, fixed before training, compares the new model with the previous one.
M2 improves on the extended EgoPro set (\nRsiTwoExtVsOneAll \nRsiTwoExtVsOneAllCI); on the standard set the difference has the same size, and its interval ends just below zero (\nRsiTwoVsOneAll \nRsiTwoVsOneAllCI).
M3 improves on M2 on both the standard set (\nRsiThreeVsTwoAll \nRsiThreeVsTwoAllCI) and the extended EgoPro set (\nRsiThreeExtVsTwoAll \nRsiThreeExtVsTwoAllCI).
Over three rounds, M3 is better than M0 on the standard set (\nRsiThreeVsZeroAll \nRsiThreeVsZeroAllCI; \cref{fig:select}, right) and on the extended EgoPro set (\nRsiThreeExtVsZeroAll \nRsiThreeExtVsZeroAllCI), and the third round gains no less than the second.
These gains are small in absolute terms, yet three rounds reach \nRsiVsCfgShare of the gain of head guidance (\cref{sec:ablation}) without its doubled sampling cost.

People agree with the selection.
In the second study, raters compare two windows of the same source video, one kept by HCC and the nearest one it drops.
They prefer the kept window in \nStudyHTwoPref \nStudyHTwoPrefCI of their answers (\cref{sec:s_study}).
Each window has a single sample, so this shows that HCC keeps windows whose synthesis people prefer, not that it would pick the better of two samples of one window.

\subsection{Ablations}
\label{sec:ablation}

\paragraph{Head guidance.}
On the validation set, the paired LPIPS difference of $w{=}3$ against $w{=}1$ is \nCfgValThreeAll \nCfgValThreeAllCI, while $w{=}5$ gives no reliable gain (\cref{tab:s_ablation}).
On the test set, run once with the chosen weight, the gain is \nCfgTestAll \nCfgTestAllCI and layout PSNR is unchanged (\nCfgTestAllPsnr\,dB).
Guidance mainly sharpens the output, and it doubles the sampling cost.

\paragraph{Explicit geometry.}
We also tried an explicit hint: the head frame reprojected into the wrist view with monocular depth and an estimated wrist pose.
With a wrist pose estimated from the real wrist frames, the hint helps by only \nWarpZeroAll, and the gain shrinks as the pose error grows and is no longer significant with 20$^\circ$ or 40$^\circ$ of pose error (\nWarpFortyAll at 40$^\circ$).
Adding zoomed crops of the hands from the head view as extra inputs makes the result slightly worse (\nZoomAll).
These are short runs against a control trained on the same schedule (\cref{sec:s_ablation}); the panel model already extracts what these hints provide.

\section{Limitations and conclusion}
\label{sec:conclusion}

Our comparison with WorldWander is not at matched compute (\cref{sec:exp_main}).
HCC agrees with people when ranking checkpoints of one model and when selecting training windows, but it penalizes blur more than errors of viewpoint and layout and cannot choose among samples of the same input, so comparisons between methods should rest on reference metrics and human judgments.
The gains of recursive self-improvement are small, and the first round shows no measurable change on the extended EgoPro set.
Finally, the generated wrist views carry less contact information than the head view they are generated from, and the timing of contact is less accurate than in real wrist video (\cref{sec:s_contact}); data that records contact or force~\cite{yang2024binding,zeng2026dtwm,li2026touchscale} could close this gap, and we have not yet trained a policy on synthesized wrist views.

We presented \method, which turns head-camera video into both wrist-camera videos, and HCC, a cross-view score that evaluates the result without wrist ground truth and selects bootstrapped training data.
The three rounds of self-improvement drew on only \nRsiCandHoursTotal hours of in-the-wild video from a corpus that holds far more, so larger pools and more rounds are a direct next step.
HCC uses neither wrist ground truth, camera parameters nor any component trained on our data, so the same loop applies to any head-camera corpus.
Together they are a step toward turning the large body of human egocentric video into the form that wrist-camera policies consume.

\section*{LLM Usage Statement}
A vision-language model, Gemini 4 Argon (Google), is part of the data pipeline: it screens the in-the-wild windows and writes their captions (\cref{sec:select,sec:s_wild}).
Beyond this, the authors used an LLM coding agent, Claude Code (Anthropic).
Under the authors' direction, it wrote much of the training, evaluation, analysis and data-processing code, and it scheduled, ran and monitored experiments and computed their statistics over many steps with limited step-by-step direction.
Claude was also used to draft and edit text, to summarize related work and to produce figures, and ChatGPT (OpenAI) was used to edit text; the authors reviewed and revised all of it.
The authors obtained reviewer-style feedback on the manuscript from Claude and from ChatGPT.
The research questions, claims and conclusions are the authors'; LLMs only helped to refine how they are worded.
Every number in the text is generated from the experiment logs, and the authors take full responsibility for all content.

{
    \small
    \bibliographystyle{ieeenat_fullname}
    \bibliography{refs}
}

\clearpage
\appendix

\renewcommand{\topfraction}{0.85}
\renewcommand{\bottomfraction}{0.5}
\renewcommand{\textfraction}{0.15}
\renewcommand{\floatpagefraction}{0.7}
\renewcommand{\dbltopfraction}{0.85}
\renewcommand{\dblfloatpagefraction}{0.7}
\setcounter{topnumber}{3}
\setcounter{dbltopnumber}{3}
\setcounter{totalnumber}{5}
\makeatletter\setlength{\@fptop}{0pt}\setlength{\@dblfptop}{0pt}\makeatother

\setcounter{figure}{0}
\setcounter{table}{0}
\setcounter{equation}{0}
\renewcommand{\thefigure}{S\arabic{figure}}
\renewcommand{\thetable}{S\arabic{table}}
\renewcommand{\theequation}{S\arabic{equation}}
\renewcommand{\theHfigure}{S\arabic{figure}}
\renewcommand{\theHtable}{S\arabic{table}}
\renewcommand{\theHequation}{S\arabic{equation}}

\twocolumn[
  \centering
  \Large
  \textbf{\thetitle}\\
  \vspace{0.5em}Appendix\\
  \vspace{1.0em}
]

\noindent This appendix gives the details that the main paper refers to.
Its sections are numbered with letters, and its figures and tables carry the prefix S.

\section{Implementation details}
\label{sec:s_impl}

\subsection{Training schedule}
\label{sec:s_train}

\method fine-tunes LoRA weights~\cite{hu2022lora} of rank \nLoraRank (alpha equal to the rank, no dropout) on all \nLoraModules linear layers of the Wan2.2-Fun-5B-InP transformer~\cite{wan2025wan}, \nLoraParams trainable parameters; the VAE and the text encoder stay frozen.
The noise level follows the flow-matching schedule of the base model with shift \nShift.
We use AdamW with $\beta=(\nAdamBetaOne, \nAdamBetaTwo)$, weight decay \nAdamWd, gradient clipping at \nStgGradClip, a linear warm-up, and one clip of \nFrames frames per GPU, with gradient accumulation to reach the batch sizes below.
The head video and the text are each dropped with probability \nStgSeedsDrop (\cref{sec:panel}).

The final model was trained in consecutive stages (\cref{tab:s_stages}), each initialized from the previous one.
The schedule grew with the project rather than being planned in advance: stages 1 to 3 used the data available at the time, EgoTouch, EgoViz, HumanTouch and the first part of EgoPro (\nHoursVTwo hours of training data), and the full \nHoursTotal hours became available during the long run, which switched to them after step \nStgLongDataSwap.
The long run trained at \nResLow per view until step \nStgLongSwitch and at \nRes afterwards, a switch point decided on the monitoring set (\cref{sec:s_evalsets}); all other stages use \nRes.
An exponential moving average of the LoRA weights (decay \nStgEma) is kept from the long run on.
The final stage starts from the long run's raw weights rather than their average, because the raw weights were slightly better on the standard test set (LPIPS \nLongRawAll against \nLongEmaAll); this is one of the three decisions that used test clips (\cref{sec:exp_setup}). Its own average starts from the weights it loads.
It switches to a cosine decay at step \nStgConvDecayAt, when a plateau rule on the monitoring set, fixed before the stage started, fired; this is the earliest step the rule allowed.
In total the model saw \nStgWindowsTotal training windows, which the main paper rounds to \nTrainWindows.
Training used about \nGpuHoursMain GPU hours on NVIDIA RTX 6000 Ada GPUs, measured from logged step times on shared machines, which include slow-downs from other jobs.

\begin{table}[t]
\centering
\footnotesize
\setlength{\tabcolsep}{2.5pt}
\begin{tabular}{@{}lrrcrrl@{}}
\toprule
Stage & Steps & Batch & LR & Windows & Data & Resolution \\
\midrule
1 & \nStgCTwoSteps & \nStgCTwoBatch & \nStgCTwoLr & \nStgCTwoWindows & \nHoursVTwo\,h & \nRes \\
2 & \nStgCTwoLongSteps & \nStgCTwoLongBatch & \nStgCTwoLongLr & \nStgCTwoLongWindows & \nHoursVTwo\,h & \nRes \\
3 & \nStgBigSteps & \nStgBigBatch & \nStgBigLr & \nStgBigWindows & \nHoursVTwo\,h & \nRes \\
4 & \nStgLongSteps & \nStgLongBatch & \nStgLongLr & \nStgLongWindows & $\to$\,\nHoursTotal\,h & $\to$\,\nRes \\
5 & \nStgConvSteps & \nStgConvBatch & \nStgConvLr & \nStgConvWindows & \nHoursTotal\,h & \nRes \\
\bottomrule
\end{tabular}
\caption{\textbf{Training stages of \method.} Batch is the number of clips per optimizer step. Stage 2 continues stage 1 with its optimizer state and ends with a cosine decay from step \nStgCTwoLongDecayAt of the joint count; stages 3 to 5 start a new optimizer, warm up for \nStgBigWarm, \nStgLongWarm and \nStgConvWarm steps, and end with cosine decays over \nStgBigDecaySteps, \nStgLongDecaySteps and \nStgConvDecaySteps steps. Stage 4 (the long run) switches from \nHoursVTwo to \nHoursTotal hours of data after step \nStgLongDataSwap and from \nResLow to \nRes after step \nStgLongSwitch; stage 5 is the final stage.}
\label{tab:s_stages}
\end{table}

\paragraph{Data mixture.}
Each training sample draws a dataset with probabilities \nMixEgoPro (EgoPro), \nMixEgoViz (EgoViz), \nMixHumanTouch (HumanTouch) and \nMixEgoTouch (EgoTouch), and then a window of \nFrames consecutive frames uniformly among all valid start frames of that dataset, so that longer episodes contribute more windows.
Relative to their size, the smaller datasets are therefore sampled more often.
Start frames at which a wrist camera is frozen are excluded.

\paragraph{Text.}
The task descriptions of the paired datasets come from their metadata: the task name or the task description that each dataset provides, followed by a fixed sentence that names the camera rig (for example, ``Seen from a head-mounted camera and two wide-angle cameras on wrist straps'').
For in-the-wild windows, the caption comes from a vision-language model (\cref{sec:s_wild}), followed by the rig sentence of EgoPro.

\subsection{Sampling}
\label{sec:s_sampling}

We integrate the flow ODE with \nSampleSteps Euler steps on the shifted noise schedule.
The initial noise of clip $i$ is drawn with seed \nSeedBase{}${}+i$, so every compared model and condition starts from the same noise.
Head guidance with $w{=}\nGuideW$ evaluates the network twice per step: once with the head video and once with the grey head latent, both with the same text.
Without guidance ($w{=}1$), generating both wrists of a \nFrames-frame clip at \nRes takes about \nSecPerClip GPU-seconds.
Fewer steps cost little: relative to \nSampleSteps steps, LPIPS changes by \nStepsTwentyAll, \nStepsFifteenAll and \nStepsTenAll with \nStepsCmpA, \nStepsCmpB and \nStepsCmpC steps on \nStepsN clips.
The analysis experiments that generate many samples (the selection pool and the requested-camera evaluation) use \nPoolSteps steps.

\subsection{Ablations}
\label{sec:s_ablation}

\Cref{tab:s_ablation} lists the ablations of \cref{sec:ablation} with their confidence intervals.

\begin{table}[t]
\centering
\small
\setlength{\tabcolsep}{4pt}
\begin{tabular}{@{}lcc@{}}
\toprule
Variant & $\Delta$ LPIPS ($\times 10^{-3}$) & Clips \\
\midrule
Guidance $w{=}3$ vs.\ $1$ (val.) & \nCfgValThreeAllK\ \nCfgValThreeAllKCI & \nCfgValThreeAllN \\
Guidance $w{=}5$ vs.\ $1$ (val.) & \nCfgValFiveAllK\ \nCfgValFiveAllKCI & \nCfgValFiveAllN \\
Guidance $w{=}3$ vs.\ $1$ (test) & \nCfgTestAllK\ \nCfgTestAllKCI & \nCfgTestAllN \\
\midrule
Reprojection, reference pose & \nWarpZeroAllK\ \nWarpZeroAllKCI & \nWarpZeroAllN \\
Reprojection, 20$^\circ$ error & \nWarpTwentyAllK\ \nWarpTwentyAllKCI & \nWarpTwentyAllN \\
Reprojection, 40$^\circ$ error & \nWarpFortyAllK\ \nWarpFortyAllKCI & \nWarpFortyAllN \\
Zoomed hand crops & \nZoomAllK\ \nZoomAllKCI & \nZoomAllN \\
\bottomrule
\end{tabular}
\caption{\textbf{Ablations.} Paired LPIPS difference of each variant against its control; negative is better. The reprojection and zoom variants are short runs compared with a control trained on the same schedule; the reference pose is estimated from the head and the real wrist frames together, and the errors are added to it.}
\label{tab:s_ablation}
\end{table}

\subsection{Camera-conditioned model}
\label{sec:s_camera}

\paragraph{Condition.}
Each panel view is divided into tokens of \nTokPx{}$\times$\nTokPx pixels, the footprint of one latent token.
For every token we compute the viewing ray of the camera at \nSubPts{}$\times$\nSubPts sub-points, each given by its unit direction in the frame of the rig's reference camera and a flag that is zero outside the valid image area.
A one-hot code of the mount (\nMounts classes: the head and \nWristMounts wrist mounts) completes a \nRayFeat-dimensional vector.
An MLP (width \nInjHidden) maps it to an embedding of size \nInjEmb, and one zero-initialized linear layer per transformer block adds it to that block's input.
With the camera branch, the model has \nCamParams trainable parameters instead of \nLoraParams.
The condition is dropped for \nCamDrop of the samples; the frames are still re-rendered in that case.

\paragraph{Virtual cameras.}
\Cref{tab:s_virtual} lists the distribution of the virtual cameras.
A virtual camera is never wider than the source camera, since the missing periphery cannot be rendered.
Images and ray maps are rendered from the same camera model (\cref{sec:s_calib}).

\begin{table}[t]
\centering
\footnotesize
\setlength{\tabcolsep}{3pt}
\begin{tabular}{@{}lll@{}}
\toprule
Parameter & Wrist views & Head view \\
\midrule
Keep the real camera & \nVcKeepWrist & \nVcKeepHead \\
Roll & \nVcRollPWrist, $\pm$\nVcRollWrist{}$^\circ$ & \nVcRollPHead, $\pm$\nVcRollHead{}$^\circ$ \\
Lens & \multicolumn{2}{l}{same, pinhole or equidistant (uniform)} \\
Horizontal FoV & \multicolumn{2}{l}{\nVcFovLo--\nVcFovHi of the source, pinhole $\leq$ \nVcPinMax{}$^\circ$} \\
Focal aspect ratio & \multicolumn{2}{l}{$\pm$\nVcAspect{}\%} \\
Principal point & \multicolumn{2}{l}{$\pm$\nVcPpX px ($x$), $\pm$\nVcPpY px ($y$)} \\
Letterbox & -- & \nVcLetterP, bars \nVcBarLo--\nVcBarHi px \\
\bottomrule
\end{tabular}
\caption{\textbf{Virtual-camera augmentation.} Probabilities and ranges per view of a training sample (canvas \nCanvas). Roll is applied with the given probability and drawn uniformly in the given range.}
\label{tab:s_virtual}
\end{table}

\paragraph{Training and evaluation.}
The camera-conditioned model and \method start from the same long-run checkpoint (step \nCamStart) and follow the same final-stage schedule: \nCamSteps steps at batch \nCamBatch, cosine decay from step \nCamDecayAt, and the same weight averaging.
They differ only in the camera condition, the virtual cameras and the removal of the rig sentence from the text, which is replaced by a short statement of whether the hands are bare or gloved.
For the requested-camera evaluation (\cref{tab:s_camera}), the narrow camera is a pinhole with a \nCamNarrowFov{}$^\circ$ horizontal field of view, the rotated camera is the native camera rolled by \nCamRollDeg{}$^\circ$, and the fisheye is an equidistant lens with the native field of view.
The reference is the real wrist video re-rendered into the requested camera, and both models are sampled with \nPoolSteps steps without guidance.
For the re-rendering oracle, \method's output in the dataset's own wrist camera (the model without the condition, \nPoolSteps steps, $w{=}1$, one sample per clip with the seeds of the ``w/o cond.'' column) is re-rendered into the requested camera with the function, lens model and parameters that produce the reference, using bilinear resampling; regions outside the native view are black in both.
One sample per clip serves all three requested cameras, since the model without the condition does not see the request.
The samples were generated again for this baseline because the original evaluation kept no videos; scored without re-rendering, they reproduce the ``w/o cond.'' column to within \nWarpReproMaxDiff LPIPS on average.
The native camera covers the whole requested frame for the narrow camera, \nWarpCoverRoll of it for the rotated camera and \nWarpCoverFisheye for the fisheye, so on the rotated camera part of the advantage of the oracle comes from matching the black corners of the reference exactly.

\begin{table}[t]
\centering
\small
\setlength{\tabcolsep}{3pt}
\begin{tabular}{@{}lccccc@{}}
\toprule
Camera & w/o cond. & with cond. & oracle & $\Delta$ cond. & $\Delta$ oracle \\
\midrule
Narrow FoV & \nCamLpNarrowWithoutAll & \nCamLpNarrowWithAll & \nCamLpNarrowWarpAll & \nCamNarrowAll & \nWarpNarrowAll \\
Rotated & \nCamLpRollWithoutAll & \nCamLpRollWithAll & \nCamLpRollWarpAll & \nCamRollAll & \nWarpRollAll \\
Fisheye & \nCamLpFisheyeWithoutAll & \nCamLpFisheyeWithAll & \nCamLpFisheyeWarpAll & \nCamFisheyeAll & \nWarpFisheyeAll \\
\midrule
Native (cost) & \nLpOursCfgOneAll & \nLpCamNativeAll & \nLpOursCfgOneAll & \nCamNativeAll & \nWarpNativeAll \\
\bottomrule
\end{tabular}
\caption{\textbf{Camera control.} LPIPS against the real wrist video re-rendered into the requested camera, without and with the camera condition, on \nCamNarrowAllN clips; both models are sampled without guidance and with \nPoolSteps steps. \emph{Oracle}: the output without the condition, generated in the native camera and re-rendered into the requested camera by the renderer of the reference. $\Delta$ cond.: with minus without the condition; $\Delta$ oracle: oracle minus with the condition; negative is better. The last row compares the camera-conditioned model with \method on each dataset's own camera, where the oracle re-renders nothing, on the standard test set (\nTestClips clips, sampled as in \cref{tab:main} with $w{=}1$).}
\label{tab:s_camera}
\end{table}

\begin{figure*}[t]
\centering
\includegraphics[width=\textwidth]{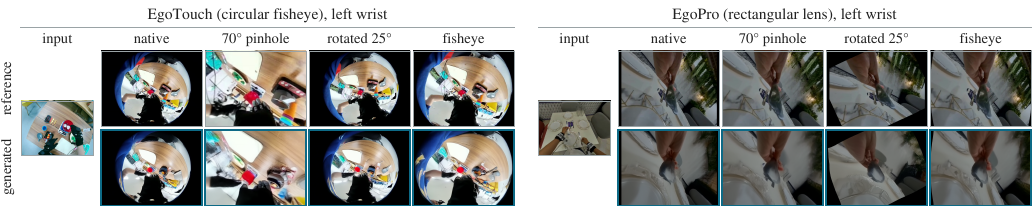}
\caption{\textbf{The generated wrist view follows the requested camera.}
Top: the real wrist video re-rendered into the requested camera (the reference of \cref{tab:s_camera}); bottom: the camera-conditioned model with that camera requested, from the same noise in every column (left wrist, frame 24). Both clips are close to their dataset's median result and were chosen without looking at the samples.}
\label{fig:s_camera}
\end{figure*}

\paragraph{Results.}
\Cref{tab:s_camera} compares the camera-conditioned model with the same model trained without the condition from the same starting point and with the same schedule, and \cref{fig:s_camera} shows examples.
The paired LPIPS difference is \nCamNarrowAll for a narrower field of view, \nCamRollAll for a rotated camera and \nCamFisheyeAll for a fisheye lens, and the condition improves \nCamSigCombos of the \nCamCombos combinations of dataset and requested camera.
The exception is a fisheye request on HumanTouch, whose native lens is already close to the requested one.
The oracle is better than the condition for every requested camera (\nWarpNarrowAll \nWarpNarrowAllCI, \nWarpRollAll \nWarpRollAllCI and \nWarpFisheyeAll \nWarpFisheyeAllCI) and in all \nWarpSigCombos combinations of dataset and requested camera; the condition closes \nCamGapNarrow, \nCamGapRoll and \nCamGapFisheye of the gap between the model without the condition and the oracle.
The oracle shares the resampling of the reference and its black regions outside the native view, which favors it on the rotated camera, but the narrow camera lies entirely inside the native view and the oracle is still better there.
Our references cannot test requests that re-rendering cannot serve, such as a camera wider than the native one or mounted elsewhere, since no recording exists for them.
On each dataset's native camera, the camera-conditioned model is worse than \method by \nCamNativeAll \nCamNativeAllCI, with the largest cost on EgoTouch (\nCamNativeEgoTouch).
We think the cause is that most wrist samples are re-rendered during training, so less training is spent on the native cameras, and the circular fisheye of EgoTouch is the most different from the virtual cameras.

\subsection{WorldWander baseline}
\label{sec:s_ww}

WorldWander~\cite{song2026worldwander} is fine-tuned following its public recipe, starting from its released LoRA weights, on the same training split and dataset weights as \method.
Its input is the head video and its target is one wrist video, with the side drawn at random per sample; the prompt is WorldWander's instruction for the target view followed by our task description.
Following its recipe, we train \nWwSteps steps at batch \nWwBatch with learning rate \nWwLr (\nWwParams LoRA parameters).
At test time we sample it with its own sampler and settings (\nWwSampleSteps steps, guidance scale \nWwCfg), once per wrist, from the same head window as \method.
The zero-shot model (the released weights without fine-tuning) was evaluated on a fixed subset of \nLpWwZsN test clips, every \nWwZsEvery{}th clip of its sampling list.
Fine-tuning used about \nGpuHoursWw logged GPU hours on shared machines.

\section{Data}
\label{sec:s_data}

\subsection{Datasets and splits}
\label{sec:s_splits}

\Cref{tab:s_datasets} lists the four training datasets with their wrist cameras, and \cref{tab:s_splits} the episodes and hours of each split.
EgoTouch~\cite{zhou2026touchanything} comes with an official split, which we use.
For the other datasets we assign whole recording units to splits by hashing their identifiers, so that no unit appears in two splits: sessions for HumanTouch~\cite{lyu2026humantouch,chen2026spark0}, recording sequences for EgoViz~\cite{humaid2026egoviz}, and environments, then episodes, for EgoPro~\cite{lightwheel2026egosuite}.
The unseen test split of EgoPro contains environments that never appear in training, and that of HumanTouch contains a task that never appears in training.
EgoPro episodes released after our split was fixed were added to training if the original rule assigned them to training, and otherwise kept in a separate held-out split that is never trained on.
All videos are stored at \nFps frames per second; each view is resized to \nRes, with black bars where the aspect ratio differs.

\begin{table}[t]
\centering
\small
\setlength{\tabcolsep}{4pt}
\begin{tabular}{@{}lrlr@{}}
\toprule
Dataset & Hours & Wrist camera & Test \\
\midrule
EgoPro~\cite{lightwheel2026egosuite} & \nHoursEgoPro & pinhole, 78--86$^\circ$ & \nTestClipsEgoPro \\
EgoTouch~\cite{zhou2026touchanything} & \nHoursEgoTouch & circular fisheye, 154--167$^\circ$ & \nTestClipsEgoTouch \\
EgoViz~\cite{humaid2026egoviz} & \nHoursEgoViz & wide angle, 107$^\circ$ & \nTestClipsEgoViz \\
HumanTouch~\cite{lyu2026humantouch,chen2026spark0} & \nHoursHumanTouch & wide angle, 111--116$^\circ$ & \nTestClipsHumanTouch \\
\bottomrule
\end{tabular}
\caption{\textbf{Training data.} Hours of paired head and wrist video in the training split, the horizontal field of view of the wrist cameras from our calibration, and the number of test clips.}
\label{tab:s_datasets}
\end{table}

\begin{table*}[t]
\centering
\small
\setlength{\tabcolsep}{6pt}
\begin{tabular}{@{}lrrrrrrrrrr@{}}
\toprule
& \multicolumn{2}{c}{Train} & \multicolumn{2}{c}{Validation} & \multicolumn{2}{c}{Test, seen} & \multicolumn{2}{c}{Test, unseen} & \multicolumn{2}{c}{Held out} \\
\cmidrule(lr){2-3}\cmidrule(lr){4-5}\cmidrule(lr){6-7}\cmidrule(lr){8-9}\cmidrule(l){10-11}
Dataset & Ep. & Hours & Ep. & Hours & Ep. & Hours & Ep. & Hours & Ep. & Hours \\
\midrule
EgoPro & \nEpEgoProTrain & \nHrEgoProTrain & \nEpEgoProVal & \nHrEgoProVal & \nEpEgoProSeen & \nHrEgoProSeen & \nEpEgoProUnseen & \nHrEgoProUnseen & \nEpEgoProHeld & \nHrEgoProHeld \\
EgoTouch & \nEpEgoTouchTrain & \nHrEgoTouchTrain & \nEpEgoTouchVal & \nHrEgoTouchVal & \nEpEgoTouchSeen & \nHrEgoTouchSeen & \nEpEgoTouchUnseen & \nHrEgoTouchUnseen & -- & -- \\
EgoViz & \nEpEgoVizTrain & \nHrEgoVizTrain & \nEpEgoVizVal & \nHrEgoVizVal & \nEpEgoVizSeen & \nHrEgoVizSeen & -- & -- & -- & -- \\
HumanTouch & \nEpHumanTouchTrain & \nHrHumanTouchTrain & \nEpHumanTouchVal & \nHrHumanTouchVal & \nEpHumanTouchSeen & \nHrHumanTouchSeen & \nEpHumanTouchUnseen & \nHrHumanTouchUnseen & -- & -- \\
\midrule
Total & \nEpAllTrain & \nHrAllTrain & \nEpAllVal & \nHrAllVal & \nEpAllSeen & \nHrAllSeen & \nEpAllUnseen & \nHrAllUnseen & \nEpAllHeld & \nHrAllHeld \\
\bottomrule
\end{tabular}
\caption{\textbf{Splits.} Episodes and hours of paired head and wrist video. The unseen test split holds new environments (EgoPro) or a new task (HumanTouch); EgoTouch uses its official split. The held-out EgoPro split is used only for the extended evaluation set and the analyses of HCC.}
\label{tab:s_splits}
\end{table*}

\begin{table*}[t]
\centering
\footnotesize
\begin{tabular}{@{}lcccccccc@{}}
\toprule
& \multicolumn{5}{c}{LPIPS $\downarrow$} & & \multicolumn{2}{c}{All clips} \\
\cmidrule(lr){2-6}\cmidrule(l){8-9}
Method & EgoPro (\nRestNEgoPro) & EgoTouch (\nRestNEgoTouch) & EgoViz (\nRestNEgoViz) & HumanTouch (\nRestNHumanTouch) & All (\nRestN) & & Layout PSNR $\uparrow$ & Ours better \\
\midrule
Untuned Wan2.2~\cite{wan2025wan} & \nRestLpBaseEgoPro & \nRestLpBaseEgoTouch & \nRestLpBaseEgoViz & \nRestLpBaseHumanTouch & \nRestLpBaseAll & & \nRestPsnrBaseAll & \nRestDiffOursVsBaseAllBetter \\
Head-view crop & \nRestLpCropEgoPro & \nRestLpCropEgoTouch & \nRestLpCropEgoViz & \nRestLpCropHumanTouch & \nRestLpCropAll & & \nRestPsnrCropAll & \nRestDiffOursVsCropAllBetter \\
Another clip's wrist video & \nRestLpOtherEgoPro & \nRestLpOtherEgoTouch & \nRestLpOtherEgoViz & \nRestLpOtherHumanTouch & \nRestLpOtherAll & & \nRestPsnrOtherAll & \nRestDiffOursVsOtherAllBetter \\
WorldWander~\cite{song2026worldwander}, fine-tuned & \nRestLpWwEgoPro & \nRestLpWwEgoTouch & \nRestLpWwEgoViz & \nRestLpWwHumanTouch & \nRestLpWwAll & & \nRestPsnrWwAll & \nRestDiffOursVsWwAllBetter \\
\midrule
\method, $w{=}1$ & \nRestLpOursCfgOneEgoPro & \nRestLpOursCfgOneEgoTouch & \nRestLpOursCfgOneEgoViz & \nRestLpOursCfgOneHumanTouch & \nRestLpOursCfgOneAll & & \nRestPsnrOursCfgOneAll & \\
\method, $w{=}3$ & \nRestLpOursCfgThreeEgoPro & \nRestLpOursCfgThreeEgoTouch & \nRestLpOursCfgThreeEgoViz & \nRestLpOursCfgThreeHumanTouch & \nRestLpOursCfgThreeAll & & \nRestPsnrOursCfgThreeAll & \\
\bottomrule
\end{tabular}
\caption{\textbf{Main comparison on the clips that were never monitored.} \Cref{tab:main} restricted to the \nRestN clips of the standard test set that are not in the monitoring set; clip counts in parentheses. Against WorldWander the paired difference is \nRestDiffOursVsWwAll \nRestDiffOursVsWwAllCI (\nRestRelOursVsWw), compared with \nDiffOursVsWwAll \nDiffOursVsWwAllCI on all \nTestClips clips and \nMonDiffOursVsWwAll \nMonDiffOursVsWwAllCI on the \nSetCompactOverlap monitored clips, so the gap does not come from the monitored clips. Only \nRestNEgoViz EgoViz clips remain, because the monitoring set contains most of the EgoViz test episodes.}
\label{tab:s_rest}
\end{table*}

\subsection{Evaluation sets}
\label{sec:s_evalsets}

Every evaluation set takes one window per episode, the valid window closest to the middle of the episode, and at most a fixed number of episodes per dataset, spread evenly over the sorted episode list.
All models are evaluated on the same windows with the same initial noise.
\begin{itemize}
\item The \emph{standard test set} (\nSetStd clips) is drawn from both test splits: \nSetStdEgoPro EgoPro clips (\nSetStdEgoProUnseen from unseen environments), \nSetStdEgoTouch EgoTouch (\nSetStdEgoTouchUnseen unseen), \nSetStdEgoViz EgoViz and \nSetStdHumanTouch HumanTouch (\nSetStdHumanTouchUnseen from the unseen task). The main results are on this set.
\item The \emph{validation set} (\nSetVal clips, \nSetValEgoPro per dataset) is used to choose the guidance weight and, in the selection experiment, the seed pair that continues to the later rounds (\cref{sec:exp_select}).
\item The \emph{extended EgoPro set} (\nSetExt clips: \nSetExtUnseen from unseen environments and \nSetExtHeld from the held-out split) is the second test set of the selection experiment (\cref{sec:exp_select}).
\item The \emph{monitoring set} (\nSetCompact clips: \nSetCompactEgoPro EgoPro, \nSetCompactEgoTouch EgoTouch, \nSetCompactEgoViz EgoViz, \nSetCompactHumanTouch HumanTouch) was monitored during training: the long run's switch to the full resolution and the start of the final learning-rate decay (\cref{sec:s_train}) were decided on it. It is drawn from the test splits, and \nSetCompactOverlap of its clips are also in the standard test set. It also serves for the requested-camera evaluation, in which both compared models follow the same schedule. \Cref{tab:s_rest} repeats the main comparison on the other \nRestN clips of the standard test set, which were never monitored.
\item The \emph{unselected set} (\nXfamN clips: \nUnselEgoPro EgoPro clips and \nUnselOther random clips of each other dataset, all from the standard test set) was chosen without looking at any score. It is used for the comparison of HCC across methods (\cref{sec:exp_score}), for \cref{fig:qual}, and for the human study (\cref{sec:s_study}).
\end{itemize}

\subsection{In-the-wild windows}
\label{sec:s_wild}

\begin{table}[t]
\centering
\small
\begin{tabular}{@{}lr@{}}
\toprule
Step & Windows \\
\midrule
Candidates from \nWildClips clips & \nWildCand \\
Both hands detected & \nWildHands \\
Passed the vision-language check, captioned & \nWildScreened \\
Synthesized wrists scorable & \nWildScorable \\
No circular-fisheye look (selection pool) & \nRsiPool \\
Kept by HCC (top \nRsiKeepShare) & \nRsiKept \\
\bottomrule
\end{tabular}
\caption{\textbf{In-the-wild windows} from Egocentric-10K. The pool holds \nWildPoolHours hours from \nWildFactoriesPool factories; the kept windows hold \nWildKeptHours hours from \nWildFactoriesKept factories.}
\label{tab:s_funnel}
\end{table}

The in-the-wild video comes from a downloaded subset of Egocentric-10K~\cite{buildai2025egocentric10k}, \nWildClips clips recorded by factory workers with a head-mounted fisheye camera and no wrist cameras.
We crop the center of each frame to the aspect ratio of our head views and resize it to \nRes.
\Cref{tab:s_funnel} shows how the selection pool of \cref{sec:exp_select} was formed.
Candidate windows are taken at regular intervals.
A hand detector~\cite{potamias2025wilor} keeps windows in which both hands appear in the lower part of the frame, one on each side.
A vision-language model (Gemini 4 Argon) then checks key frames for the wearer's own two hands, bare or in dark gloves, handling an object, and writes a one-sentence caption in the same call.
We synthesize both wrist videos for every remaining window with \method (\nPoolSteps steps, without guidance, one sample per window) and score them with HCC.
Windows whose synthesized wrists cannot be scored, or look like a circular fisheye image, which the target wrist camera never produces, are dropped.
HCC then keeps the top \nRsiKeepShare of the pool.

\section{Camera calibration}
\label{sec:s_calib}

\paragraph{Radius-to-angle tables.}
Only the EgoPro head camera and the Egocentric-10K camera come with vendor intrinsics.
For every other camera we estimate the lens with AnyCalib~\cite{tiradogarin2025anycalib}, which predicts a per-pixel ray field from a single image.
We run it on sampled frames of each device, convert each prediction to the angle between every pixel's ray and the optical axis, and pool the angles in narrow bins of the distance to the principal point.
The median angle per bin gives a monotone table from image radius to ray angle.
Projection and unprojection interpolate this table, so the same code renders images into a virtual camera and computes the ray maps of the condition (\cref{sec:camera}).
Pixels that are black in every sampled frame lie outside the valid image area, which is a circle for the EgoTouch fisheye lenses and a rectangle otherwise.
\Cref{tab:s_calib} lists the resulting fields of view.

\begin{table}[t]
\centering
\small
\setlength{\tabcolsep}{4pt}
\begin{tabular}{@{}lcccccc@{}}
\toprule
& \multicolumn{2}{c}{Head} & \multicolumn{2}{c}{Left wrist} & \multicolumn{2}{c}{Right wrist} \\
\cmidrule(lr){2-3}\cmidrule(lr){4-5}\cmidrule(l){6-7}
Dataset & H & V & H & V & H & V \\
\midrule
EgoPro & \nCalEgoProHeadH & \nCalEgoProHeadV & \nCalEgoProLeftH & \nCalEgoProLeftV & \nCalEgoProRightH & \nCalEgoProRightV \\
EgoTouch & \nCalEgoTouchHeadH & \nCalEgoTouchHeadV & \nCalEgoTouchLeftH & \nCalEgoTouchLeftV & \nCalEgoTouchRightH & \nCalEgoTouchRightV \\
EgoViz & \nCalEgoVizHeadH & \nCalEgoVizHeadV & \nCalEgoVizLeftH & \nCalEgoVizLeftV & \nCalEgoVizRightH & \nCalEgoVizRightV \\
HumanTouch & \nCalHumanTouchHeadH & \nCalHumanTouchHeadV & \nCalHumanTouchLeftH & \nCalHumanTouchLeftV & \nCalHumanTouchRightH & \nCalHumanTouchRightV \\
Egocentric-10K & \nCalWildHeadH & \nCalWildHeadV & -- & -- & -- & -- \\
\bottomrule
\end{tabular}
\caption{\textbf{Fields of view} (degrees, horizontal and vertical, across the valid area of the \nCanvas canvas). The EgoTouch wrist cameras are circular fisheye lenses; the EgoPro head and the Egocentric-10K camera use vendor intrinsics, all others the AnyCalib tables.}
\label{tab:s_calib}
\end{table}

\paragraph{An error in the EgoTouch right-wrist table.}
After the experiments were finished, we found an error in the table of the EgoTouch right wrist camera.
Its first entry after the center maps a radius of \nCalBugR pixels to \nCalBugDeg{}$^\circ$, and the angle then stays flat up to a radius of about \nCalBugFlatR pixels.
All rays within \nCalBugDeg{}$^\circ$ of the optical axis therefore project to within \nCalBugR pixels of the principal point.
When a real frame of this camera is re-rendered into a pinhole or equidistant camera, the center of the result is filled with a uniform disk, with a radius of about \nCalBugDisk pixels for the \nCamNarrowFov{}$^\circ$ pinhole camera of the evaluation.
The left-wrist table does not have this jump.
The error affects two things: the references of the EgoTouch right wrist in the narrow and fisheye rows of \cref{tab:s_camera}, and the virtual-camera augmentations of the EgoTouch right wrist during training of the camera-conditioned model.
The native-camera evaluation re-renders nothing; there the table enters only through the ray map, which the model saw in the same form during training.

We did not retrain, and instead measured how much the error can change the reported numbers.
\Cref{tab:s_calibbug} repeats the camera-control comparison with the EgoTouch right wrists left out (EgoTouch clips scored on the left wrist only).
The gains are within a few thousandths of those in \cref{tab:s_camera}.
On EgoTouch, the unaffected left wrist gains as much as or more than the right wrist for the narrow and fisheye requests, so the error does not create the gain.
The native-camera cost of the condition on EgoTouch is also the same for the left (\nCalNativeL \nCalNativeLCI) and the right wrist (\nCalNativeR \nCalNativeRCI), so the error does not explain why EgoTouch has the largest cost.

\begin{table}[t]
\centering
\small
\setlength{\tabcolsep}{3pt}
\begin{tabular}{@{}lccc@{}}
\toprule
& All wrists & \multicolumn{2}{c}{EgoTouch only (\nCalEtN clips)} \\
\cmidrule(l){3-4}
Requested camera & \cref{tab:s_camera} & left & right \\
\midrule
Narrow & \nCamNarrowAll & \nCalEtNarrowL & \nCalEtNarrowR \\
Rotated & \nCamRollAll & \nCalEtRollL & \nCalEtRollR \\
Fisheye & \nCamFisheyeAll & \nCalEtFisheyeL & \nCalEtFisheyeR \\
\midrule
& \multicolumn{3}{c}{Without EgoTouch right wrists} \\
\cmidrule(l){2-4}
Narrow & \multicolumn{3}{c}{\nCalImpNarrow\ \nCalImpNarrowCI} \\
Rotated & \multicolumn{3}{c}{\nCalImpRoll\ \nCalImpRollCI} \\
Fisheye & \multicolumn{3}{c}{\nCalImpFisheye\ \nCalImpFisheyeCI} \\
\bottomrule
\end{tabular}
\caption{\textbf{Impact of the EgoTouch right-wrist calibration error} on the camera-control gain ($\Delta$ LPIPS of the camera-conditioned model against the model without the condition; negative is better).}
\label{tab:s_calibbug}
\end{table}

\section{Hand-centric cross-view consistency (HCC)}
\label{sec:s_score}

\begin{table*}[t]
\centering
\small
\setlength{\tabcolsep}{8pt}
\begin{tabular}{@{}lcc@{}}
\toprule
Score & Across checkpoints & Same model \\
\midrule
HCC, both directions & \nHumOursCross\ \nHumOursCrossCI & \nHumOursSame\ \nHumOursSameCI \\
\quad wrist $\to$ hand region & \nHumPrecCross\ \nHumPrecCrossCI & \nHumPrecSame\ \nHumPrecSameCI \\
\quad hand region $\to$ wrist & \nHumRecCross\ \nHumRecCrossCI & \nHumRecSame\ \nHumRecSameCI \\
\quad whole head frame & \nHumWholeCross\ \nHumWholeCrossCI & \nHumWholeSame\ \nHumWholeSameCI \\
MASt3R inliers~\cite{leroy2024grounding} & \nHumMastrCross\ \nHumMastrCrossCI & \nHumMastrSame\ \nHumMastrSameCI \\
CLIP similarity~\cite{radford2021learning} & \nHumClipCross\ \nHumClipCrossCI & \nHumClipSame\ \nHumClipSameCI \\
MEt3R~\cite{asim2025met3r} & \nHumMetrCross\ \nHumMetrCrossCI & \nHumMetrSame\ \nHumMetrSameCI \\
MEt3R overlap & \nHumMetrOvCross\ \nHumMetrOvCrossCI & \nHumMetrOvSame\ \nHumMetrOvSameCI \\
\midrule
LPIPS (needs the real wrist) & \nHumLpipsCross\ \nHumLpipsCrossCI & \nHumLpipsSame\ \nHumLpipsSameCI \\
Two people & \nHumHumanCross & \nHumHumanSame \\
\bottomrule
\end{tabular}
\caption{\textbf{Agreement with human choices} (Spearman correlation with 95\% intervals; \nHumNCross answers across checkpoints, \nHumNSame between samples of one model). Two people: $2p-1$ from their agreement rate $p$.}
\label{tab:s_agree}
\end{table*}

\begin{table*}[t]
\centering
\small
\setlength{\tabcolsep}{4.5pt}
\begin{tabular}{@{}lccccccccc@{}}
\toprule
& \multicolumn{4}{c}{Ours} & \multicolumn{2}{c}{Auxiliary tests} & \multicolumn{3}{c}{Other scores} \\
\cmidrule(lr){2-5}\cmidrule(lr){6-7}\cmidrule(l){8-10}
Corruption & HCC & Wrist $\to$ region & Region $\to$ wrist & Whole frame & Temporal & Reflection & MASt3R & MEt3R & CLIP \\
\midrule
Swap with another clip & \nSAucSwapOurs & \nSAucSwapPrec & \nSAucSwapRec & \nSAucSwapWhole & \nSAucSwapTime & \nSAucSwapRefl & \nSAucSwapMastr & \nSAucSwapMetr & \nSAucSwapClip \\
Drift, weak & \nSAucDriftWeakOurs & \nSAucDriftWeakPrec & \nSAucDriftWeakRec & \nSAucDriftWeakWhole & \nSAucDriftWeakTime & \nSAucDriftWeakRefl & \nSAucDriftWeakMastr & \nSAucDriftWeakMetr & \nSAucDriftWeakClip \\
Drift, strong & \nSAucDriftOurs & \nSAucDriftPrec & \nSAucDriftRec & \nSAucDriftWhole & \nSAucDriftTime & \nSAucDriftRefl & \nSAucDriftMastr & \nSAucDriftMetr & \nSAucDriftClip \\
\midrule
Upside down & \nSAucFlipOurs & \nSAucFlipPrec & \nSAucFlipRec & \nSAucFlipWhole & \nSAucFlipTime & \nSAucFlipRefl & \nSAucFlipMastr & \nSAucFlipMetr & \nSAucFlipClip \\
Half turn & \nSAucRotateOurs & \nSAucRotatePrec & \nSAucRotateRec & \nSAucRotateWhole & \nSAucRotateTime & \nSAucRotateRefl & \nSAucRotateMastr & \nSAucRotateMetr & \nSAucRotateClip \\
Halves swapped & \nSAucHalvesOurs & \nSAucHalvesPrec & \nSAucHalvesRec & \nSAucHalvesWhole & \nSAucHalvesTime & \nSAucHalvesRefl & \nSAucHalvesMastr & \nSAucHalvesMetr & \nSAucHalvesClip \\
Mirror & \nSAucMirrorOurs & \nSAucMirrorPrec & \nSAucMirrorRec & \nSAucMirrorWhole & \nSAucMirrorTime & \nSAucMirrorRefl & \nSAucMirrorMastr & \nSAucMirrorMetr & \nSAucMirrorClip \\
\midrule
Hand warp & \nSAucHandwarpOurs & \nSAucHandwarpPrec & \nSAucHandwarpRec & \nSAucHandwarpWhole & \nSAucHandwarpTime & \nSAucHandwarpRefl & \nSAucHandwarpMastr & \nSAucHandwarpMetr & \nSAucHandwarpClip \\
Elastic warp & \nSAucElasticOurs & \nSAucElasticPrec & \nSAucElasticRec & \nSAucElasticWhole & \nSAucElasticTime & \nSAucElasticRefl & \nSAucElasticMastr & \nSAucElasticMetr & \nSAucElasticClip \\
\midrule
Shuffled frames & \nSAucShuffleOurs & \nSAucShufflePrec & \nSAucShuffleRec & \nSAucShuffleWhole & \nSAucShuffleTime & \nSAucShuffleRefl & \nSAucShuffleMastr & \nSAucShuffleMetr & \nSAucShuffleClip \\
Frozen frames & \nSAucFreezeOurs & \nSAucFreezePrec & \nSAucFreezeRec & \nSAucFreezeWhole & \nSAucFreezeTime & \nSAucFreezeRefl & \nSAucFreezeMastr & \nSAucFreezeMetr & \nSAucFreezeClip \\
Flicker & \nSAucFlickerOurs & \nSAucFlickerPrec & \nSAucFlickerRec & \nSAucFlickerWhole & \nSAucFlickerTime & \nSAucFlickerRefl & \nSAucFlickerMastr & \nSAucFlickerMetr & \nSAucFlickerClip \\
Time warp & \nSAucTimewarpOurs & \nSAucTimewarpPrec & \nSAucTimewarpRec & \nSAucTimewarpWhole & \nSAucTimewarpTime & \nSAucTimewarpRefl & \nSAucTimewarpMastr & \nSAucTimewarpMetr & \nSAucTimewarpClip \\
\midrule
Jitter & \nSAucJitterOurs & \nSAucJitterPrec & \nSAucJitterRec & \nSAucJitterWhole & \nSAucJitterTime & \nSAucJitterRefl & \nSAucJitterMastr & \nSAucJitterMetr & \nSAucJitterClip \\
Breathing & \nSAucBreatheOurs & \nSAucBreathePrec & \nSAucBreatheRec & \nSAucBreatheWhole & \nSAucBreatheTime & \nSAucBreatheRefl & \nSAucBreatheMastr & \nSAucBreatheMetr & \nSAucBreatheClip \\
\bottomrule
\end{tabular}
\caption{\textbf{AUROC of each score on corrupted real wrist videos}: the probability that the real video scores higher than its corrupted copy (\nSAucN videos per corruption; chance is one half). The drift is shown at two strengths, the other corruptions at one. Groups: content, geometry, deformation, time and framing.}
\label{tab:s_corrupt}
\end{table*}

\subsection{Implementation}
\label{sec:s_score_impl}
HCC compares the head frame and the wrist frame of the same time at \nXvFrames frames of the window, every \nXvStride{}th frame.
In the head frame, the hand detector of WiLoR~\cite{potamias2025wilor} (confidence at least \nXvConf) gives the box of the hand on the same side as the wrist camera; if it finds none, we take the most confident hand in the corresponding half of the image.
For gloved hands, which this detector often misses, OWLv2~\cite{minderer2023scaling} with text queries for a hand and a black glove (score at least \nXvOwl) provides the box.
The region is a square of \nXvRoi times the longer side of the box, centered on it.
Both frames are resized to \nRes and encoded by DINOv3 ViT-7B/16~\cite{simeoni2025dinov3}; we use the patch tokens of the last layer after the final normalization, normalized to unit length, and cosine similarity.
A frame counts if its region covers at least \nXvMinPatches patches, and HCC is defined for a wrist video if at least \nXvMinFrames frames count; videos for which it is undefined are left out of every analysis.
Almost every real and generated video of the validation studies could be scored, and in the selection pool \nWildRowsNan of the synthesized wrist videos could not (\cref{tab:s_funnel}).
The same frames, region and features are used for every variant in \cref{tab:s_agree,tab:s_corrupt}; the whole-frame variant matches the central half of the wrist frame against all patches of the head frame.

\subsection{Design: nearest neighbors instead of correspondences}
\label{sec:s_score_design}
HCC matches every patch to its nearest neighbor in the other view, in both directions, and does not ask for one-to-one or mutual correspondences.
Three properties of the head-to-wrist pair motivate this choice.
First, the scales differ: the hand and the object it holds cover a few patches of the head region but much of a close-up, often fisheye, wrist view, so many wrist patches correctly match the same head patch.
Second, the views overlap only partly: the wrist camera sees surfaces that the hand hides from the head camera, and the region contains scene that the wrist camera does not see; a one-to-one assignment would have to match these parts as well.
Third, the two directions measure different errors, like precision and recall in BERTScore~\cite{zhang2020bertscore}: content that the wrist view invents lowers the first term, which uses the central half of the wrist frame, and an object that the wrist view omits lowers the second, which uses the region.
A score built on verified correspondences, the MASt3R inlier count~\cite{leroy2024grounding}, agrees less with people across checkpoints (\nHumMastrCross against \nHumOursCross for HCC, \cref{tab:s_agree}) and detects swapped and drifting videos less well (\cref{tab:s_corrupt}).
It is better on corruptions that keep the content and change its layout or timing, such as mirrored videos (AUROC \nSAucMirrorMastr against \nSAucMirrorOurs; \cref{tab:s_corrupt}): nearest-neighbor matching checks what the wrist view shows, not where it shows it (\cref{sec:s_score_fail}).

\subsection{Which part of HCC agrees with people}
\label{sec:s_score_roi}

\Cref{tab:s_agree} extends \cref{fig:score}a.
The two directions of the matching agree with people about equally across checkpoints, and HCC uses their mean.
Matching against the whole head frame is lower, and the region size matters little: squares of \nRoiSmall, \nXvRoi and \nRoiLarge times the hand box reach \nRoiCrossOneFive, \nRoiCrossTwoFive and \nRoiCrossFour across checkpoints.
We chose the region on these answers, so the agreement of the chosen setting is optimistic; the other two settings show how much this choice can matter.

\subsection{Corrupted wrist videos}
\label{sec:s_score_corrupt}

\Cref{tab:s_corrupt} lists every corruption we tested, each applied to \nSAucN real wrist videos of EgoPro, EgoViz and HumanTouch test clips, and the probability that the real video scores higher than its corrupted copy.
The swap replaces the wrist video by that of another clip of the same dataset, and the drift blends it progressively into another clip.
The flip turns the video upside down, the rotation turns it by half a turn, the halves swap exchanges its left and right halves, and the mirror flips it horizontally.
The hand warp deforms the lower part of the frame, where the wearer's hand usually is, and the elastic warp deforms the whole frame.
Shuffle, freeze, flicker and time warp change the order or the rate of the frames, and jitter and breathing change the framing slightly.
HCC detects changes of content (swap, drift) and is weak on changes of geometry that keep the content.
Two auxiliary tests, a temporal test that compares each wrist frame with the head frame of the same time and of other times, and a reflection test that counts mirrored matches, cover the temporal corruptions and the mirror; they are not part of HCC.
Jitter, breathing, flicker and time warp stay near chance for every score: matching single frames across views does not see them.

\subsection{Choosing among samples of one input}
\label{sec:s_score_ceiling}
To test whether HCC can choose the best of several samples of one input, we generated \nCeilSeeds samples for each of \nCeilClips validation and held-out clips with \method (\nPoolSteps steps, without guidance).
Most of the variation of LPIPS lies between clips (\nCeilVarClips of the variance) rather than between samples of one clip (\nCeilVarSeeds).
Choosing the sample with the lowest LPIPS, which needs the ground truth, improves LPIPS over a random sample by \nCeilOracleTwo, \nCeilOracleFour and \nCeilOracleEight for the best of \nCeilNA, \nCeilNB and \nCeilNC samples.
HCC captures \nCeilScoreCapture of the best-of-\nCeilNC gain; the best of the other single features without ground truth, matching against the whole head frame, captures \nCeilFeatCapture, measured on the same data on which it was found to be best among them.
A gradient-boosted ranker trained on several features computed alongside HCC, including the auxiliary tests and the position of the hand in the wrist frame, captures \nCeilRankerCapture on held-out clips.
HCC is therefore useful where differences between clips dominate, for ranking checkpoints and for selecting inputs, and not for choosing among samples of one input.

\begin{table}[t]
\centering
\small
\begin{tabular}{@{}lcc@{}}
\toprule
& HCC $\uparrow$ & LPIPS $\downarrow$ \\
\midrule
Real wrist video & \nCropHccSubReal & -- \\
Head-view crop & \nCropHccSub & \nCropSubLpCrop \\
\method ($w{=}3$) & \nCropHccSubOurs & \nCropSubLpOursThree \\
\method ($w{=}1$) & \nCropHccSubOursOne & \nCropSubLpOursOne \\
WorldWander, fine-tuned & \nCropHccSubWw & \nCropSubLpWw \\
\bottomrule
\end{tabular}
\caption{\textbf{A copy of the head view scores high.} HCC and LPIPS on the \nCropHccSubN clips of the unselected set (\cref{sec:s_evalsets}), the clips with HCC for every entry. The head-view crop is the baseline of \cref{tab:main}.}
\label{tab:s_copy}
\end{table}

\begin{figure*}[t]
\centering
\includegraphics[width=\textwidth]{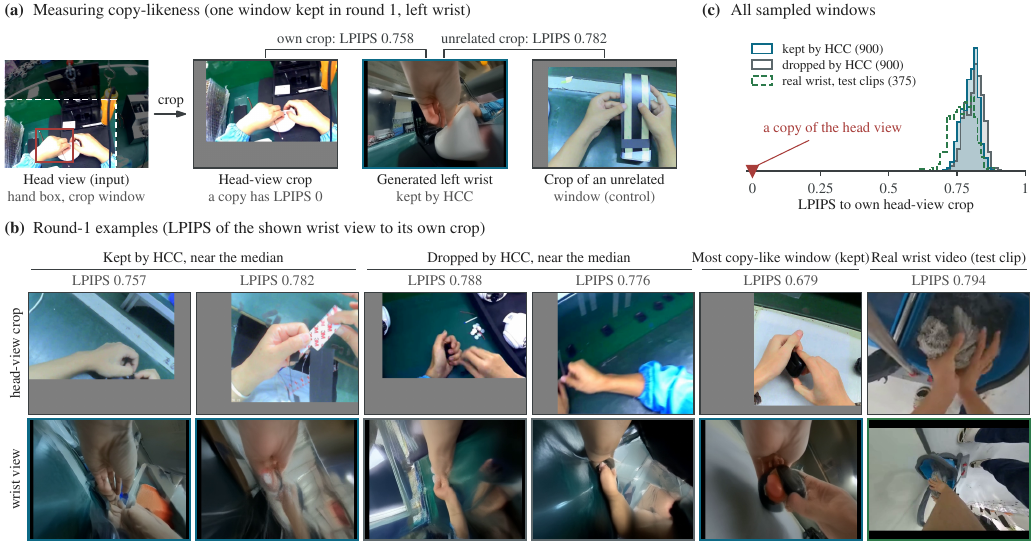}
\caption{\textbf{Measuring whether HCC selection favors copies of the head view.}
(a)~For a window that HCC kept in round~1, the hand on the same side is detected in the head view (red box) and the head-view crop of \cref{tab:main} is cut around it (dashed window, mid-grey outside the frame). A copy of the head view would match this crop exactly (LPIPS 0). We measure LPIPS between the generated wrist view and its own crop and, as a control, the crop of an unrelated window of the same round and group.
(b)~Round-1 windows closest to the median of their group, the most copy-like of all \nCopyNTotal sampled windows, and a real wrist video of a test clip with its crop. Numbers are LPIPS of the shown wrist to its crop, averaged over the 13 analyzed frames.
(c)~LPIPS to the own crop, averaged over both wrists, for all sampled windows of the three rounds and for the real wrist videos of the test clips (\cref{tab:s_copycheck}).}
\label{fig:s_copycheck}
\end{figure*}

\begin{table*}[t]
\centering
\small
\begin{tabular}{@{}lccccc@{}}
\toprule
& \multicolumn{2}{c}{LPIPS to own crop $\downarrow$} & \multicolumn{3}{c}{Kept minus dropped} \\
\cmidrule(lr){2-3}\cmidrule(l){4-6}
Round & Kept & Dropped & Own crop & Unrelated crop & Own minus unrelated \\
\midrule
1 & \nCopyOwnKeptROne & \nCopyOwnDropROne & \nCopyOwnDiffROne\ \nCopyOwnDiffROneCI & \nCopyMisDiffROne\ \nCopyMisDiffROneCI & \nCopySpecDiffROne\ \nCopySpecDiffROneCI \\
2 & \nCopyOwnKeptRTwo & \nCopyOwnDropRTwo & \nCopyOwnDiffRTwo\ \nCopyOwnDiffRTwoCI & \nCopyMisDiffRTwo\ \nCopyMisDiffRTwoCI & \nCopySpecDiffRTwo\ \nCopySpecDiffRTwoCI \\
3 & \nCopyOwnKeptRThree & \nCopyOwnDropRThree & \nCopyOwnDiffRThree\ \nCopyOwnDiffRThreeCI & \nCopyMisDiffRThree\ \nCopyMisDiffRThreeCI & \nCopySpecDiffRThree\ \nCopySpecDiffRThreeCI \\
\midrule
All & \nCopyOwnKeptPool & \nCopyOwnDropPool & \nCopyOwnDiffPool\ \nCopyOwnDiffPoolCI & \nCopyMisDiffPool\ \nCopyMisDiffPoolCI & \nCopySpecDiffPool\ \nCopySpecDiffPoolCI \\
\bottomrule
\end{tabular}
\caption{\textbf{HCC selection does not favor copies of the head view.} LPIPS between the generated wrist views and the crop of the head view around the same-side hand (0 for a copy) for \nCopyNPerGroup kept and \nCopyNPerGroup dropped windows per round of recursive self-improvement. Kept windows are closer to their own crop, but by as much as they are closer to the crop of an unrelated window of the same round and group; the part specific to their own head view does not differ. Real wrist videos are at \nCopyReal from their crop on the test clips. 95\% bootstrap intervals.}
\label{tab:s_copycheck}
\end{table*}

\subsection{Known failure modes}
\label{sec:s_score_fail}

HCC checks content, not geometry: it rewards a wrist view that shows the objects around the hand, but it barely penalizes a correct view at the wrong orientation or mirrored (\cref{tab:s_corrupt}).
The periphery of the wrist view is checked only by the recall direction, so errors near the image border, such as the edge of the table or an unrelated object, cost little.
HCC is sensitive to blur: unguided samples ($w{=}1$) score below WorldWander on the same clips (higher on only \nXfamOursHigher of \nXfamN) although LPIPS prefers them, and head guidance, which raises HCC by \nXfamGuidanceGain \nXfamGuidanceGainCI, reverses the order (\cref{sec:exp_score}).
Gloved hands are often missed by the hand detector and need the open-vocabulary fallback.
When neither detector finds the hand in enough frames, HCC is undefined and the sample is left out, which happens for a few in-the-wild windows (\cref{sec:s_score_impl}).
Finally, the region size was chosen on the human answers that we also use to validate HCC.

\subsection{Copying the head view}
\label{sec:s_score_copy}

\Cref{tab:s_copy} scores the head-view crop of \cref{tab:main}, a zoomed crop of the head video around the same-side hand, with HCC.
On the \nCropHccN scorable test clips it scores \nCropHcc \nCropHccCI, above the real wrist video (\nCropHccReal \nCropHccRealCI) on \nCropHccOverRealShare of the clips.
On the \nCropHccSubN clips with HCC for every entry, it is above \method ($w{=}3$) by \nCropHccMinusOurs \nCropHccMinusOursCI, higher than every other entry on all clips but one (against WorldWander), while LPIPS ranks it last.
Both halves of the score are high for the crop (precision \nCropHccPartCropPrec and recall \nCropHccPartCropRec, against \nCropHccPartRealPrec and \nCropHccPartRealRec for the real video), so neither direction alone is gamed.
The crop is cut from the region that HCC matches against, so it shows that region's content at a similar scale, and HCC, which checks whether the right content appears, accepts it.
A real wrist video scores high, but a high score does not imply a good video; this is why we use HCC to select the data of one model and not as a reward or to compare methods (\cref{sec:exp_score}).

If HCC rewards copying, the windows that it keeps for recursive self-improvement might be the ones that copy the head view.
We test this on \nCopyNPerGroup kept and \nCopyNPerGroup dropped windows drawn at random from each round, with the analysis fixed before any video was read.
For each window we detect the hands in the head view with the rule of HCC, cut the crop of \cref{tab:s_copy} around each hand, and measure LPIPS between each generated wrist view and its crop, which is 0 for a copy.
As a control we measure the same distance to the crop of another window of the same round and group (\cref{fig:s_copycheck}a).
In every round the kept windows are closer to their own crop (\cref{tab:s_copycheck}), by \nCopyOwnDiffPool \nCopyOwnDiffPoolCI over all rounds, while their HCC is higher by \nCopyHccDiffPool.
They are closer by the same amount to the crop of an unrelated window, so the part of the distance that is specific to their own head view does not differ (\nCopySpecDiffPool \nCopySpecDiffPoolCI).
HCC thus keeps windows that look somewhat more like head-view images in general, for example with larger hands in view or from certain factories (with factory held fixed the difference halves to \nCopyFactoryDiff), but not windows that reproduce their own head view.
The shift is \nCopyShiftShare of the distance between a real wrist video and a copy (\nCopyReal \nCopyRealCI for real wrist videos on the test clips).
No sampled window comes close to a copy: the closest is at \nCopyMin, and none is closer than the 5th percentile of real wrist videos (\nCopyRealPfive).
Two caveats apply: the real-wrist level is measured on the paired test sets, whose scenes differ from the in-the-wild windows, and LPIPS would miss a copy that exists only in the feature space of HCC, although it does detect a pixel copy such as the head-view crop.

\section{Contact in generated wrist views}
\label{sec:s_contact}

\begin{table*}[t]
\centering
\footnotesize
\setlength{\tabcolsep}{5pt}
\begin{tabular}{@{}lcccc@{}}
\toprule
& \multicolumn{2}{c}{EgoTouch} & \multicolumn{2}{c}{HumanTouch} \\
\cmidrule(lr){2-3}\cmidrule(l){4-5}
Input & AUROC & Within-clip AUROC & AUROC & Within-clip AUROC \\
\midrule
Task prior & \nCtEgoTouchPriorAuc\ \nCtEgoTouchPriorAucCI & \nCtEgoTouchPriorWithin & \nCtHumanTouchPriorAuc\ \nCtHumanTouchPriorAucCI & \nCtHumanTouchPriorWithin \\
Head frame, crop around the hand & \nCtEgoTouchHeadCropAuc\ \nCtEgoTouchHeadCropAucCI & \nCtEgoTouchHeadCropWithin\ \nCtEgoTouchHeadCropWithinCI & \nCtHumanTouchHeadCropAuc\ \nCtHumanTouchHeadCropAucCI & \nCtHumanTouchHeadCropWithin\ \nCtHumanTouchHeadCropWithinCI \\
Head frame, whole & \nCtEgoTouchHeadFullAuc\ \nCtEgoTouchHeadFullAucCI & \nCtEgoTouchHeadFullWithin\ \nCtEgoTouchHeadFullWithinCI & \nCtHumanTouchHeadFullAuc\ \nCtHumanTouchHeadFullAucCI & \nCtHumanTouchHeadFullWithin\ \nCtHumanTouchHeadFullWithinCI \\
Real wrist & \nCtEgoTouchWristAuc\ \nCtEgoTouchWristAucCI & \nCtEgoTouchWristWithin\ \nCtEgoTouchWristWithinCI & \nCtHumanTouchWristAuc\ \nCtHumanTouchWristAucCI & \nCtHumanTouchWristWithin\ \nCtHumanTouchWristWithinCI \\
Real wrist through the VAE & \nCtEgoTouchVaeAuc\ \nCtEgoTouchVaeAucCI & \nCtEgoTouchVaeWithin\ \nCtEgoTouchVaeWithinCI & \nCtHumanTouchVaeAuc\ \nCtHumanTouchVaeAucCI & \nCtHumanTouchVaeWithin\ \nCtHumanTouchVaeWithinCI \\
Generated wrist (\method) & \nCtEgoTouchGenAuc\ \nCtEgoTouchGenAucCI & \nCtEgoTouchGenWithin\ \nCtEgoTouchGenWithinCI & \nCtHumanTouchGenAuc\ \nCtHumanTouchGenAucCI & \nCtHumanTouchGenWithin\ \nCtHumanTouchGenWithinCI \\
\bottomrule
\end{tabular}
\caption{\textbf{Contact probes on each input.} AUROC pooled over frames and within-clip AUROC (contact timing within a clip, with task and scene priors removed), with 95\% intervals over clips. The task prior is constant within a clip. The wrist probe is trained on real wrist frames and applied unchanged to the VAE reconstructions and to the generated frames.}
\label{tab:s_contact}
\end{table*}

\begin{table*}[t]
\centering
\footnotesize
\setlength{\tabcolsep}{3pt}
\begin{tabular}{@{}lcccc@{}}
\toprule
& \multicolumn{2}{c}{EgoTouch} & \multicolumn{2}{c}{HumanTouch} \\
\cmidrule(lr){2-3}\cmidrule(l){4-5}
Difference & AUROC & Within-clip AUROC & AUROC & Within-clip AUROC \\
\midrule
Real wrist $-$ head crop & \nCtEgoTouchWristMinusCrop\ \nCtEgoTouchWristMinusCropCI & \nCtEgoTouchWristMinusCropWithin\ \nCtEgoTouchWristMinusCropWithinCI & \nCtHumanTouchWristMinusCrop\ \nCtHumanTouchWristMinusCropCI & \nCtHumanTouchWristMinusCropWithin\ \nCtHumanTouchWristMinusCropWithinCI \\
Generated $-$ task prior & \nCtEgoTouchGenMinusPrior\ \nCtEgoTouchGenMinusPriorCI & \nCtEgoTouchGenMinusPriorWithin\ \nCtEgoTouchGenMinusPriorWithinCI & \nCtHumanTouchGenMinusPrior\ \nCtHumanTouchGenMinusPriorCI & \nCtHumanTouchGenMinusPriorWithin\ \nCtHumanTouchGenMinusPriorWithinCI \\
Real wrist $-$ generated & \nCtEgoTouchWristMinusGen\ \nCtEgoTouchWristMinusGenCI & \nCtEgoTouchWristMinusGenWithin\ \nCtEgoTouchWristMinusGenWithinCI & \nCtHumanTouchWristMinusGen\ \nCtHumanTouchWristMinusGenCI & \nCtHumanTouchWristMinusGenWithin\ \nCtHumanTouchWristMinusGenWithinCI \\
Generated $-$ head crop & \nCtEgoTouchGenMinusCrop\ \nCtEgoTouchGenMinusCropCI & \nCtEgoTouchGenMinusCropWithin\ \nCtEgoTouchGenMinusCropWithinCI & \nCtHumanTouchGenMinusCrop\ \nCtHumanTouchGenMinusCropCI & \nCtHumanTouchGenMinusCropWithin\ \nCtHumanTouchGenMinusCropWithinCI \\
Real wrist $-$ VAE reconstruction & \nCtEgoTouchWristMinusVae\ \nCtEgoTouchWristMinusVaeCI & \nCtEgoTouchWristMinusVaeWithin\ \nCtEgoTouchWristMinusVaeWithinCI & \nCtHumanTouchWristMinusVae\ \nCtHumanTouchWristMinusVaeCI & \nCtHumanTouchWristMinusVaeWithin\ \nCtHumanTouchWristMinusVaeWithinCI \\
\bottomrule
\end{tabular}
\caption{\textbf{Paired differences} of the probes in \cref{tab:s_contact}, with 95\% intervals from resampling clips. The last row comes from the control run that scored the VAE reconstructions, in which the probes were fit again; its real-wrist values differ slightly from those of \cref{tab:s_contact}.}
\label{tab:s_contact_diff}
\end{table*}

The main paper states that the generated wrist views carry contact information but place contact less accurately in time than real wrist video (\cref{sec:conclusion}).
This section reports the study behind that statement.
Its analysis plan, including how each possible outcome would be read, was written before the confirmatory run.

\paragraph{Labels.}
Two of the training datasets record touch.
EgoTouch~\cite{zhou2026touchanything} provides pressure grids from tactile gloves, and HumanTouch~\cite{lyu2026humantouch,chen2026spark0} provides the force on several regions of each hand.
A frame is labeled as contact when the summed pressure reaches \nCtThrEgoTouchHi on EgoTouch, or when the largest regional force reaches \nCtThrHumanTouchHi\,N on HumanTouch.
It is labeled as no contact below \nCtThrEgoTouchLo and \nCtThrHumanTouchLo\,N, respectively, and frames in between are left out.
The thresholds were set once from the distribution of training frames.

\paragraph{Probes.}
For each dataset, we train linear contact probes on frozen DINOv3 ViT-7B~\cite{simeoni2025dinov3} features, the class token and the mean patch token, standardized and reduced to \nCtPcs principal components.
The probes see real frames of training episodes only.
The regularization strength is chosen by cross-validation grouped by episode, and the two classes are weighted equally.
There is one probe per input: the crop of the head frame around the hand (a square of \nCtRoi times the hand box, resized to \nCtCrop pixels), the whole head frame (one probe per hand), and the real wrist frame.
The probe trained on real wrist frames also scores the generated wrist frames, so a generated frame is judged as if it were real.
As a reference for what can be guessed without looking at the frame, a task prior predicts the training contact rate of the clip's task.

\paragraph{Test data.}
We generate wrist videos with \method ($w{=}1$, as in all analysis experiments) for \nCtEgoTouchClips EgoTouch and \nCtHumanTouchClips HumanTouch test clips and evaluate every second frame.
Only frames in which the hand is visible in the head view are used (\nCtEgoTouchRows and \nCtHumanTouchRows hand-frames), which favors the head-view inputs.
AUROC is pooled over frames.
Within-clip AUROC is averaged over the hand-clips that contain both classes (\nCtEgoTouchWithinN on EgoTouch and \nCtHumanTouchWithinN on HumanTouch).
It removes task and scene priors and asks whether contact is placed at the right time.
Intervals come from \nCtBoot bootstrap resamples of clips.
To separate the content of the generated frames from the quality of the VAE decoder, we also score real wrist frames that were encoded and decoded by the VAE of the video model.

\paragraph{Reading.}
The probes work: on both datasets the real wrist frame is far above the task prior.
The head crop is as accurate as the real wrist frame on both datasets (\cref{tab:s_contact_diff}, first row), so contact can be read from the head view.
We therefore do not motivate wrist views by a claim that the head camera misses contact; the main paper motivates them by the viewpoint that wrist-camera policies use.
The generated wrist frames carry per-frame contact information: they are above the task prior on both datasets.
On EgoTouch they are less accurate than real wrist frames by \nCtEgoTouchWristMinusGen; on HumanTouch the difference (\nCtHumanTouchWristMinusGen \nCtHumanTouchWristMinusGenCI) is not significant, which is not evidence that the two are equally accurate.
The AUROC of real wrist frames minus that of their VAE reconstructions is \nCtEgoTouchWristMinusVae on EgoTouch and \nCtHumanTouchWristMinusVae on HumanTouch, so the gap on EgoTouch comes from the generated content rather than from the decoder.
Since the head crop is as accurate as the real wrist and the generated wrist is below the head crop on both datasets, the model does not carry all of the contact information that is visible in its input into the wrist view.
Paired data with contact or force signals could supervise this directly.

\section{Selecting synthetic training data}
\label{sec:s_select}

\begin{table}[t]
\centering
\small
\setlength{\tabcolsep}{4pt}
\begin{tabular}{@{}lc@{}}
\toprule
$\Delta$ LPIPS ($\times 10^{-3}$) & All \\
\midrule
Selected $-$ random & \nRsiDvsCAllK\ \nRsiDvsCAllKCI \\
Selected $-$ start & \nRsiDvsMAllK\ \nRsiDvsMAllKCI \\
Random $-$ start & \nRsiCvsMAllK\ \nRsiCvsMAllKCI \\
\midrule
Selected $-$ random, seed pair 1 & \nRsiPairOneDvsCAllK\ \nRsiPairOneDvsCAllKCI \\
Selected $-$ random, seed pair 2 & \nRsiPairTwoDvsCAllK\ \nRsiPairTwoDvsCAllKCI \\
Selected $-$ start, seed pair 1 & \nRsiPairOneDvsMAllK\ \nRsiPairOneDvsMAllKCI \\
Selected $-$ start, seed pair 2 & \nRsiPairTwoDvsMAllK\ \nRsiPairTwoDvsMAllKCI \\
\bottomrule
\end{tabular}
\caption{\textbf{First-round data selection on the standard test set} (\nRsiDvsCAllN clips). Paired LPIPS differences with 95\% intervals; negative is better. ``Start'' is \method without guidance (averaged weights); both arms continue from its weights before averaging. The upper rows pool the two seed pairs per clip.}
\label{tab:s_select}
\end{table}

\paragraph{Recipe.}
Both arms of \cref{sec:exp_select} start from the weights of \method before averaging, with a new optimizer.
The learning rate warms up over \nRsiWarm steps to \nRsiLr, stays constant until step \nRsiDecayAt and decays to zero at step \nRsiSteps; we evaluate the averaged weights at the end.
The synthetic windows are added to the paired data as a fifth source with probability \nRsiSynShare, and the four real datasets keep their relative weights.
The batch grows to \nRsiBatch clips so that each step sees about as many real clips as a step of the final stage (\nStgConvBatch).
The selected arm uses the \nRsiKept windows with the highest HCC, and the random arm uses as many windows drawn at random; \nRsiOverlap windows are in both.
Each seed pair trains both arms with the same seed, and the evaluation pools the two pairs per clip.

\paragraph{An earlier recipe that did not work.}
We first tried a different recipe on the same windows: the synthetic data replaced \nRoneSynShare of the real samples instead of being added, and the learning rate was raised again to \nRoneLr for \nRoneSteps steps.
Selected and random data then gave the same result (\nRoneDvsCAll \nRoneDvsCAllCI), and the selected arm ended worse than the starting model (\nRoneDvsMAll \nRoneDvsMAllCI).
A control trained on real data only with the same schedule was worse than the starting model by a similar amount (\nRoneAvsMAll \nRoneAvsMAllCI, an interval that includes zero), which points to the schedule rather than to the synthetic data.
The recipe above changes only these two parts, and its decision rule was fixed before training.

\paragraph{Results per seed pair.}
\Cref{tab:s_select} gives the first round of \cref{sec:exp_select} pooled over the two seed pairs and for each seed pair.
M1 of the later rounds is the selected arm of the seed pair with the lower validation LPIPS (seed pair 2).
On the extended EgoPro set, no difference between these arms is significant (\cref{sec:exp_select}).

\paragraph{Windows of the later rounds.}
The windows of the second round come from workers that the first round did not use, in the half of the factories whose scenes are closest to the training split (ranked by CLIP similarity to head frames of the training split, before round 2).
The third round uses new windows from the videos of the second round.
HCC keeps the top 30\% in each round: \nRsiTwoKept of \nRsiTwoPool windows in the second round and \nRsiThreeKept of \nRsiThreePool in the third.

\paragraph{More rounds.}
With more rounds we observe a tendency toward collapse on in-the-wild input: the wrist views that the later models synthesize for different windows become more similar to each other, while their outputs on the test clips do not change.
Each round already mixes the synthetic windows into the paired data (Recipe above); running the loop for more rounds likely needs a larger share of replayed real training data to avoid catastrophic forgetting, and different noise for each window.

\section{Training progress}
\label{sec:s_progress}

\begin{table*}[t]
\centering
\small
\setlength{\tabcolsep}{6pt}
\begin{tabular}{@{}lrlcccccc@{}}
\toprule
After stage & Windows seen & Data & EgoPro & unseen env. & EgoTouch & EgoViz & HumanTouch & All \\
\midrule
1 & \nProgCTwoWindows & \nHoursVTwo\,h & \nProgCTwoEgoPro & \nProgCTwoUnseen & \nProgCTwoEgoTouch & \nProgCTwoEgoViz & \nProgCTwoHumanTouch & \nProgCTwoAll \\
2 & \nProgCTwoLongWindows & \nHoursVTwo\,h & \nProgCTwoLongEgoPro & \nProgCTwoLongUnseen & \nProgCTwoLongEgoTouch & \nProgCTwoLongEgoViz & \nProgCTwoLongHumanTouch & \nProgCTwoLongAll \\
3 & \nProgBigWindows & \nHoursVTwo\,h & \nProgBigEgoPro & \nProgBigUnseen & \nProgBigEgoTouch & \nProgBigEgoViz & \nProgBigHumanTouch & \nProgBigAll \\
4 (long run) & \nProgLongWindows & \nHoursTotal\,h & \nProgLongEgoPro & \nProgLongUnseen & \nProgLongEgoTouch & \nProgLongEgoViz & \nProgLongHumanTouch & \nProgLongAll \\
5 (final, \method) & \nProgConvWindows & \nHoursTotal\,h & \nProgConvEgoPro & \nProgConvUnseen & \nProgConvEgoTouch & \nProgConvEgoViz & \nProgConvHumanTouch & \nProgConvAll \\
\bottomrule
\end{tabular}
\caption{\textbf{LPIPS on the standard test set after each training stage} (\nSetStd clips, $w{=}1$, lower is better). Windows seen are cumulative. Stages 1 to 3 have no weight averaging, so their raw weights are evaluated; stages 4 and 5 use the averaged weights. ``Unseen env.'' is the subset of EgoPro clips from environments that never appear in training.}
\label{tab:s_progress}
\end{table*}

\Cref{tab:s_progress} reports LPIPS on the standard test set after each training stage of \cref{tab:s_stages}, all sampled without guidance.
The amount of paired data, the number of training windows and the resolution changed together during training, so the table shows how the model improved over the course of training and does not isolate the effect of data.
A controlled data-scaling experiment, with the amount of paired data varied at a fixed training budget, is left to future work.

\section{Human studies}
\label{sec:s_study}

\begin{table*}[t]
\centering
\small
\setlength{\tabcolsep}{4pt}
\begin{tabular}{@{}llcc@{}}
\toprule
Study & Items & Share preferring & Majority vote \\
\midrule
H1a, with ground truth & \nStudyHOneAItems clips & \nStudyHOneAPref\ \nStudyHOneAPrefCI & \nStudyHOneAMajFor\ of \nStudyHOneAMajN \\
\quad EgoPro & 12 clips & \nStudyHOneAEgoPro\ \nStudyHOneAEgoProCI & \\
\quad EgoTouch & 12 clips & \nStudyHOneAEgoTouch\ \nStudyHOneAEgoTouchCI & \\
\quad EgoViz & 12 clips & \nStudyHOneAEgoViz\ \nStudyHOneAEgoVizCI & \\
\quad HumanTouch & 12 clips & \nStudyHOneAHumanTouch\ \nStudyHOneAHumanTouchCI & \\
H1b, in the wild & \nStudyHOneB windows & \nStudyHOneBPref\ \nStudyHOneBPrefCI & \nStudyHOneBMajFor\ of \nStudyHOneBMajN \\
H2, kept vs.\ dropped window & \nStudyHTwo videos & \nStudyHTwoPref\ \nStudyHTwoPrefCI & \nStudyHTwoMajFor\ of \nStudyHTwoMajN \\
\midrule
HCC agrees with raters (H1b) & \nStudyHOneB windows & \nStudyHOneBHcc\ \nStudyHOneBHccCI & \nStudyHOneBHccMaj\ of \nStudyHOneBHccMajN \\
\bottomrule
\end{tabular}
\caption{\textbf{Second human study} (\nStudyRaters raters, \nStudyAnswers answers including attention checks and repeats; \nStudyBlocksPerQ answers on each of the \nStudyUnique unique questions). Share of answers preferring \method (H1a, H1b) or the window HCC keeps (H2), with 95\% bootstrap intervals over items; majority vote per item. Last row: share of answers whose choice HCC predicts.}
\label{tab:s_study}
\end{table*}

\paragraph{HCC validation (complete).}
The agreement study of \cref{sec:exp_score} collected \nHumAnswers answers from \nHumRaters raters.
Each question showed the head video and the real video of one wrist camera of an EgoPro, EgoViz or HumanTouch test clip, then two generated videos of that wrist, and asked which is closer to the real one.
The two videos came either from two checkpoints of our model or from two samples of the same checkpoint.
The analysis leaves out the repeated questions and \nHumExcluded rater, who was close to chance on subtle questions; of the remaining \nHumAnalysed answers from \nHumAnalysedRaters raters, every score can be computed on \nHumScored (\nHumNCross across checkpoints, \nHumNSame within a model).
The agreement between two people (\nHumHumanCross across checkpoints, \nHumHumanSame within a model) is $2p-1$, where $p$ is the rate at which two people who answered the same question made the same choice.
HCC was designed after these answers were collected (\cref{sec:s_score_roi}).

\paragraph{Comparison with WorldWander and test of the data selection.}
The second study answers two questions.
H1 asks whether people prefer \method over WorldWander fine-tuned on our data, with ground truth (H1a) and in the wild (H1b).
H2 asks whether a window that HCC keeps for training is better synthesized, in people's eyes, than a window it drops.
The design and the analysis below were registered before the first answer was collected; the results follow the design.

\emph{Material.}
H1a uses \nStudyHOneA test clips, \nStudyHOneAPerDs per dataset, drawn at random from the \nXfamN clips that no score selected; both methods are sampled as in \cref{tab:main}.
H1b uses \nStudyHOneB Egocentric-10K windows drawn at random from the screened and captioned windows of \cref{tab:s_funnel}, without the source videos used in H2 and a few windows we had inspected before (\nStudyHOneBEligible eligible windows), with no two windows from the same source video and at most \nStudyHOneBPerFactory from one factory (\nStudyHOneBFactories factories).
These windows are not filtered by HCC.
Both methods receive the same head video and the same caption.
H2 uses \nStudyHTwo pairs of windows from the selection pool of \cref{sec:exp_select}.
In each pair, the two windows come from the same source video: one kept by HCC (top \nRsiKeepShare) and the closest window in time that was not kept.
The pairs are drawn at random from the \nStudyHTwoEligible eligible source videos, at most \nStudyHTwoPerFactory per factory (\nStudyHTwoFactories factories); the median gap is \nStudyHTwoGapFrames frames and the median HCC difference \nStudyHTwoScoreGap.
Each H2 window shows the synthesis that the selection experiment trained on.

\emph{Questions.}
Every question is a two-alternative forced choice; method names and scores are never shown, and the sides are random.
An H1a page shows the head video and the real wrist videos on top, labeled as real, and the two candidate pairs below.
An H1b page shows the head video and the two candidate pairs, and asks which pair looks more like what wrist cameras would record.
An H2 page shows the two windows side by side, each with its own head video above its wrists, and asks raters not to choose by scene.
The criteria are listed on every page: hand pose, held object, viewpoint and timing for H1a; consistency with the head video, a plausible wrist viewpoint and stability over time for H1b; realistic hands, consistency with the window's own head video and stability for H2.

\emph{Blocks, checks and repeats.}
The \nStudyUnique unique questions are split into \nStudyBlocks blocks so that every question appears in \nStudyBlocksPerQ blocks; with \nStudyRatersTarget raters, each question receives \nStudyBlocksPerQ answers.
A block holds \nStudyUniqueMin to \nStudyUniqueMax unique questions.
Every rater also answers \nStudyChecks attention checks, the same for everyone, in which one side is clearly wrong: in H1a, the real wrist videos of a clip from another dataset; in H1b and H2, the wrists of another factory with the frames in random order.
Every rater further sees \nStudyRepeats of their own questions again with the sides swapped, at least \nStudyRepeatGap questions after or before the original, which measures self-consistency.
A session therefore has \nStudyPerRaterMin to \nStudyPerRaterMax questions, in an order drawn for each rater.

\emph{Exclusion and analysis.}
A rater who fails \nStudyFailExclude or more checks, or answers fewer than \nStudyMinChecks of them, is excluded from the main analysis and reported separately.
For each study we report the share of answers that prefer \method (H1) or the kept window (H2), with a percentile bootstrap interval (\nStudyBoot resamples) over clips, windows or source videos.
A claim holds only if the lower end of the interval is above chance.
H1a is also reported per dataset, and every study by majority vote per item.
For H2 we report the same share by tercile of the HCC difference, which tests whether larger differences predict stronger preferences.
On the H1b pairs we also compute HCC, to measure how often it agrees with people across methods.
Sensitivity analyses repeat everything with all raters, and without raters who took part in our earlier studies.

\emph{Results.}
Collection ended when each of the \nStudyBlocks blocks had been completed by one rater.
The \nStudyRaters raters who completed a block passed all \nStudyChecks attention checks and enter the main analysis (\nStudyAnswers answers, checks and repeats included); answers from an incomplete session that answered no check are excluded by the rule above.
\Cref{tab:s_study} gives the results.
With ground truth (H1a), raters prefer \method over WorldWander in \nStudyHOneAPref \nStudyHOneAPrefCI of their answers and on \nStudyHOneAMajFor of \nStudyHOneAMajN clips by majority vote; the preference holds on EgoTouch, EgoViz and HumanTouch and is not significant on EgoPro.
In the wild (H1b), raters show no preference: \nStudyHOneBPref \nStudyHOneBPrefCI of the answers prefer \method, and the majority vote splits \nStudyHOneBMajFor to \nStudyHOneBMajAgainst.
We think the missing reference explains this.
Without the real wrist video, a rater can judge whether a wrist view looks plausible and consistent with the head video, but not whether it shows the scene from the right viewpoint, and the errors of WorldWander lie mostly in viewpoint and layout (\cref{fig:qual}); its views of a plausible hand and object are hard to tell apart from ours without a reference.
This is our interpretation; the study does not test it.
Raters are about as consistent with each other in H1b as in H1a (pairwise agreement \nStudyHOneBAgree and \nStudyHOneAAgree), so they do judge the two sides consistently per window, but which side they prefer varies from window to window.
On the same H1b pairs, HCC agrees with the rater's choice in \nStudyHOneBHcc \nStudyHOneBHccCI of the answers and with the majority on \nStudyHOneBHccMaj of \nStudyHOneBHccMajN windows; these answers were collected after HCC was fixed.
For the data selection (H2), raters prefer the window that HCC keeps in \nStudyHTwoPref \nStudyHTwoPrefCI of their answers and on \nStudyHTwoMajFor of \nStudyHTwoMajN videos by majority vote, in every tercile of the HCC difference (\nStudyHTwoLow, \nStudyHTwoMid and \nStudyHTwoHigh from the smallest to the largest difference; highest minus lowest \nStudyHTwoHighLow \nStudyHTwoHighLowCI).
A rater shown one of their own questions again with the sides swapped makes the same choice in \nStudyRepeatSame of the cases (\nStudyRepeatSameK of \nStudyRepeatSameN), which shows how noisy single answers are.
The sensitivity analyses agree: with all raters, H1a \nStudyHOneAPrefAllRaters, H1b \nStudyHOneBPrefAllRaters and H2 \nStudyHTwoPrefAllRaters; without the raters of our earlier studies (\nStudyRatersNoEarlier raters), \nStudyHOneAPrefNoEarlier, \nStudyHOneBPrefNoEarlier and \nStudyHTwoPrefNoEarlier.

\emph{What the design cannot show.}
Each H2 window has a single sample, so H2 cannot tell whether HCC prefers easy inputs or recognizes good synthesis.
Both readings are useful for selecting training data, but only the second would make HCC a sample-level judge, which \cref{sec:exp_score} already shows it is not.

\section{More qualitative results}
\label{sec:s_qual}

\begin{figure*}[t]
\centering
\includegraphics[width=\textwidth]{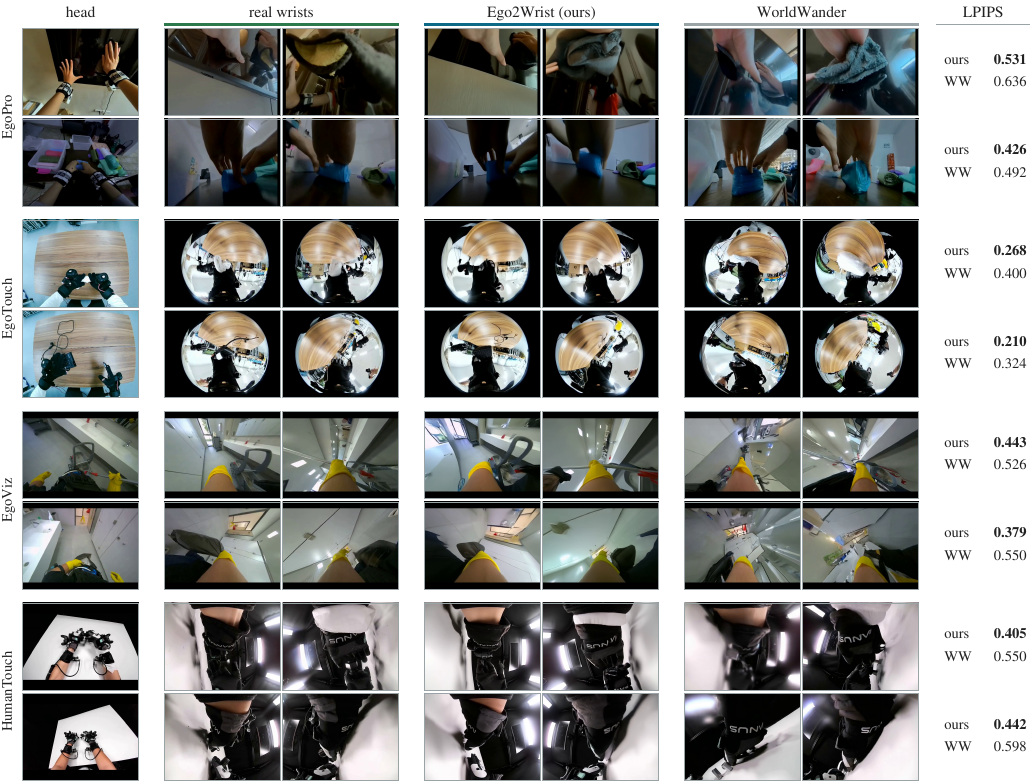}
\caption{\textbf{Random test clips} (the middle frame of each video): the head input, the real wrists, \method with head guidance, and WorldWander fine-tuned on our data, with the LPIPS of each clip averaged over both wrists. The clips were drawn at random before any frame was looked at.}
\label{fig:s_qual}
\end{figure*}

\begin{figure*}[p]
\centering
\includegraphics[width=\textwidth]{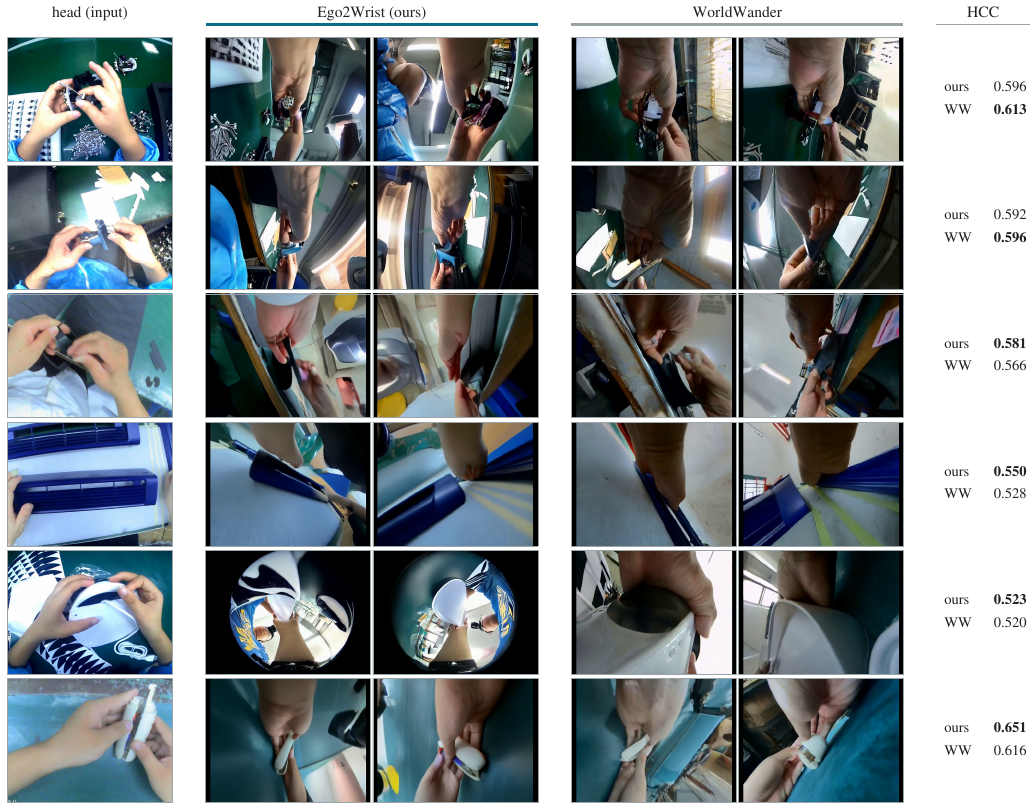}
\caption{\textbf{Random in-the-wild windows} (Egocentric-10K, frame 24; no wrist ground truth): the head input, \method with head guidance, and WorldWander fine-tuned on our data, with the HCC of each sample (mean over both wrists; higher means more consistent with the head view). The windows were drawn at random before any frame was looked at.}
\label{fig:s_wild}
\end{figure*}

\Cref{fig:s_qual} shows test clips drawn at random, two per dataset, in the layout of \cref{fig:qual}.
They come from the unselected set (\cref{sec:s_evalsets}), excluding the clips that are questions of the human study (\cref{sec:s_study}) and those already shown in the main paper.
\Cref{fig:s_wild} shows in-the-wild windows of Egocentric-10K, drawn at random from the H1b windows of the second human study (\cref{sec:s_study}) after its collection ended, excluding the two windows of \cref{fig:teaser}.
There is no wrist ground truth, so the figure gives the HCC of each sample instead of LPIPS.
In the fifth row our model renders a circular-fisheye view, which the target wrist camera never produces; the selection pool of \cref{sec:exp_select} drops such windows before scoring (\cref{sec:s_wild}).

\end{document}